\RequirePackage{fix-cm}
\documentclass[10pt, twocolumn]{svjour3}          

\smartqed  
\usepackage{graphicx}
\usepackage{times}
\usepackage{amsmath}
\usepackage{amssymb}
\usepackage{appendix}
\usepackage{pifont}
\usepackage{cite}
\usepackage{booktabs}
\usepackage{array}
\usepackage{color}
\usepackage{tabularx}
\usepackage{longtable}
\usepackage{tabu}
\usepackage{url}
\usepackage[misc]{ifsym}
\usepackage{float}
\usepackage{bm}
\usepackage{threeparttable}
\usepackage{footnote}
\usepackage{booktabs}
\usepackage{multirow}
\usepackage{mathtools}
\usepackage{natbib}
\usepackage[ruled,linesnumbered]{algorithm2e}
\usepackage{algorithmic}
\usepackage{subfigure}
\newcommand{\eg}{\textit{e}.\textit{g}.}
\newcommand{\ie}{\textit{i}.\textit{e}.}

\begin{document}

\title{Exposing Semantic Segmentation Failures via Maximum Discrepancy Competition}




\author{Jiebin Yan \textsuperscript{1} \and Yu Zhong \textsuperscript{1} \and Yuming Fang \textsuperscript{1}  \and Zhangyang Wang \textsuperscript{2} \and Kede Ma \textsuperscript{3}
}


\institute{Jiebin Yan\\
          \vspace{0.1cm}\email{jiebinyan@foxmail.com}\\
           Yu Zhong\\
          \vspace{0.1cm}\email{zhystu@qq.com}\\
           Yuming Fang (Corresponding Author)\\
          \vspace{0.1cm}\email{fa0001ng@e.ntu.edu.sg}\\
           Zhangyang Wang\\
           \vspace{0.1cm}\email{atlaswang@utexas.edu}\\
           Kede Ma\\
          \vspace{0.1cm}\email{kede.ma@cityu.edu.hk}\\
            \at
            {1} School of Information Technology, Jiangxi University of Finance and Economics, Nanchang, Jiangxi, China.
             \at
            {2} Department of Electrical and Computer Engineering, The University of Texas at Austin, Austin, Texas, USA.
            \at
            {3} Department of Computer Science, City University of Hong Kong, Kowloon, Hong Kong.
}

\date{}


\maketitle

\begin{abstract}
Semantic segmentation is an extensively studied task in computer vision, with numerous methods proposed every year. Thanks to the advent of deep learning in semantic segmentation, the performance on existing benchmarks is close to saturation. A natural question then arises: Does the superior performance on the closed (and frequently re-used) test sets transfer to the open visual world with unconstrained variations? In this paper, we take steps toward answering the question by exposing failures of existing semantic segmentation methods in the open visual world under the constraint of very limited human labeling effort. Inspired by previous research on model falsification, we start from an arbitrarily large image set, and automatically sample a small image set by MAximizing the Discrepancy (MAD) between
two segmentation methods. The selected images have the greatest potential in falsifying either (or both) of the two  methods. We also explicitly enforce several conditions to diversify the exposed failures, corresponding to different underlying root causes. A segmentation method, whose failures are more difficult to be exposed in the MAD competition, is considered better. We conduct a thorough MAD diagnosis of  ten PASCAL VOC  semantic segmentation algorithms. With detailed analysis of experimental results, we point out strengths and weaknesses of the competing algorithms, as well as potential research directions for further advancement in semantic segmentation. The codes are publicly available at \url{https://github.com/QTJiebin/MAD_Segmentation}.

\end{abstract}


\section{Introduction}
\label{sec_intro}
Deep learning techniques have been predominant in various computer vision tasks, such as image classification~\citep{krizhevsky2012imagenet,Simonyan2014, He2016ResNet}, object detection~\citep{Ren2015}, and semantic segmentation~\citep{Long2015}. One important reason for the remarkable successes of deep learning techniques is the establishment of large-scale human-labeled databases for different vision tasks. However, years of model development on the same benchmarks may raise the risk of overfitting due to excessive re-use of the test data. This
poses a new challenge for performance comparison in computer vision:
\begin{itemize}
    \item[] \textit{How to probe the generalization of ``top-performing'' computer vision methods (measured on closed and well-established test sets) to the open visual world with much greater content variations?}
\end{itemize}

Here we focus our study on semantic segmentation, which involves partitioning a digital image into semantically meaningful segments, for two main reasons. First, previous studies on generalizability testing are mostly restricted to image classification \citep{Goodfellow2014},  while little attention has been received for semantic segmentation despite its close relationship to many high-stakes applications, such as self-driving cars and computer-aided diagnosis systems. Second, semantic segmentation requires pixel-level dense labeling, which is a much more expensive and time-consuming endeavor compared to that of image classification and object detection. According to \cite{Everingham2015}, it can easily take ten times as long to segment an object than to draw a bounding box around it, making the human labeling budget even
more unaffordable for this particular task. As it is next to impossible to create a larger human-labeled image set as a closer approximation of the open visual world, the previous question may be rephrased in the context of semantic segmentation:
\begin{itemize}
    \item[] \textit{How to leverage massive unlabeled images from the open visual world to test the generalizability of semantic segmentation methods under a very limited human labeling budget?}
\end{itemize}


Inspired by previous studies on model falsification in the field of computational neuroscience \citep{golan2019controversial}, software testing \citep{mckeeman1998}, image processing \citep{ma2018group}, and computer vision \citep{wang2020going}, we take steps toward answering the question by efficiently exposing semantic segmentation failures using the  MAximum Discrepancy (MAD) competition\footnote{Note that MAD is different from ``Maximum Mean Discrepancy" \citep{tzeng2014deep}. The former means maximizing the difference between the predictions of two test models, while the latter is a method of computing distances between probability distributions.} \citep{wang2020going}. Specifically, given a web-scale unlabeled image set, we automatically mine a small image set by maximizing the discrepancy between two semantic segmentation methods. The selected images are the most informative in terms of discriminating between two methods. Therefore, they have the great potential to be the
failures of at least one method. We also specify a few conditions to encourage spotting more diverse failures, which may correspond to  different underlying  causes. We seek such small sets of images for every distinct pair of the competing segmentation methods, and merge them into one, which we term as the \textit{MAD image set}. Afterwards,
 we allocate the limited human labeling budget to the MAD set, sourcing the  ground-truth segmentation map for each selected image. Comparison of model predictions and human labels on the MAD set strongly indicates the relative performance of the competing models. In this case, a better segmentation method is the one with fewer failures exposed by other algorithms.
 We apply this general model falsification method to state-of-the-art semantic segmentation methods trained on the popular benchmark of PASCAL VOC \citep{Everingham2015}. Analysis of the resulting MAD set yields a number of insightful and previously overlooked findings on the  PASCAL VOC dataset.

In summary, our contributions include:
\begin{itemize}
\item A computational framework to efficiently expose diverse and even rare semantic segmentation failures based on the MAD competition.
\item An extensive demonstration of our method to diagnose ten semantic segmentation methods on PASCAL VOC, where we spot the generalization pitfalls of even the strongest segmentation method so far, pointing to potential directions for improvement.
\end{itemize}

\section{Related Work}
In this section, we first give a detailed overview of semantic segmentation, with emphasis on several key modules that have been proven useful to boost performance on standard benchmarks. We then review emerging ideas to test different aspects of model generalizability in computer vision.
\label{sec_relat}
\subsection{Semantic Segmentation}
\label{subsec_seg}
Traditional methods~\citep{Csurka2008, Shotton2008, Cao2007} relied exclusively on hand-crafted features such as textons~\citep{Julesz1981}, histogram of oriented gradients~\citep{Dalal2005}, and bag-of-visual-words \citep{kadir2001saliency,Lowe2004,russell2006using}, coupled with  different machine learning techniques, including random forests~\citep{Shotton2008}, support vector machines~\citep{Yang2012}, and conditional random fields \citep{verbeek2008scene}. The performance of these knowledge-driven methods is highly dependent on the quality of  the hand-crafted features, which are bound to have limited expressive power. With the
introduction of fully convolutional networks \citep{Long2015}, many existing semantic segmentation algorithms practiced end-to-end joint optimization of feature extraction, grouping and classification, and obtained promising results on closed and well-established benchmarks. Here, we summarize four key design choices of convolutional neural networks (CNNs) that significantly advance the state-of-the-art of semantic segmentation.

\emph{1) Dilated Convolution}~\citep{Yu2016}, also known as  atrous convolution, is specifically designed  for dense prediction. With an extra parameter to control the rate of dilation, it aggregates multi-scale context information without sacrificing  spatial resolution. However, dilated convolution comes with the so-called  gridding problem. \cite{wang2018understanding} proposed  hybrid dilated convolution to effectively alleviate the gridding issue by gradually increasing the dilation rates in each stage of convolutions. \cite{yu2017dilated} developed dilated residual networks for semantic segmentation with  max pooling removed. DeepLabv3+~\citep{Chen2018v3+}, the latest version of the well-known DeepLab family~\citep{Chen2017v2, Chen2017v3, Chen2018v3+} combined dilated convolution with spatial pyramid pooling for improved extraction of multi-scale contextual information.

\emph{2) Skip Connection}. In the study of fully convolutional networks for semantic segmentation, \cite{Long2015} adopted skip connections to fuse low-level and high-level information, which is helpful for generating sharper boundaries.  U-Net~\citep{ronneberger2015u} and its variant~\citep{zhou2018unet++} used dense skip connections to combine feature maps of the encoder and the corresponding decoder, and achieved great successes in medical image segmentation. SegNet~\citep{badrinarayanan2017segnet} implemented a similar network architecture for semantic segmentation, and compensated for the reduced spatial resolution with index-preserving max pooling. \cite{Fu2019} proposed a stacked deconvolutional network by stacking multiple small deconvolutional networks (also called ``units'') for fine recovery of localization information. The skip connections were made for both inter-units and intra-units to enhance feature fusion and assist network training.

\emph{3) Multi-Scale Computation}. Multi-scale perception is an important characteristic of the human visual system (HVS), and there have been many computational models, trying to mimic this type of human perception for various tasks in signal and image processing \citep{simoncelli1995steerable}, computer vision \citep{Lowe2004}, and computational neuroscience \citep{wandell1997foundations}. In the context of CNN-based semantic segmentation methods, the multi-scale perception can be formed in three ways. The most straightforward implementation is to feed different resolutions of an input image to one CNN, as a way of encouraging the network to capture  discriminative features at multiple scales~\citep{chen2019collaborative}. An advanced implementation of the same idea is to first construct a pyramid representation of the input image (\eg, a Gaussian, Laplacian \citep{burt1981fast}, or steerable pyramid \citep{simoncelli1995steerable}), and then train CNNs to process different subbands, followed by fusion. As CNNs with spatial pooling are born with multi-scale representations, the second type of implementation explores  feature pyramid pooling. Specifically, \cite{Zhao2017}
developed  a pyramid pooling module  to summarize both local and global context information. An atrous spatial pyramid pooling model was introduced in~\citep{Chen2017v3}. The third type of implementation makes use of skip connections to directly concatenate features of different scales for subsequent processing. RefineNet~\citep{LinGS2017} proposed a multi-resolution fusion block to predict the segmentation map from the skip-connected feature maps of different resolutions. \cite{chen2019fasterseg} utilized neural architecture search techniques to automatically identify and integrate multi-resolution branches.

\emph{4) Attention Mechanism}. Visual attention refers to the cognitive process of the HVS that allows us to concentrate on potentially valuable features, while ignoring less important visual information. The attention mechanism implemented in computer vision is  essentially similar to that of the HVS, whose goal is to extract and leverage the most relevant information to the vision task at hand. Some studies attempted to advance semantic segmentation by incorporating attention mechanism. \cite{li2018pyramid} proposed a feature pyramid attention module to combine attention mechanism with pyramid representation, resulting in more precise dense prediction. \cite{fu2019dual} presented a self-attention mechanism in semantic segmentation, which involves a position attention module to aggregate important spatial features scattered across different locations, and a channel attention module to emphasize interdependent feature maps. \cite{zhao2018psanet} proposed a point-wise spatial attention network to aggregate long-range contextual information and to enable bi-directional information propagation. Similarly, \cite{huang2019ccnet} introduced a criss-cross network for efficiently capturing full-image contextual information with reduced computational complexity. \cite{Lixia2019} developed  expectation maximization attention networks to compute  attention maps on a compact set of basis vectors.

Apart from the pursuit of accuracy, much effort has also been made toward resource-efficient and/or real-time semantic segmentation. Separable convolution \citep{Peng2017,Mehta2019dicenet,mehta2019espnetv2} is a common practice to maintain a large kernel size, while reducing floating point operations. \cite{dvornik2017blitznet} proposed BlitzNet for real-time semantic segmentation and object detection.~\cite{zhao2018icnet} proposed an image cascade network to strike a balance between efficiency and accuracy in processing low-resolution and  high-resolution images. The low-resolution branch is responsible for coarse prediction, while the medium- and high-resolution branches are used for segmentation refinement. A similar idea was leveraged in \cite{chen2019collaborative} to process high-resolution images with memory efficiency. \cite{nekrasov2018light} proposed a lighter version of RefineNet by reducing model parameters and floating point operations. \cite{chen2019fasterseg} presented an AutoML-designed semantic segmentation network with not only state-of-the-art performance but also faster speed than current methods. In this work, we only evaluate and compare the semantic segmentation methods in terms of their accuracies in the open visual world, and do not touch on the efficiency aspect (\ie, computational complexity).

\subsection{Tests of Generalizability in Computer Vision}
\label{subsec_gener}
In the pre-dataset era, the vast majority of computer vision algorithms for various tasks were hand-engineered, whose successes were declared on a very limited number of visual examples \citep{canny}. Back then, computer vision was more of an art than a science; it was unclear which methods  would generalize better for a particular task. \cite{amfm_pami2011} compiled one of the first ``large-scale'' human-labeled database for evaluating edge detection algorithms. Interestingly, this dataset is still in active use for the development of CNN-based edge detection methods \citep{xie2015holistically}. \cite{mikolajczyk2005performance} conducted a systematic performance evaluation of several local descriptors on a set of images with different geometric and photometric transformations, \eg, viewpoint change and JPEG compression. The scale-invariant feature transform (SIFT) \citep{Lowe2004} cut a splendid figure in this synthetic dataset with easily inferred ground-truth correspondences. This turned out to be a huge success in computer vision. Similar dataset-driven breakthroughs include the structural similarity index \citep{wang2004image} for image quality assessment thanks to the LIVE database \citep{sheikh2006statistical}, deformable part models \citep{felzenszwalb2009object} for object detection thanks to the PASCAL VOC \citep{Everingham2015}, and CNNs for image classification (and general computer vision tasks) thanks to the ImageNet \citep{Deng2009}.

Nowadays, nearly all computer vision methods rely on human-labeled data for algorithm development. As a result, the \textit {de facto} means of probing generalization is to split the available dataset into two (or three) subsets: one for training, (one for validation), and the other for testing. To facilitate fair comparison, the splitting is typically fixed. The goodness of fit on the test set provides an approximation to generalization beyond training. Recent years have witnessed the successful application of the independent and identically distributed (i.i.d.) evaluation to monitor the progress of many subfields of computer vision, especially where large-scale labeled datasets are available.

Despite demonstrated successes, computer vision researchers began to reflect on several pitfalls of their evaluation strategy. First, the test sets are pre-selected, fixed, and re-used. This leaves open the door for adapting methods to the test sets (intentionally or unintentionally) via extensive hyperparameter tuning, raising the risk of overfitting. For example, the ImageNet Large Scale Visual Recognition Challenge~\citep{russakovsky2015imagenet} was terminated in 2017, which might give us an impression that image classification has been solved. Yet \cite{Recht2019} investigated the generalizability of ImageNet-trained classifiers by building a new test set with the same protocol of ImageNet. They, however, found that the accuracy drops noticeably (from $11\%$ to $14\%$) for a broad range of image classifiers. This reveals the much underestimated difficulty of ensuring a sufficiently representative and unbiased test set. The situation would only worsen if human labeling requires extensive domain expertise (\eg, medical applications \citep{jia2019cone}), or if failures are rare but fatal in high-stakes and failure-sensitive applications (\eg, autonomous cars \citep{mohseni2019practical,chen2020automated}).

More importantly, the i.i.d. evaluation method may not reveal the shortcut learning problem \citep{zhang2019interpreting}, where shortcuts mean a set of decision rules that perform well on the training set and i.i.d. test set due to the dataset insufficiency or selection bias, but fail to generalize to challenging testing environments. \cite{geirhos2018imagenet} showed that CNNs trained on ImageNet are overly relying on texture appearances, instead of object shapes as used by humans for image classification. To leverage the spotted shortcut, \cite{brendel2019} proposed a learnable bag-of-feature model to classify an image by ``counting texture patches" without considering their spatial ordering, and achieved high performance on ImageNet.

Recently, researchers began to create various \textit{fixed} out-of-distribution (o.o.d.) test sets to probe model generalizability mainly in the area of image classification. Here, we use the o.o.d. test set to broadly include: 1) samples that are drawn from the same underlying distribution of the i.i.d. training and test data, but belong to different subpopulations \citep{jacobsen2018excessive}, and 2) samples that are from a systematically different distribution. For example, \cite{hendrycks2019natural} compiled a set of natural adversarial examples that represent a hard subpopulation of the space of all natural images, causing ImageNet classifiers to degrade significantly. \cite{hendrycks2019benchmarking} evaluated the robustness of ImageNet classifiers to common corruptions and perturbations, such as noise, blur, and weather. \cite{geirhos2018imagenet} proposed to use synthetic stimuli of conflicting cues by transferring the textures of one object (\eg, an Indian elephant) to another object (\eg, a tabby cat), while keeping the main structures. \cite{li2017deeper} tested classifiers on sketch, cartoon, and painting domains. Although these fixed o.o.d. test sets successfully reveal some shortcuts of image classifiers, they suffer from the potential problem of extensive re-use, similar as fixed i.i.d. test sets.

\begin{figure*}[!ht]
  \includegraphics[width=1\linewidth]{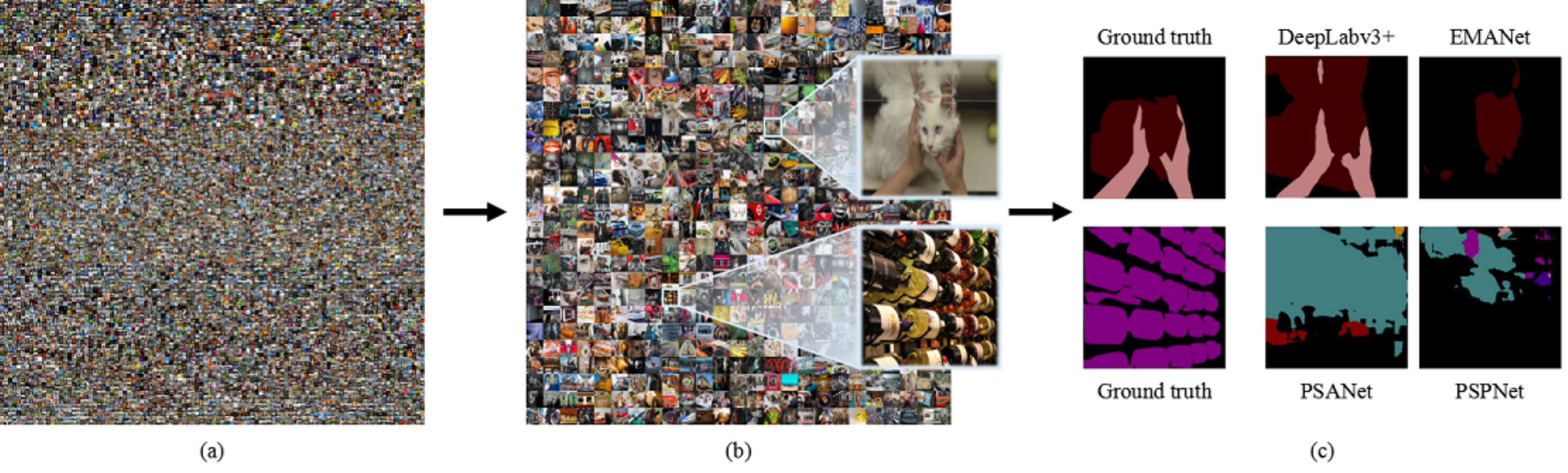}
\caption{The pipeline of the MAD competition for semantic segmentation. \textbf{(a)} A web-scale image set $\mathcal{D}$ that provides a closer approximation to the space of all possible natural images. It is worth noting that dense labeling each image in $\mathcal{D}$ is prohibitively expensive (if not impossible). \textbf{(b)} The MAD image set $\mathcal{S}\subset\mathcal{D}$, in which each image is able to best discriminate two out of $J$ semantic segmentation models. $\mathcal{S}$ is constructed by maximizing the discrepancy between two models while satisfying a set of constraints to ensure the diversity of exposed failures. The size of $\mathcal{S}$ can be adjusted to accommodate the available human labeling budget. \textbf{(c)} Representative images from $\mathcal{S}$ together with the ground-truth maps and the predictions associated with the two segmentation models. Although many of the chosen models have achieved excellent performance on PASCAL VOC, MAD is able to automatically expose their respective failures, which are in turn used to rank their generalization to the open visual world.}
\label{fig:framework}
\end{figure*}

Over decades, Ulf Grenander suggested to study computer vision from the pattern theory perspective \citep{grenander2012pattern, grenander2012lectures, grenander2012regular}. \cite{mumford1994pattern} defined pattern theory as:
\begin{itemize}
    \item [] \textit{The analysis of the patterns generated by the world in any modality, with all their naturally occurring complexity and ambiguity, with
the goal of reconstructing the processes, objects and events that
produced them.}
\end{itemize}
According to the pattern theory, if one wants to test whether a computational method relies on intended (rather than shortcut) features for a specific vision task, the set of features should be tested in a generative (not a discriminative) way. This is the idea of ``analysis by synthesis'', and is well-demonstrated in the field of texture modeling \citep{julesz1962visual}. Specifically, classification based on some fixed example textures provides a fairly weak test of the sufficiency of a set of statistics in capturing the appearances of visual textures. A more efficient test is trying to synthesize a texture image by forcing it to satisfy the set of statistical constraints given by an example texture image \citep{portilla2000parametric}. If the synthesized image looks identical to the example image for a wide variety of texture patterns, the candidate set of features is probably sufficient (\ie, intended) to characterize the statistical regularities of visual textures, and is also expected to perform well in the task of texture classification. By contrast, if the set of features is insufficient for texture synthesis, the synthesized image would satisfy the identical statistical constraints, but look different. In other words, a strong failure of the texture model under testing is identified,  which is highly unlikely to be included in the pre-selected and fixed test set for the purpose of texture classification.

The recent discovery of adversarial samples~\citep{Goodfellow2014} of CNN-based image classifiers can also be cast into the framework of ``analysis by synthesis''. That is, imperceptible perturbations are added to synthesize o.o.d. images, leading to well-trained classifiers to make wrong predictions. \cite{jacobsen2018excessive} described a means of synthesizing images that have exact the same probabilities over all $1,000$ classes (\ie, logits), but contain arbitrary object. Segmentation models also suffer from large performance drop under adversarial examples. \cite{Arnab208} investigated the sensitivity of segmentation methods to adversarial perturbations, and observed that residual connection and multi-scale processing contribute to robustness improvement. \cite{guo2019degraded} found that image degradation has a great impact on segmentation models.

The closest work to ours is due to \cite{wang2008maximum}, who practiced ``analysis by synthesis'' in the field of computational vision, and proposed the maximum differentiation competition method for efficiently comparing computational models of perceptual quantities. In the context of image quality assessment \citep{wang2006modern}, given two image quality models, they first synthesized a pair of images that maximize (and minimize) the predictions of one model while fixing the other. This process is repeated by reversing the roles of the two models. The synthesized images are the most informative in revealing their relative advantages and disadvantages. Nevertheless, the image synthesis process may be computationally expensive, and the synthesized images may be highly unnatural and of less practical relevance. \cite{ma2018group} alleviated these issues by restricting the sampling process from the space of all possible images to a large-scale image subdomain of interest. \cite{wang2020going} extended this idea to compare multiple ImageNet classifiers, where the discrepancy between two classifiers is measured by a weighted hop distance over the WordNet hierarchy \citep{miller1998wordnet}. However, the above methods do not take into full consideration the diversity of the selected samples, which is crucial for fully characterizing erroneous corner-case behaviors of the competing models.

We conclude this section by mentioning that similar ideas have emerged in software engineering, in the form of differential testing \citep{mckeeman1998}. It is a classic idea to find bug-exposing test samples, by providing the same tests to multiple software implementations of the same function. If some implementation runs differently from the others, it is likely to be buggy. Inspired by differential testing, \cite{pei2017deepxplore} developed DeepXplore, an automated whitebox testing of deep driving systems on synthetic weather conditions.

\begin{figure*}[!ht]
\centering
\subfigure[]{
\begin{minipage}[]{0.48\linewidth}
\includegraphics[width=0.31\linewidth]{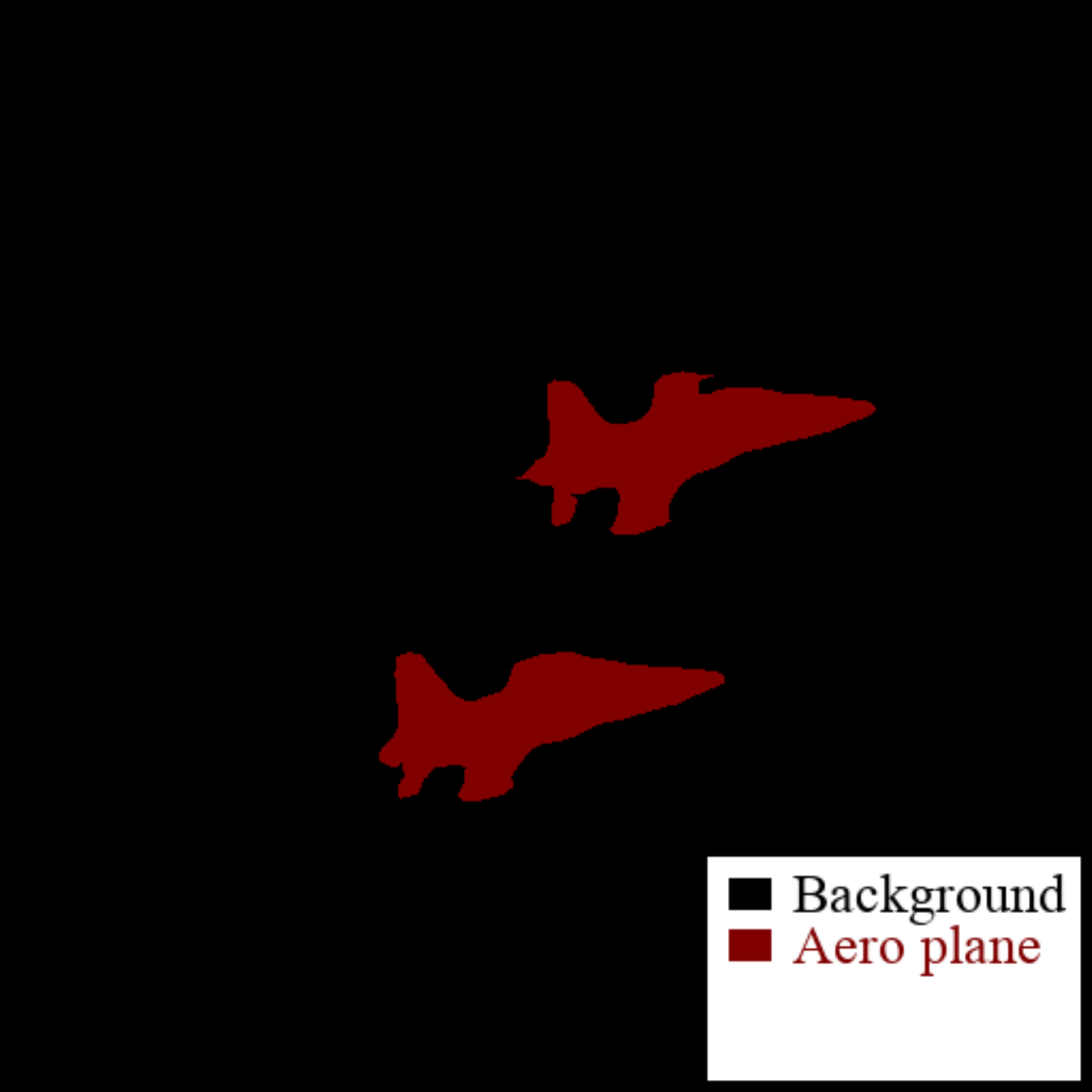}
\includegraphics[width=0.31\linewidth]{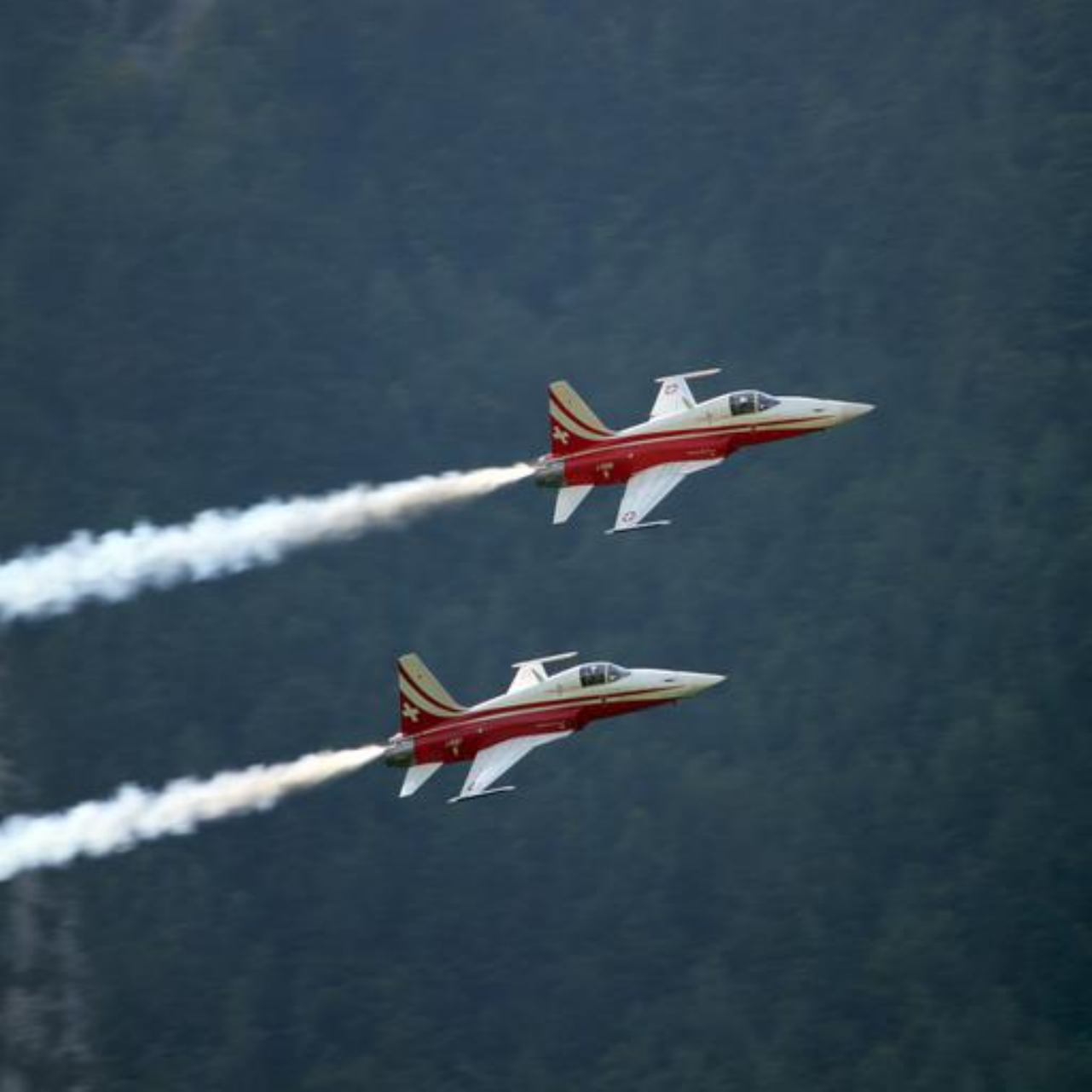}\vspace{0.51pt}
\includegraphics[width=0.31\linewidth]{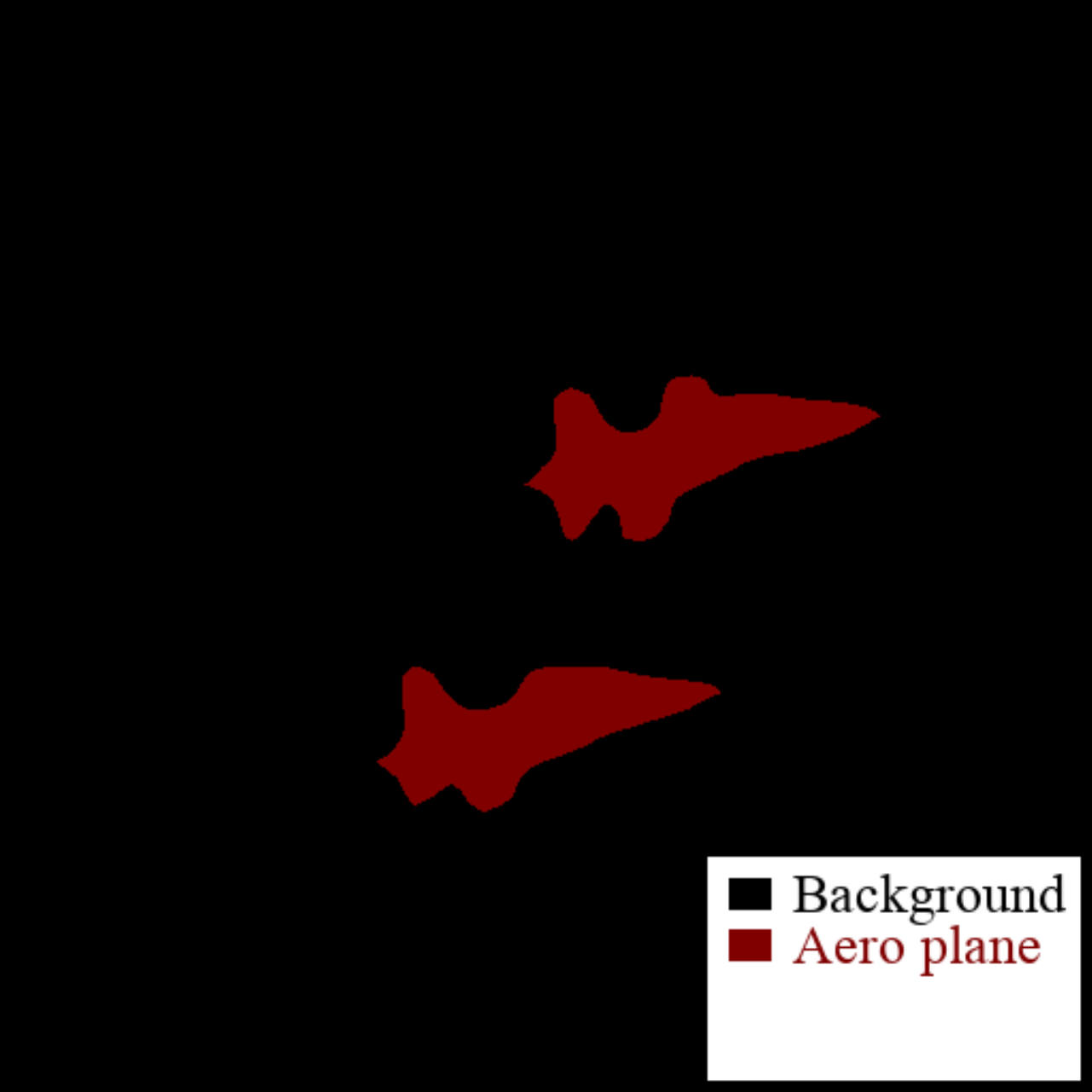}\vspace{0.51pt}
\end{minipage}}
\subfigure[]{
\begin{minipage}[]{0.48\linewidth}
\includegraphics[width=0.31\linewidth]{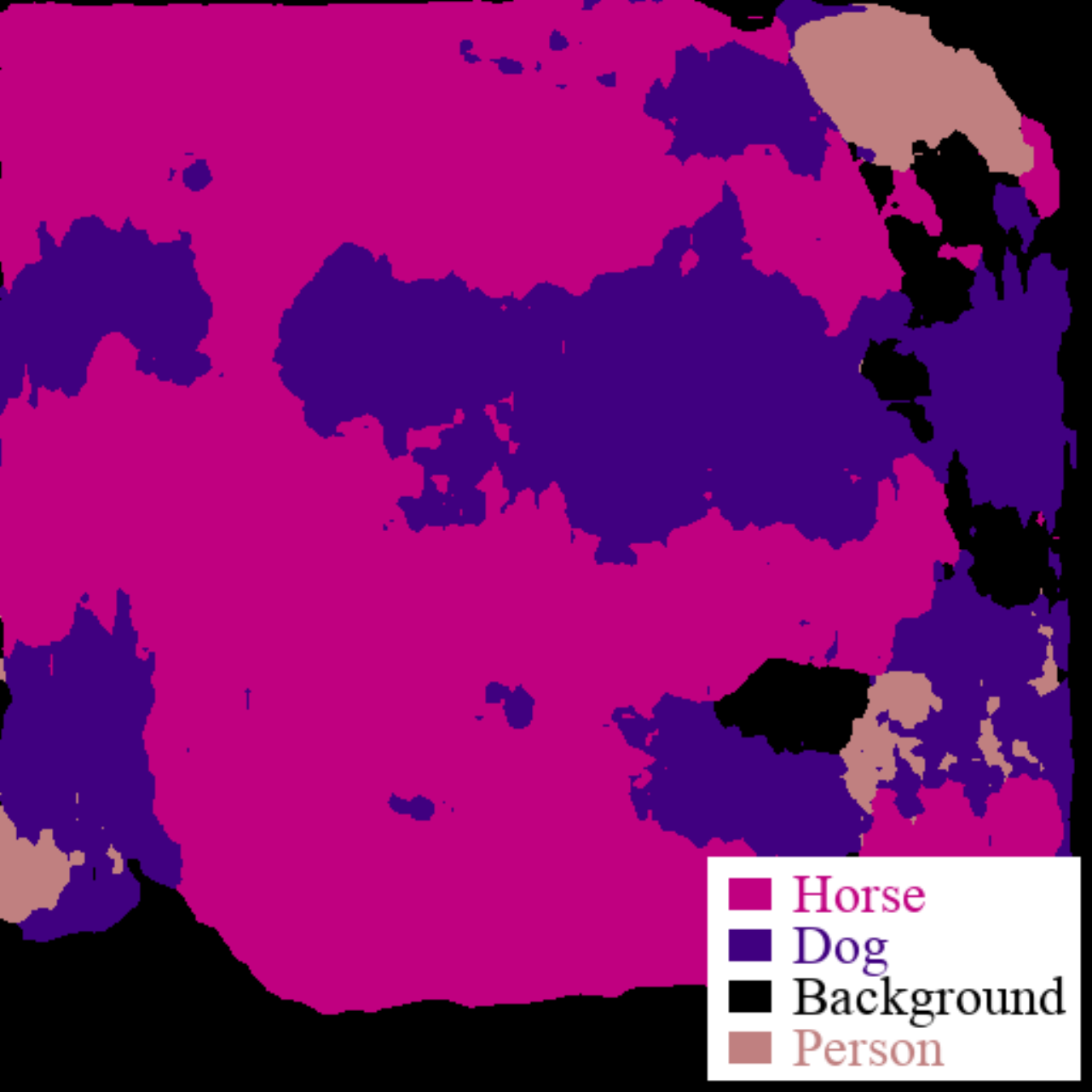}\vspace{0.51pt}
\includegraphics[width=0.31\linewidth]{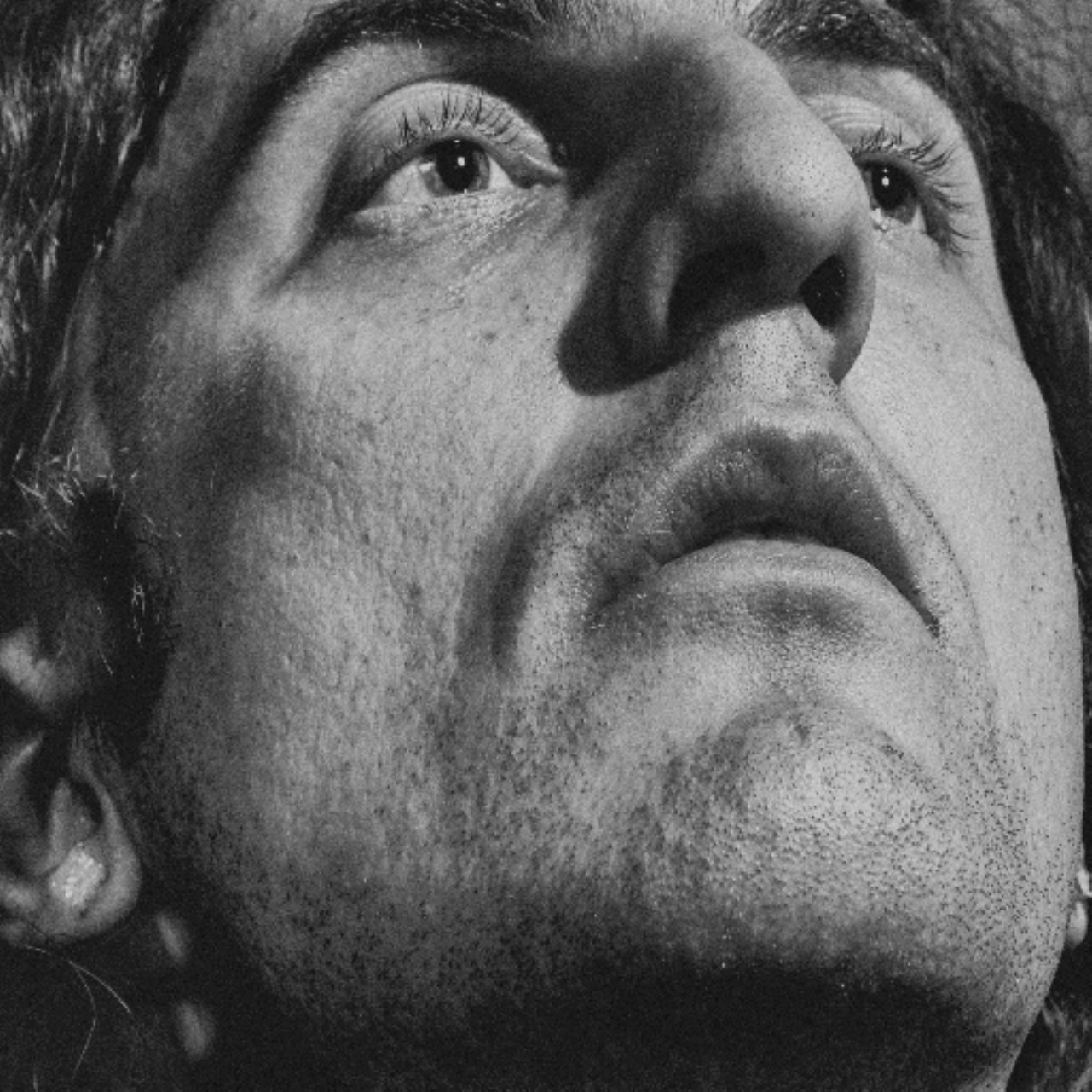}\vspace{0.51pt}
\includegraphics[width=0.31\linewidth]{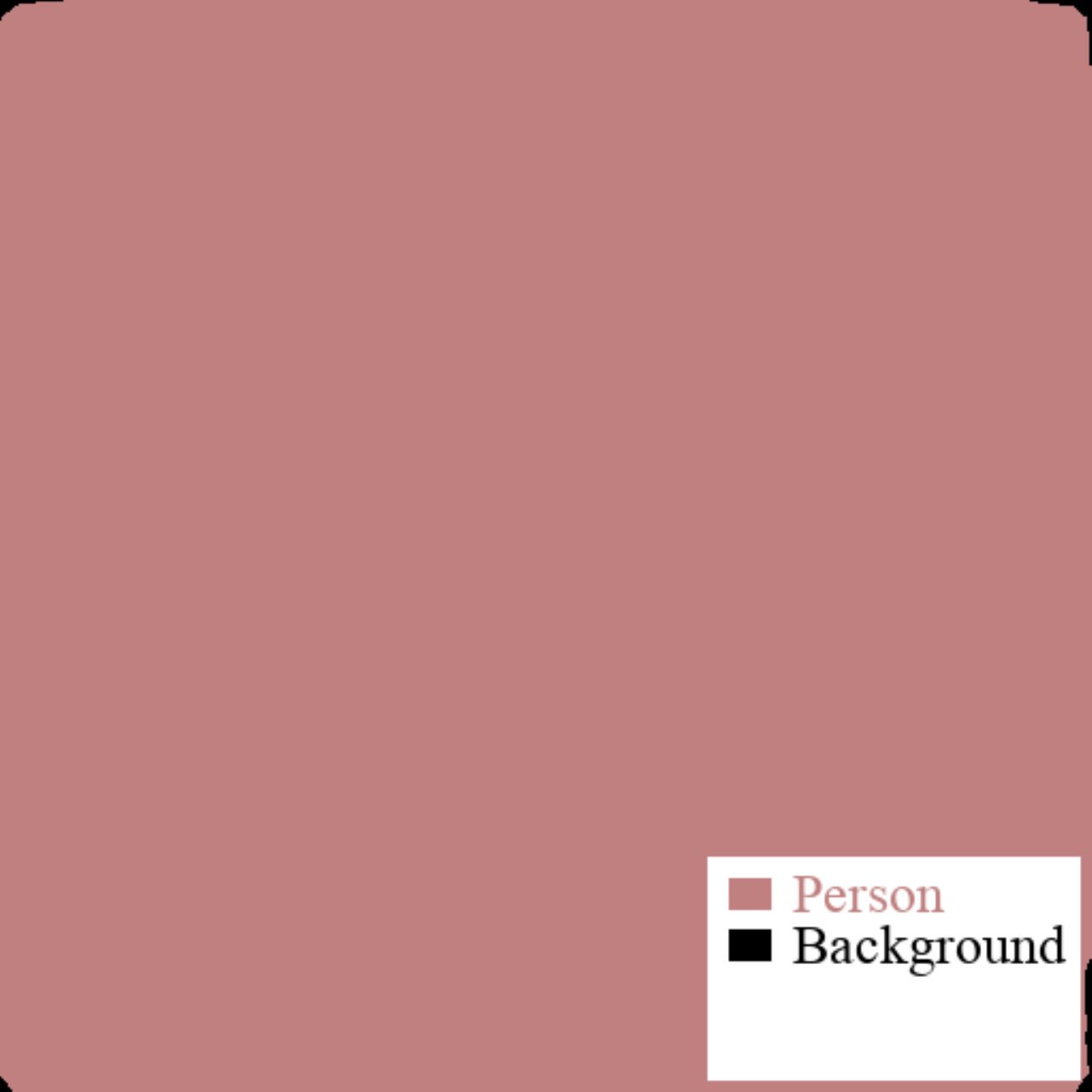}\vspace{0.51pt}
\end{minipage}}\\
\subfigure[]{
\begin{minipage}[]{0.48\linewidth}
\includegraphics[width=0.31\linewidth]{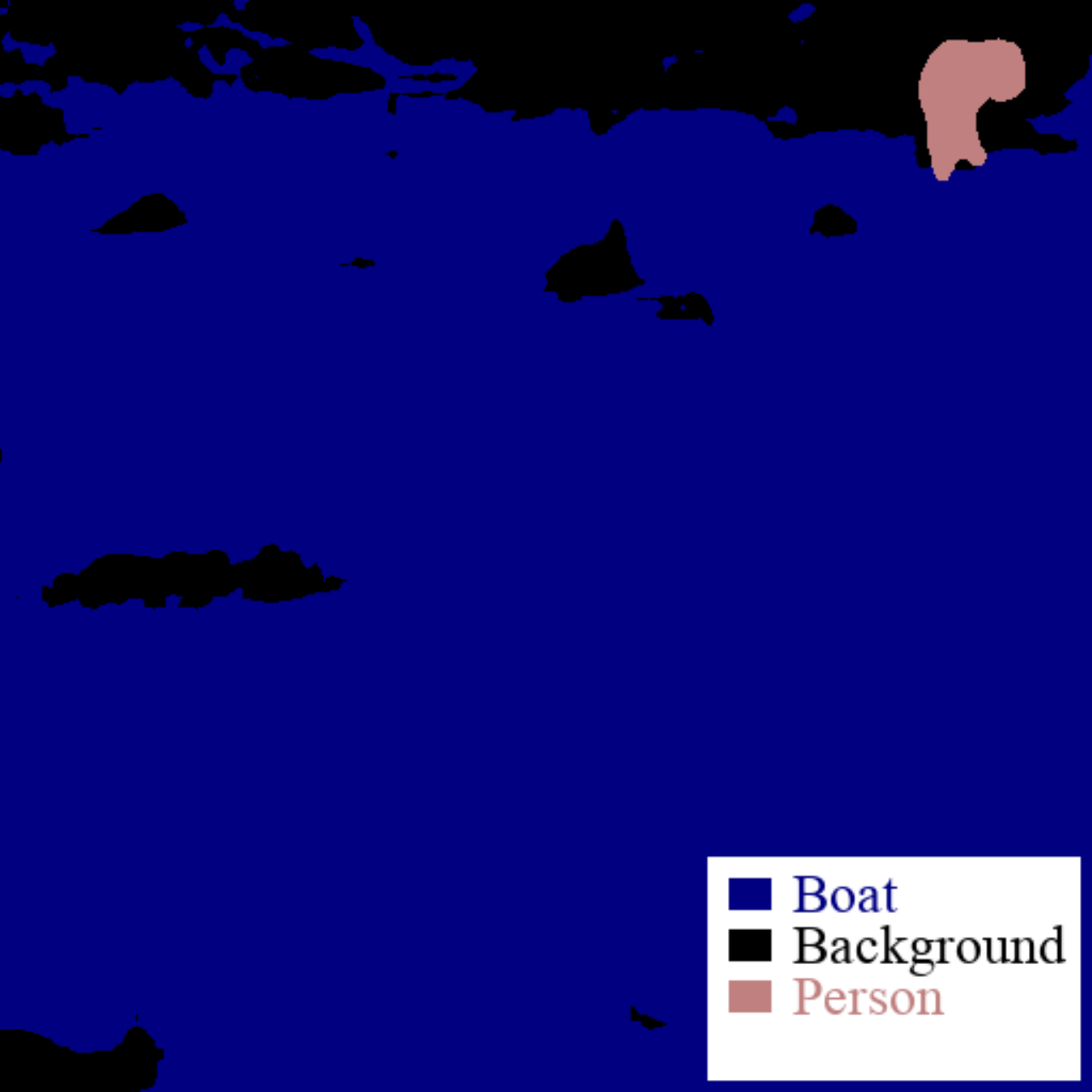}\vspace{0.51pt}
\includegraphics[width=0.31\linewidth]{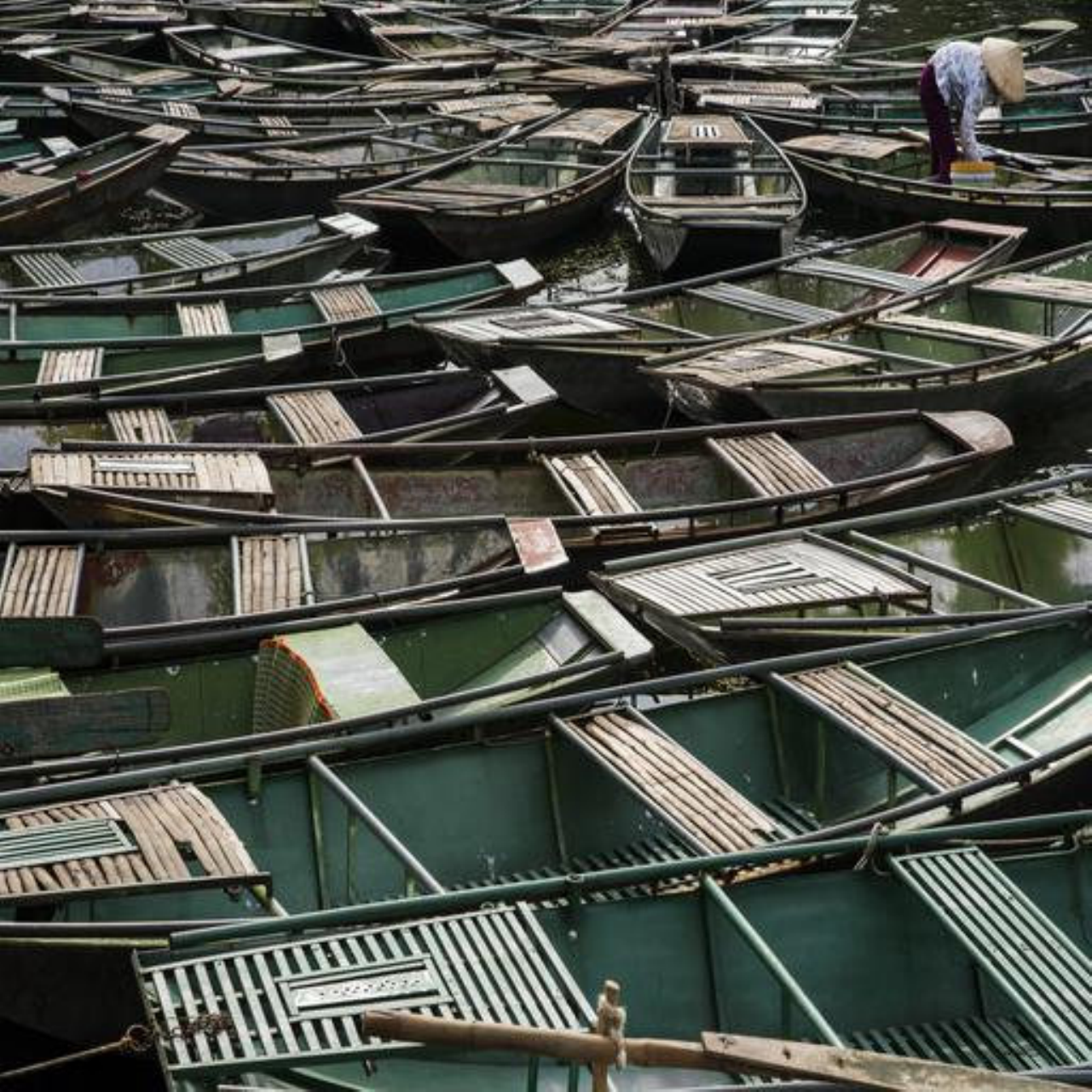}\vspace{0.51pt}
\includegraphics[width=0.31\linewidth]{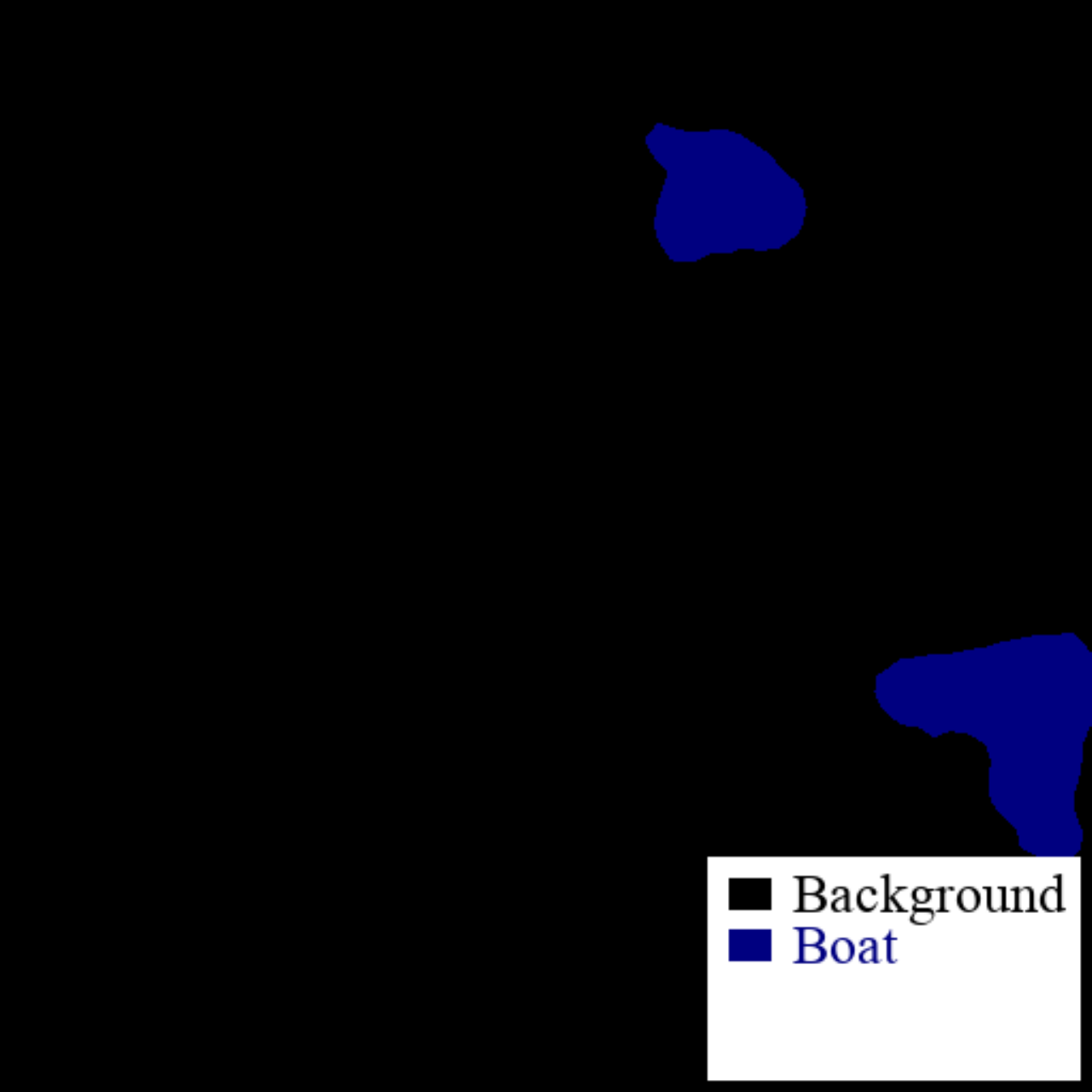}\vspace{0.51pt}
\end{minipage}}
\subfigure[]{
\begin{minipage}[]{0.48\linewidth}
\includegraphics[width=0.31\linewidth]{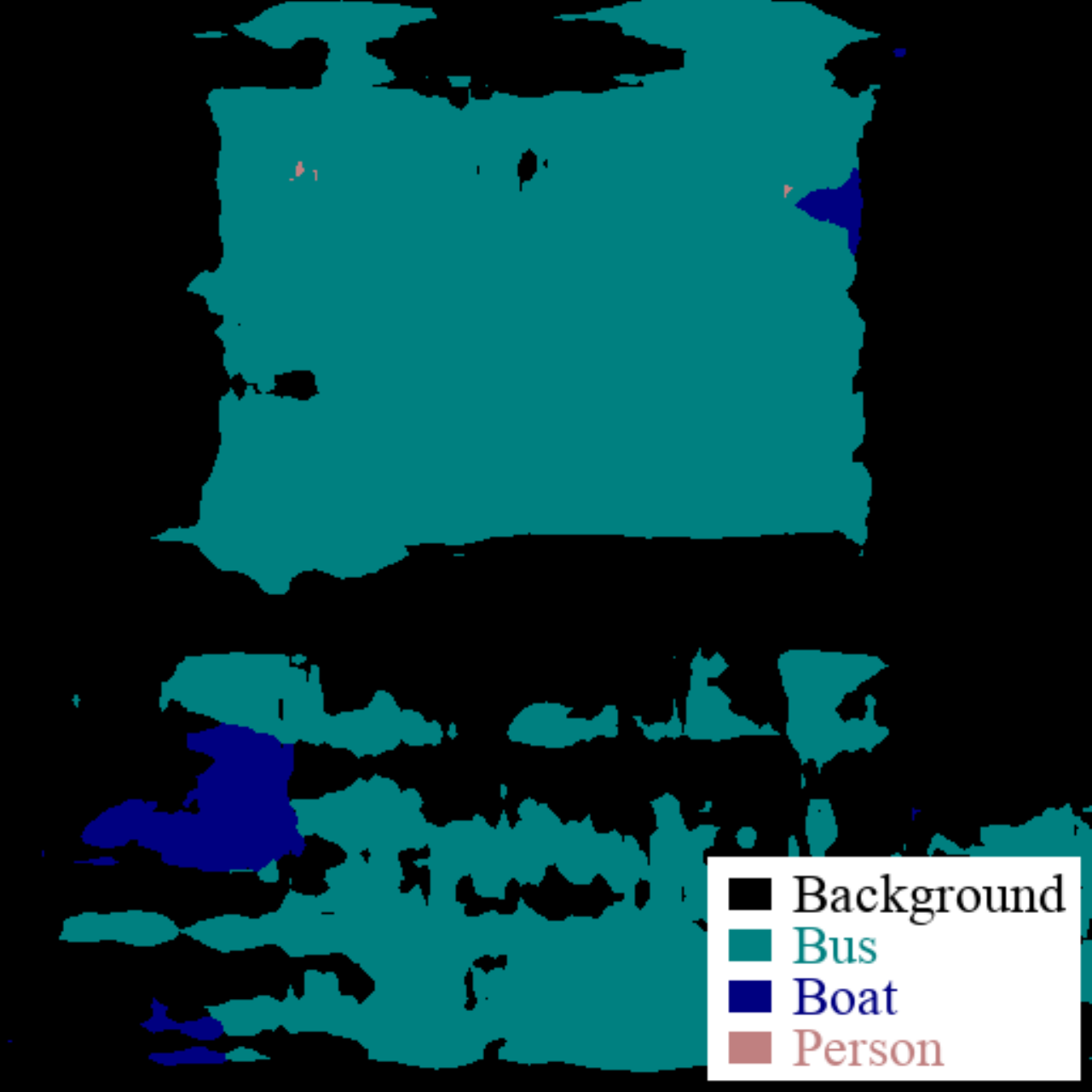}\vspace{0.51pt}
\includegraphics[width=0.31\linewidth]{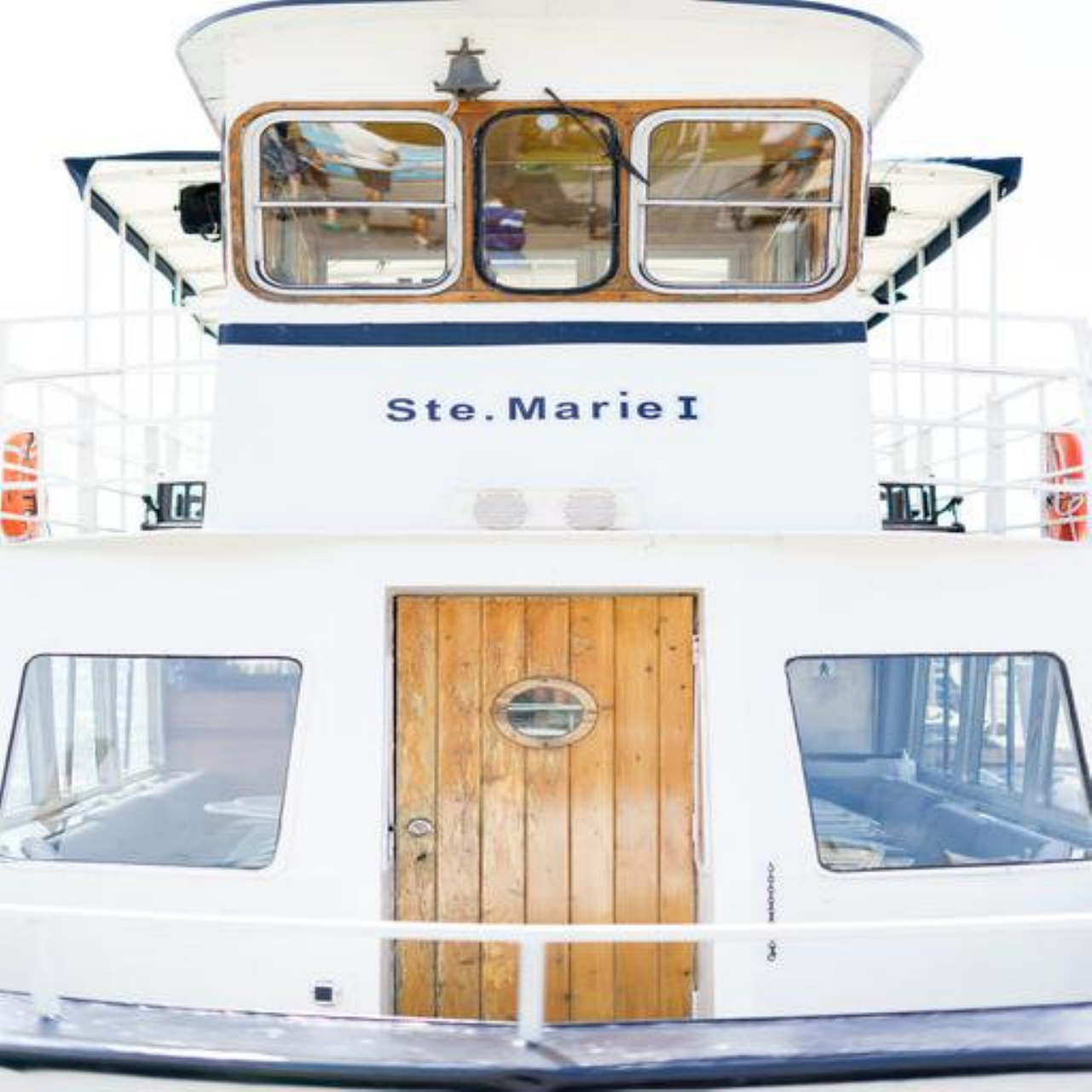}\vspace{0.51pt}
\includegraphics[width=0.31\linewidth]{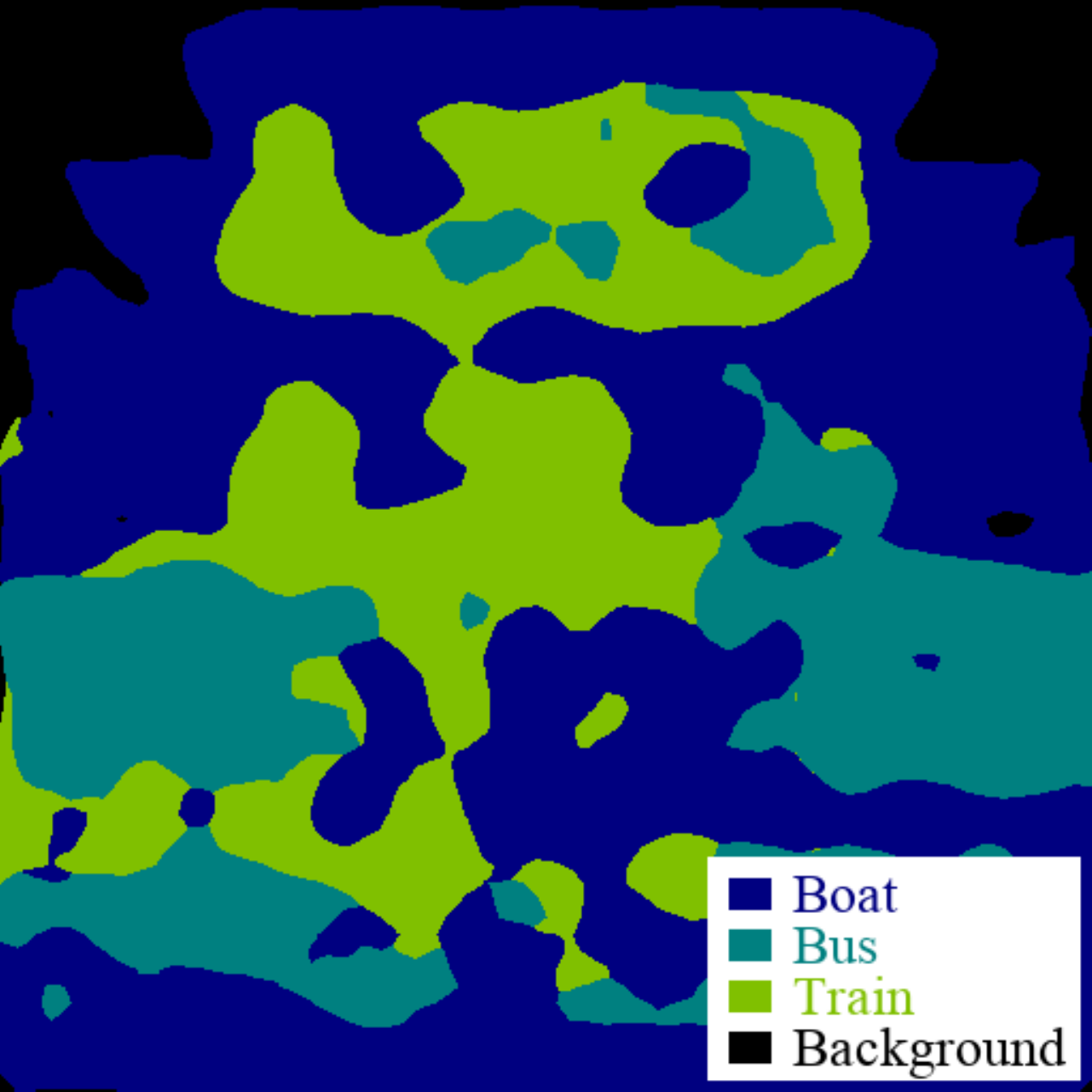}\vspace{0.51pt}
\end{minipage}}
\caption{An illustration of three plausible outcomes of the MAD competition between two segmentation methods (DeepLabv3+~\citep{Chen2018v3+} and EMANet~\citep{Lixia2019}). (a) Both methods give satisfactory segmentation results. (b) and (c) One method makes significantly better prediction than the other. (d) Both methods make poor predictions.}
\label{fig:cases}
\end{figure*}


\section{Proposed Method}
\label{sec_propo}


In this section, we investigate the generalizability of semantic segmentation methods by efficiently exposing their failures using the MAD competition~\citep{wang2008maximum, ma2018group}. Rather than creating another fixed  test set with human annotations, we automatically sample  a small set of algorithm-dependent images by maximizing the discrepancy between the methods, as measured by common metrics in semantic segmentation (\eg, mean intersection over union (mIoU)). Moreover, we allow the resulting MAD set to cover diverse sparse mistakes by enforcing a set of additional conditions. Subjective testing indicates the relative performance of the segmentation methods in the MAD competition, where a better method is more aggressive in spotting others' failures and more resistant to others' attacks. The pipeline of the proposed method is shown in Fig. \ref{fig:framework}.

\begin{figure*}[!ht]
\centering
\subfigure[]{
\begin{minipage}[]{0.48\linewidth}
\includegraphics[width=0.31\linewidth]{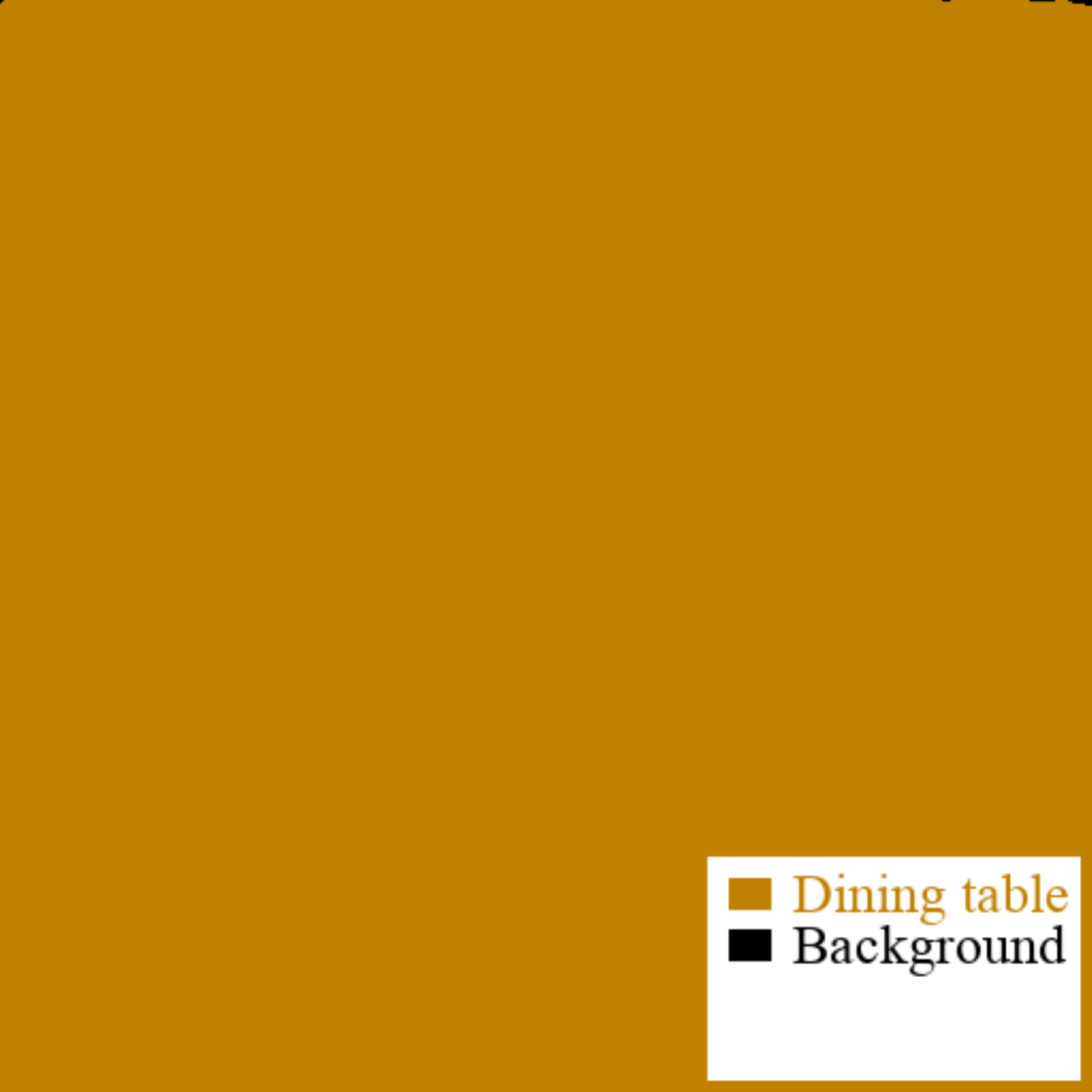}\vspace{0.51pt}
\includegraphics[width=0.31\linewidth]{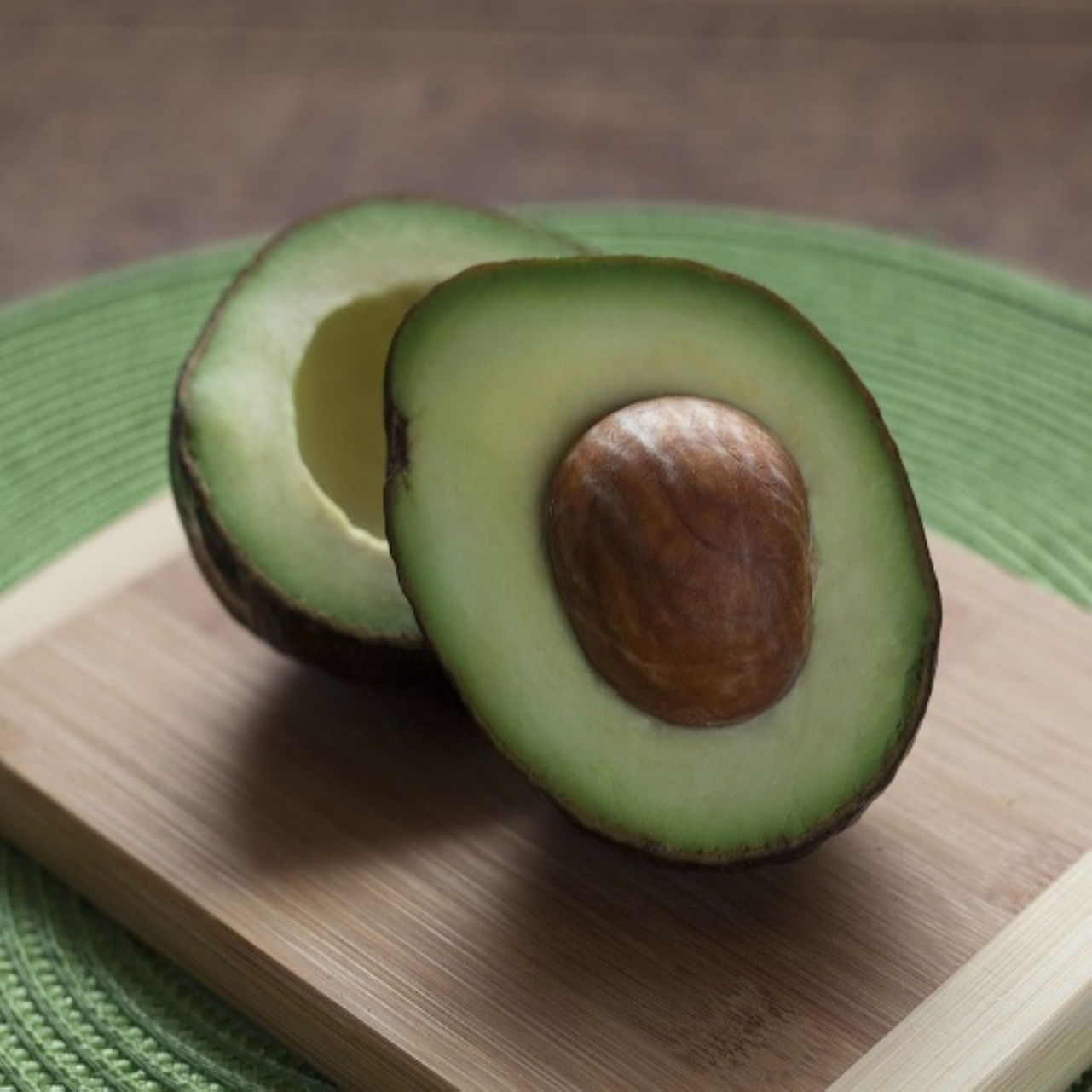}\vspace{0.51pt}
\includegraphics[width=0.31\linewidth]{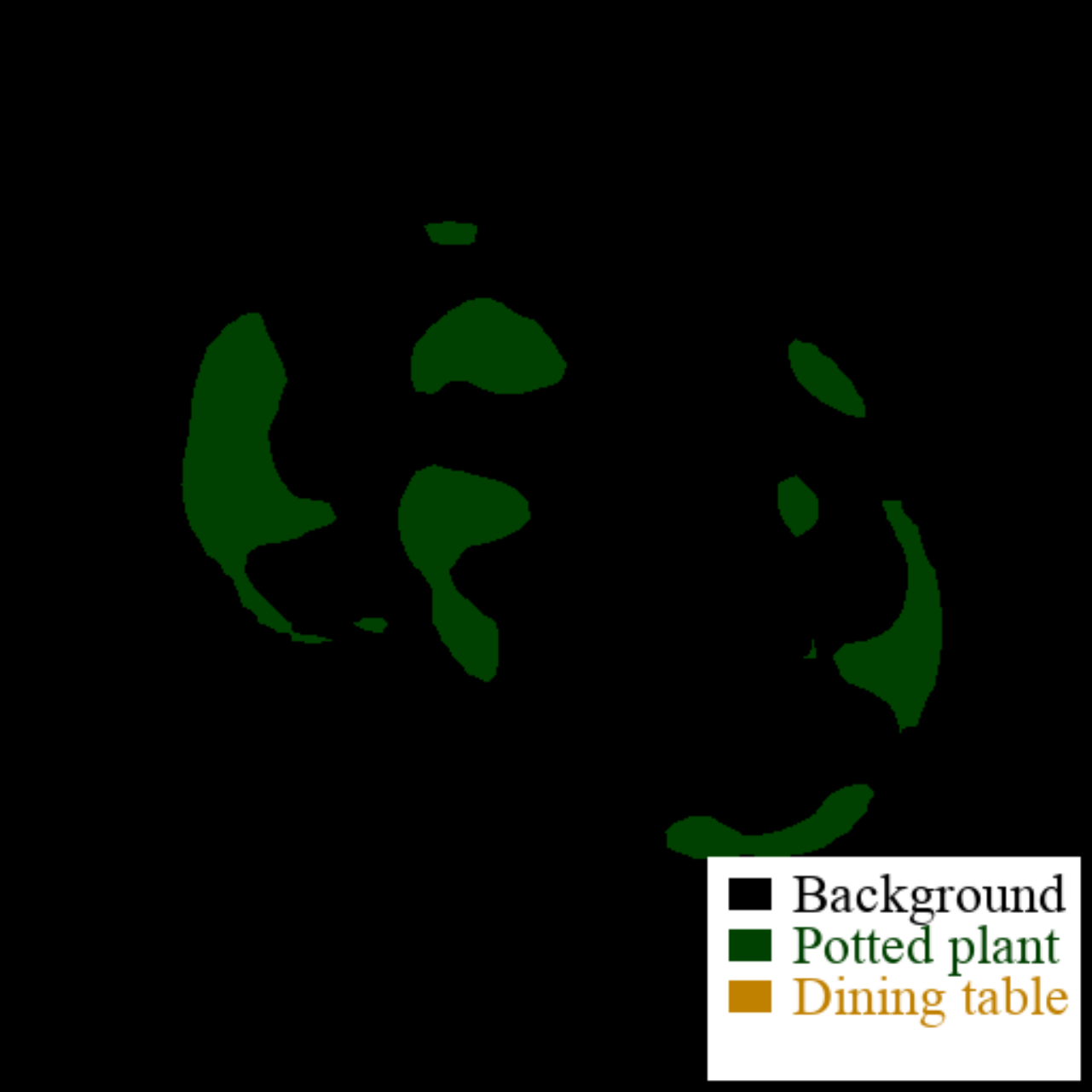}\vspace{0.51pt}\\
\includegraphics[width=0.31\linewidth]{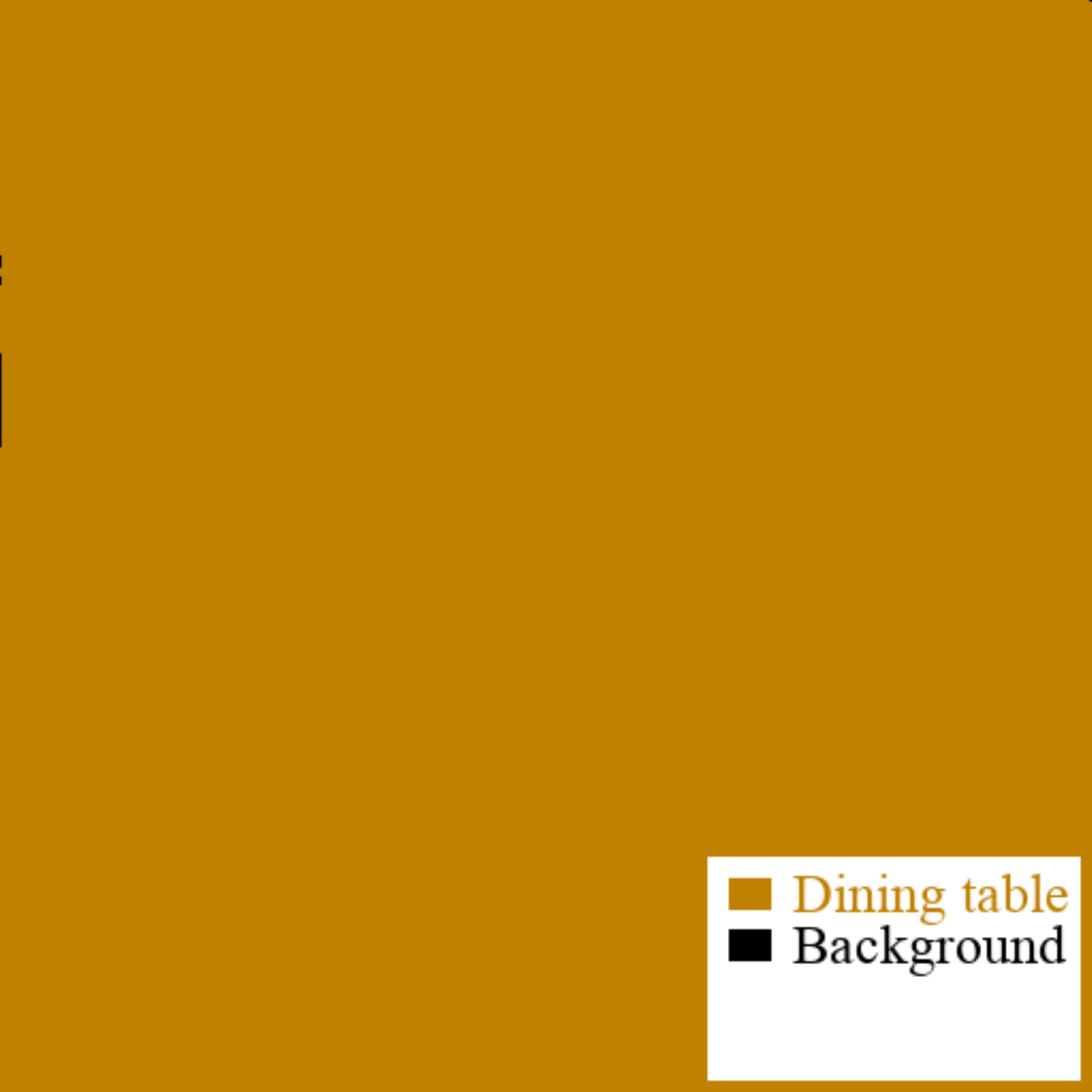}\vspace{0.51pt}
\includegraphics[width=0.31\linewidth]{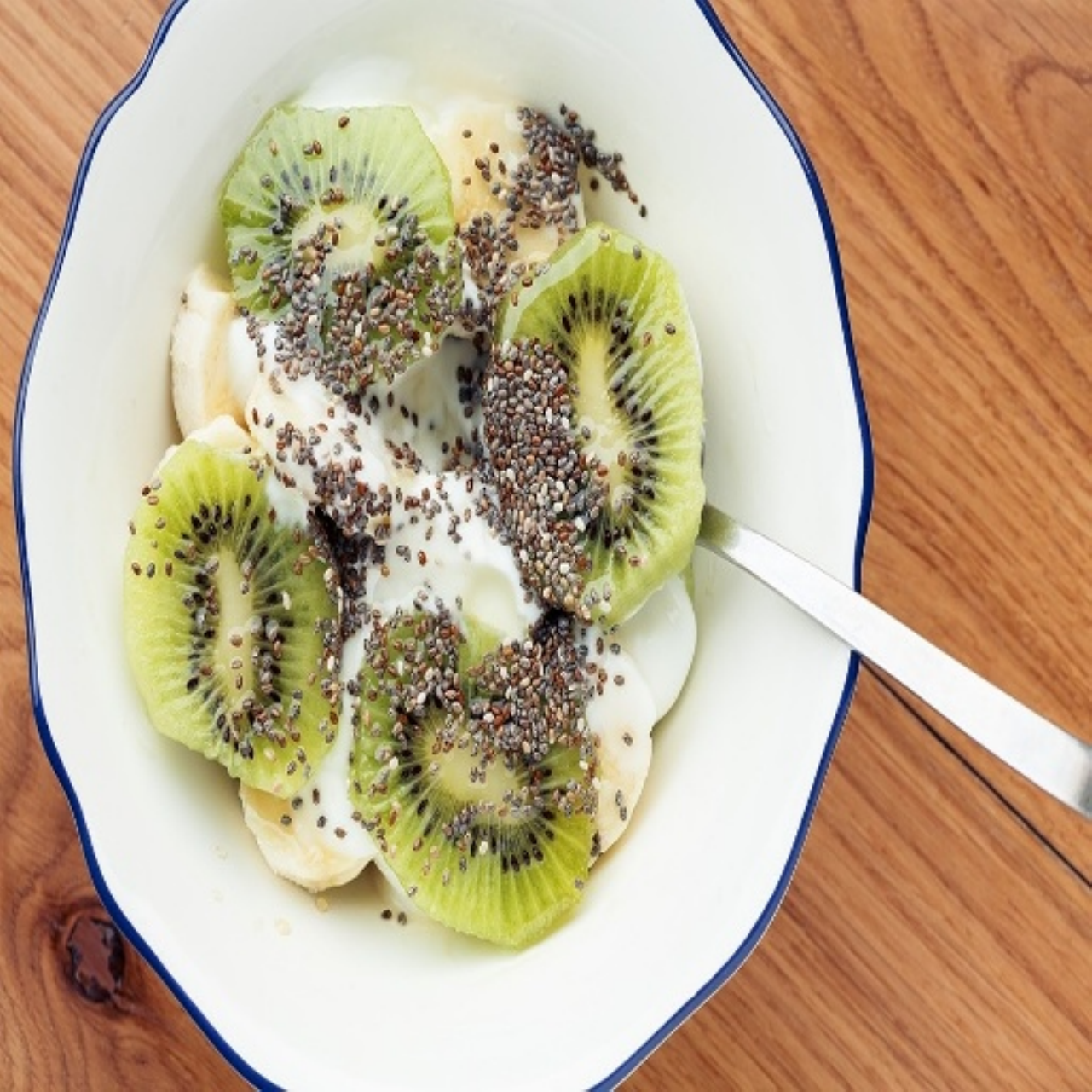}\vspace{0.51pt}
\includegraphics[width=0.31\linewidth]{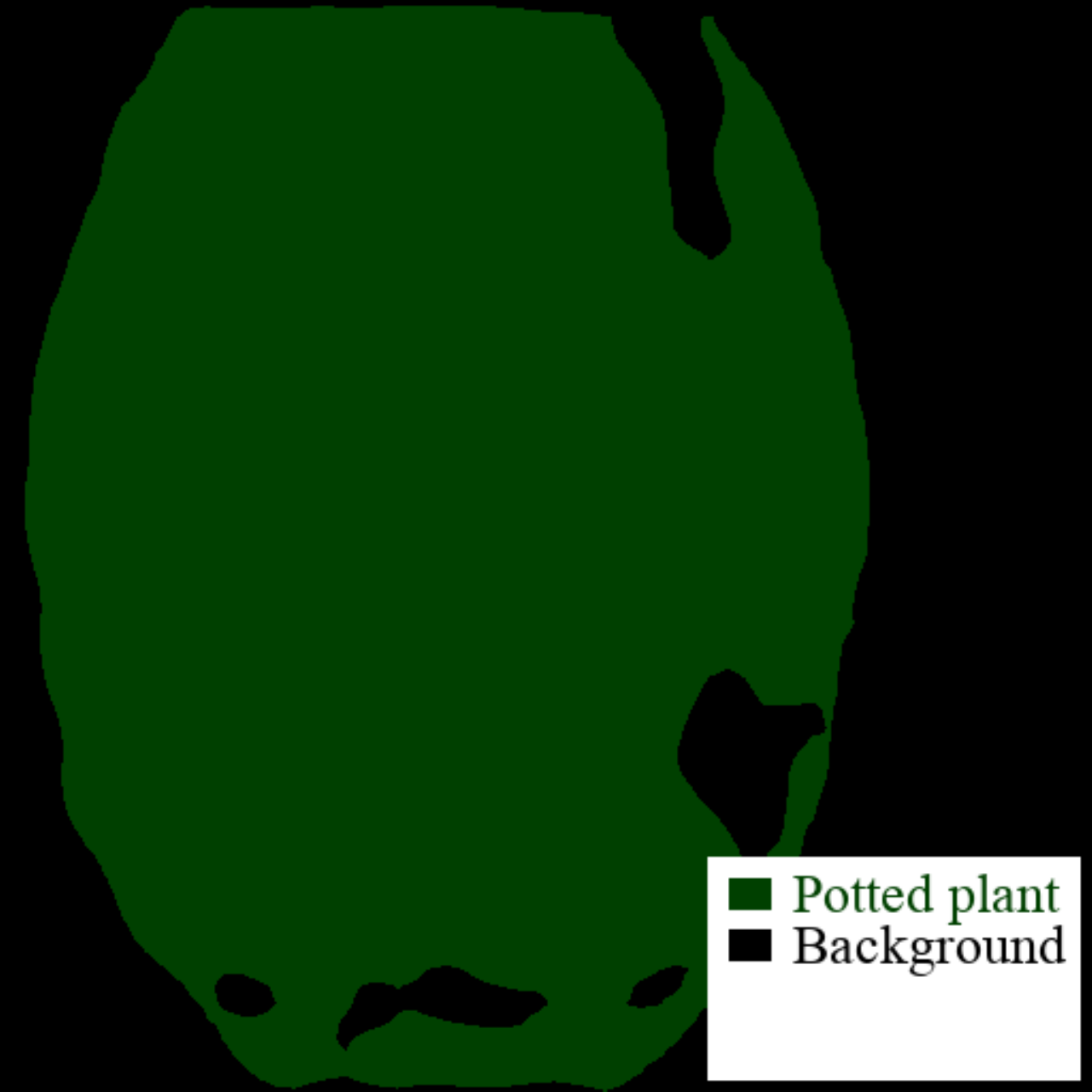}\vspace{0.51pt}
\end{minipage}
\begin{minipage}[]{0.48\linewidth}
\includegraphics[width=0.31\linewidth]{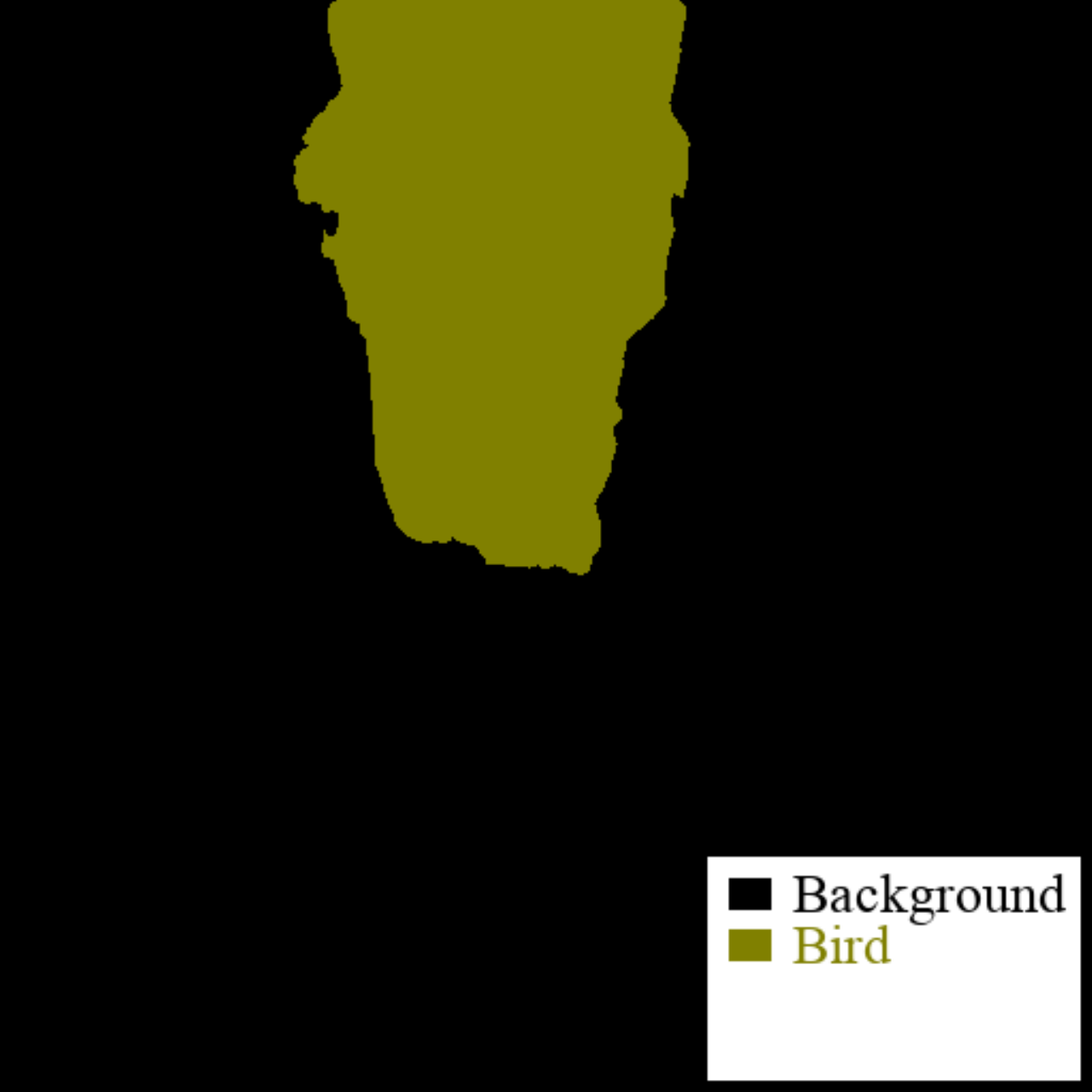}\vspace{0.51pt}
\includegraphics[width=0.31\linewidth]{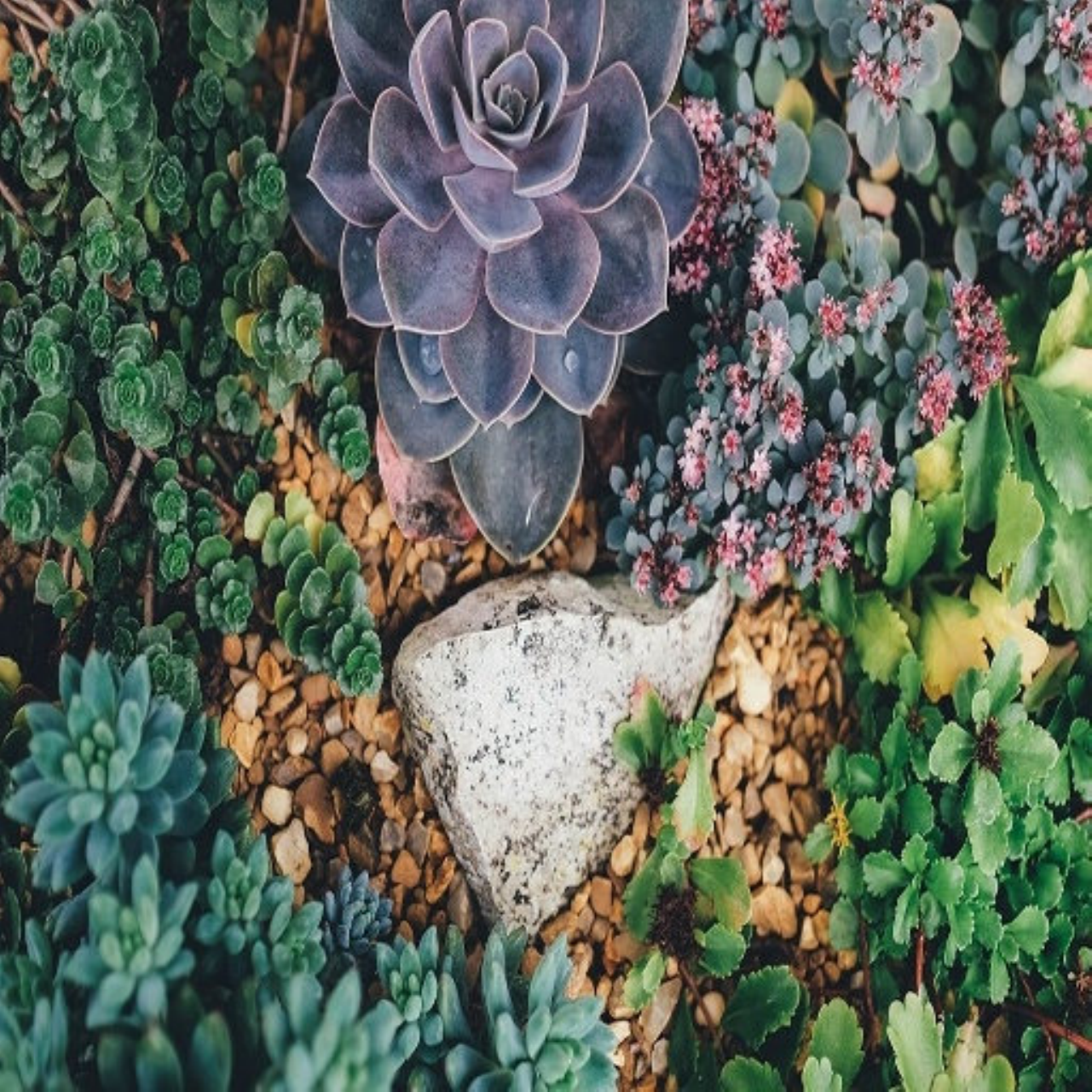}\vspace{0.51pt}
\includegraphics[width=0.31\linewidth]{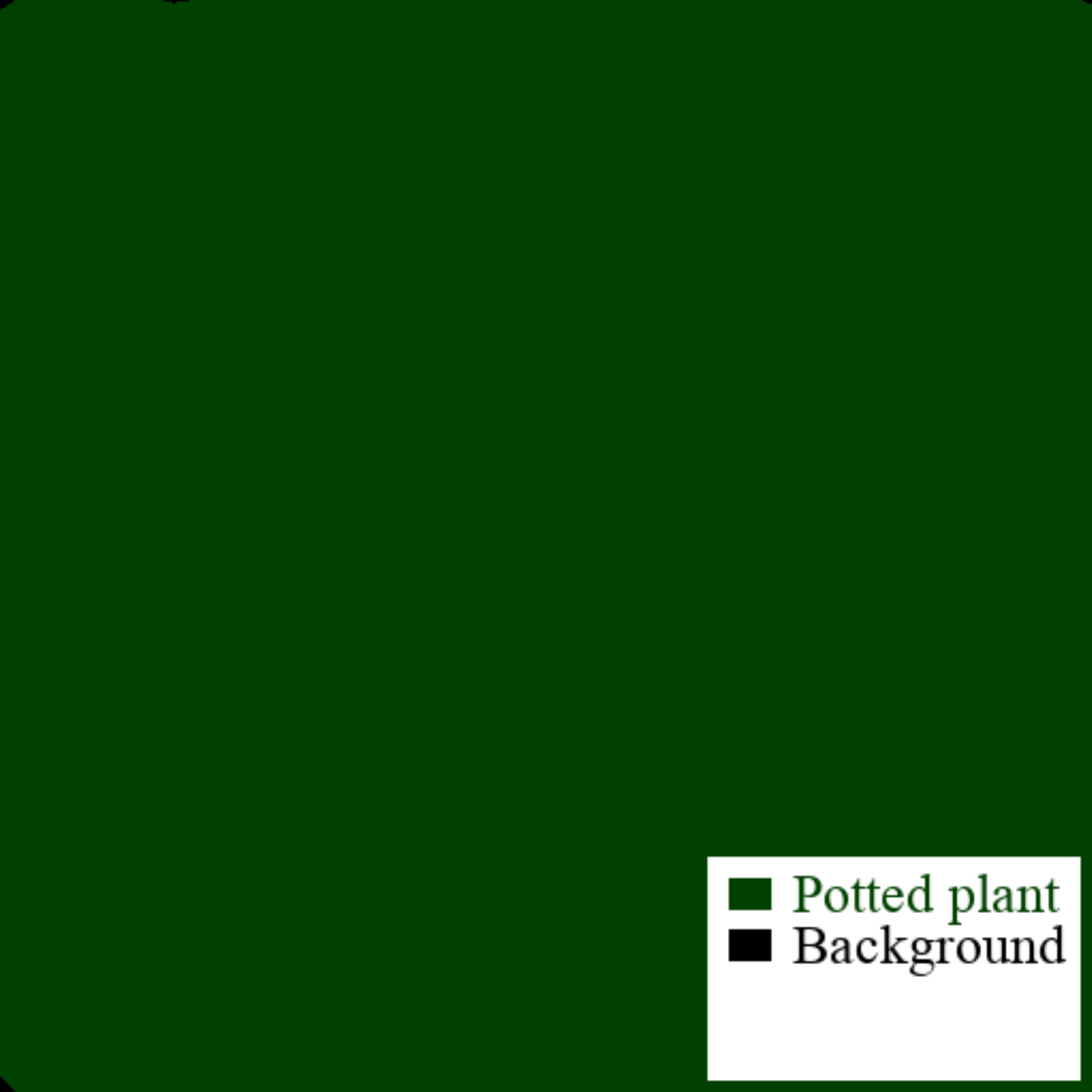}\vspace{0.51pt}\\
\includegraphics[width=0.31\linewidth]{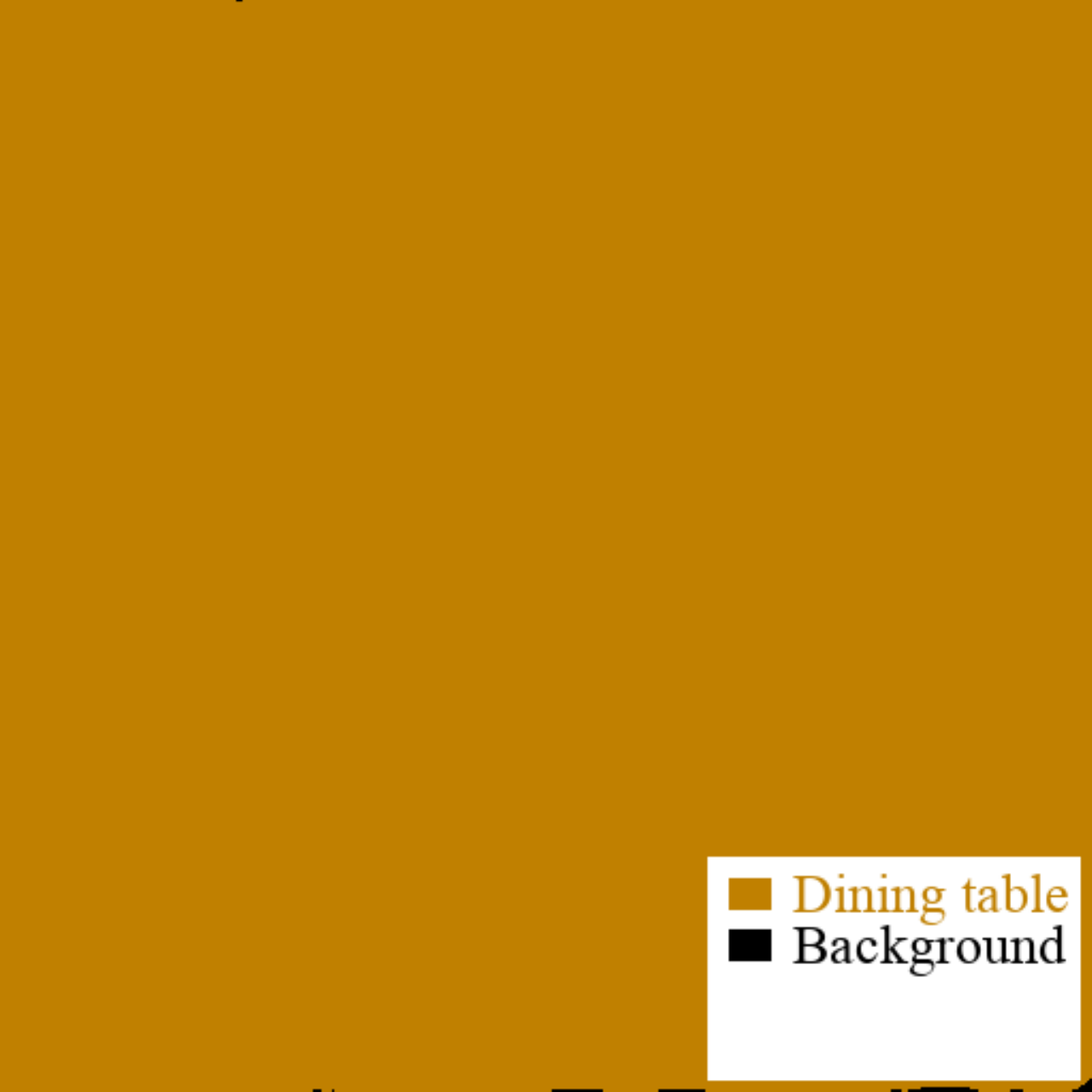}\vspace{0.51pt}
\includegraphics[width=0.31\linewidth]{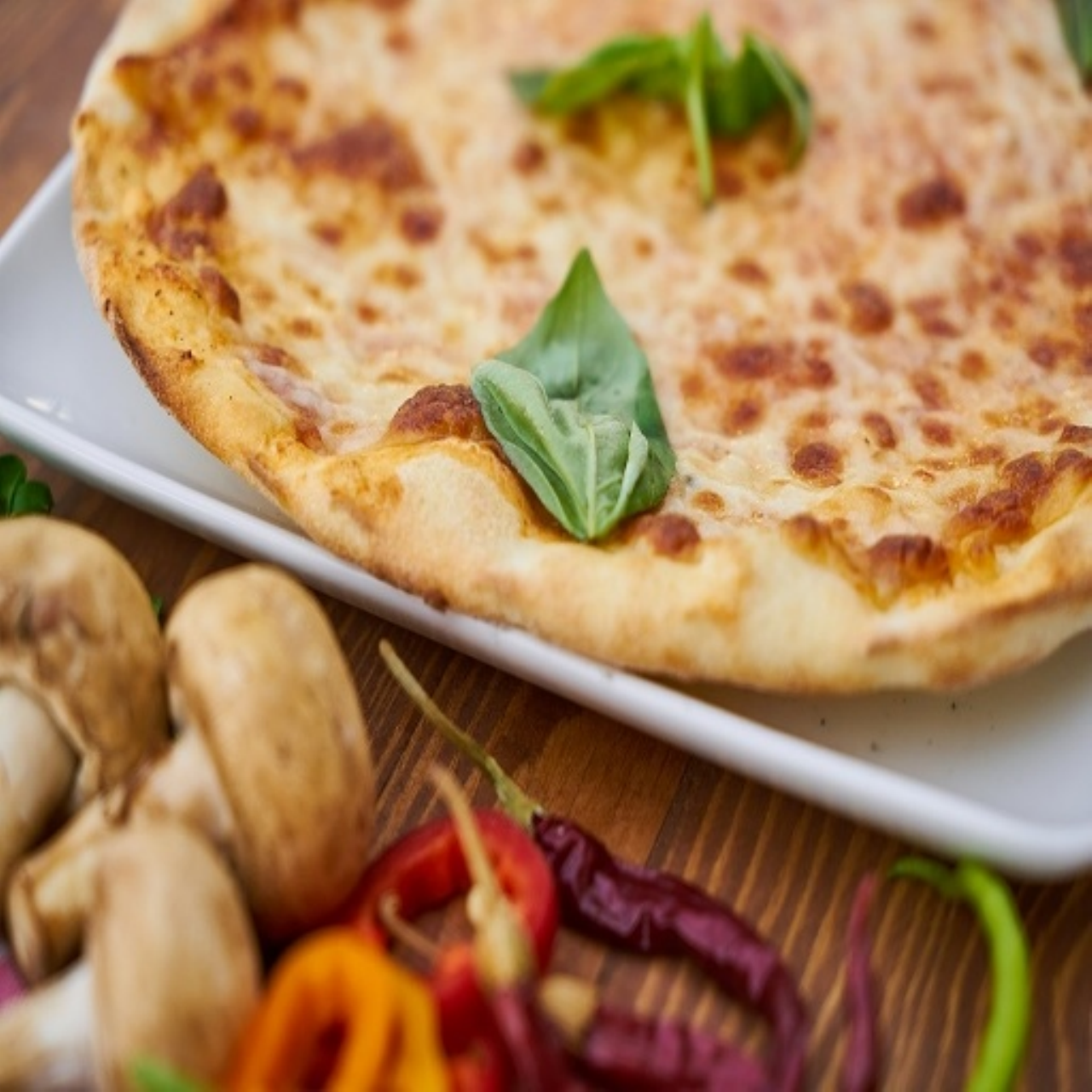}\vspace{0.51pt}
\includegraphics[width=0.31\linewidth]{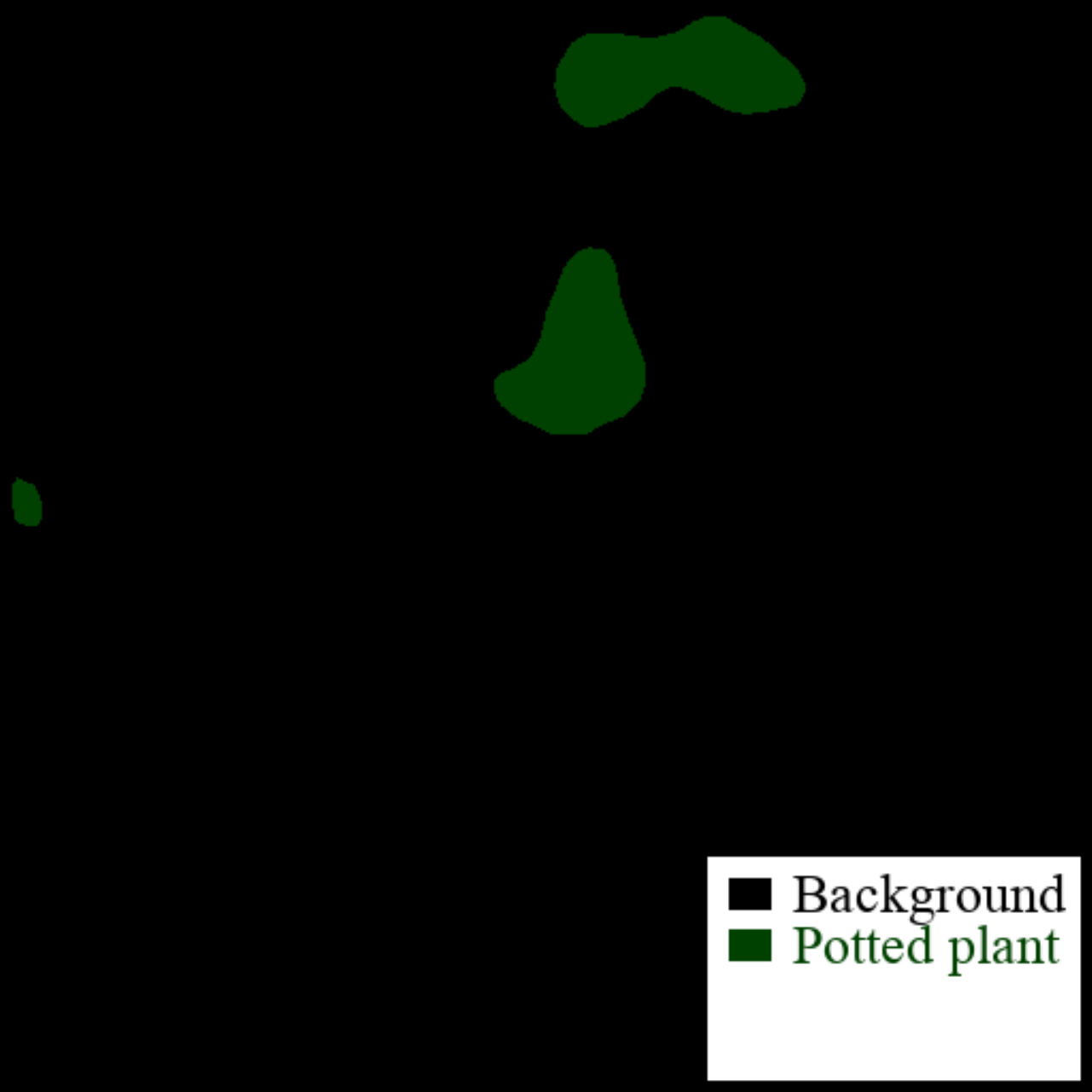}\vspace{0.51pt}
\end{minipage}}
\subfigure[]{
\begin{minipage}[]{0.48\linewidth}
\includegraphics[width=0.31\linewidth]{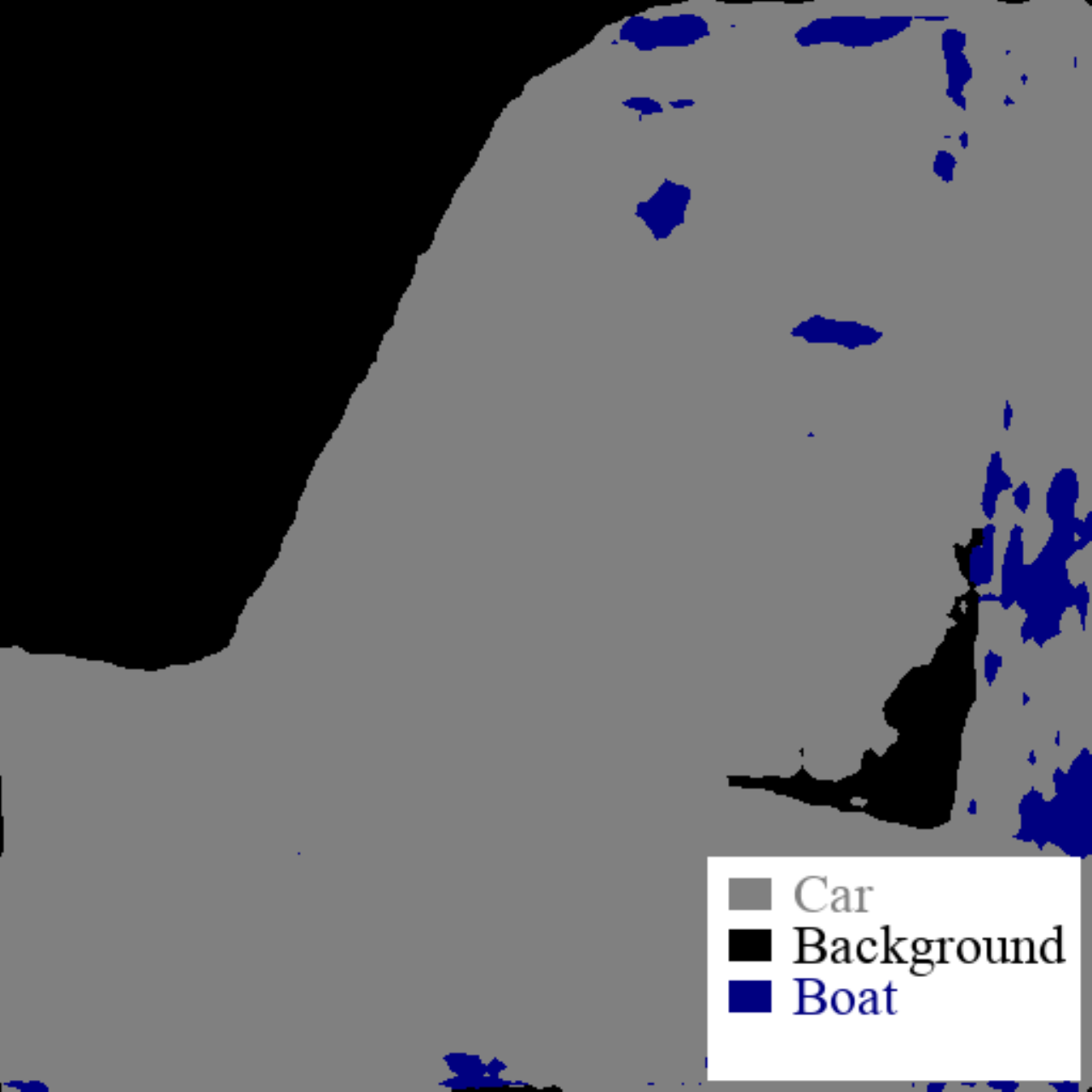}\vspace{0.51pt}
\includegraphics[width=0.31\linewidth]{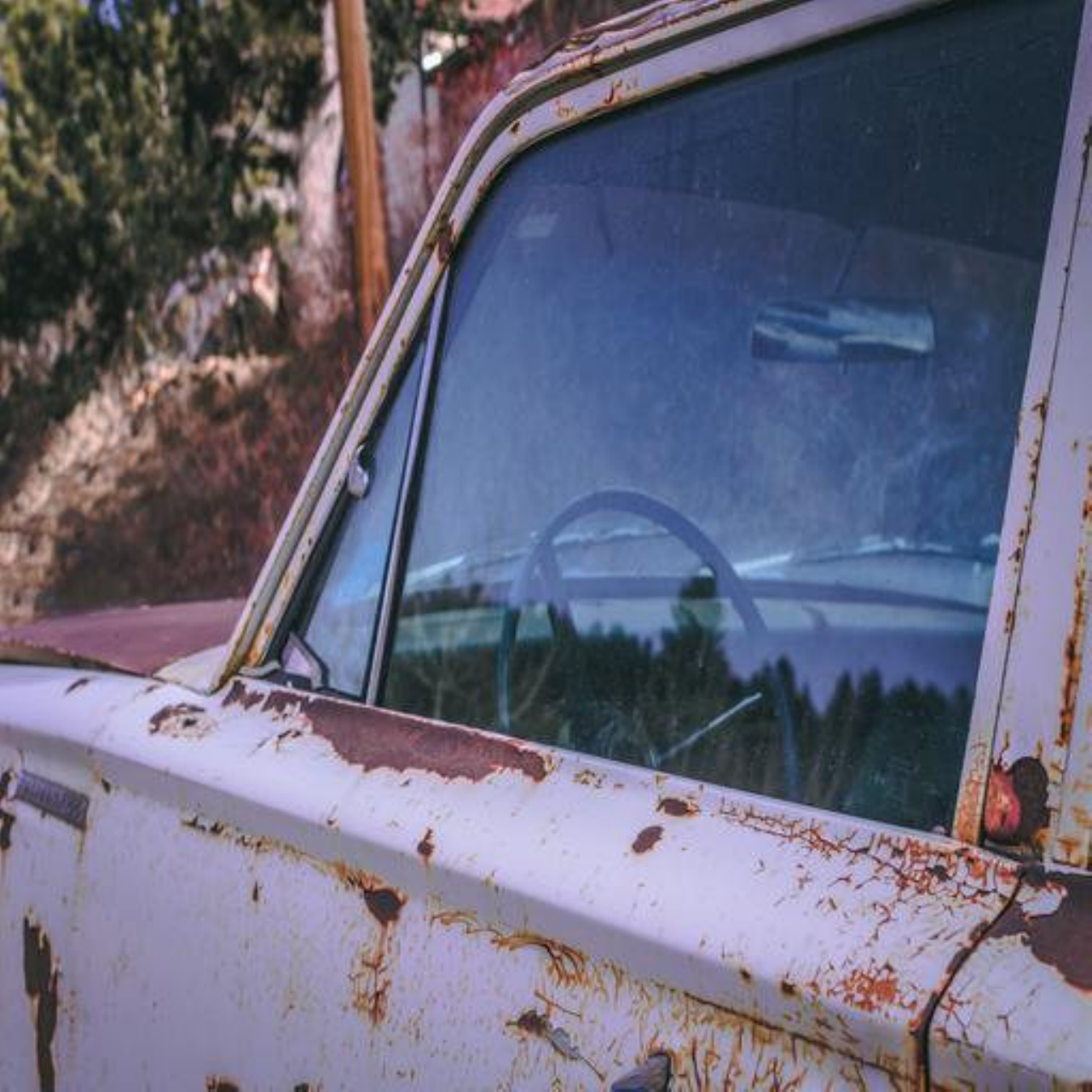}\vspace{0.51pt}
\includegraphics[width=0.31\linewidth]{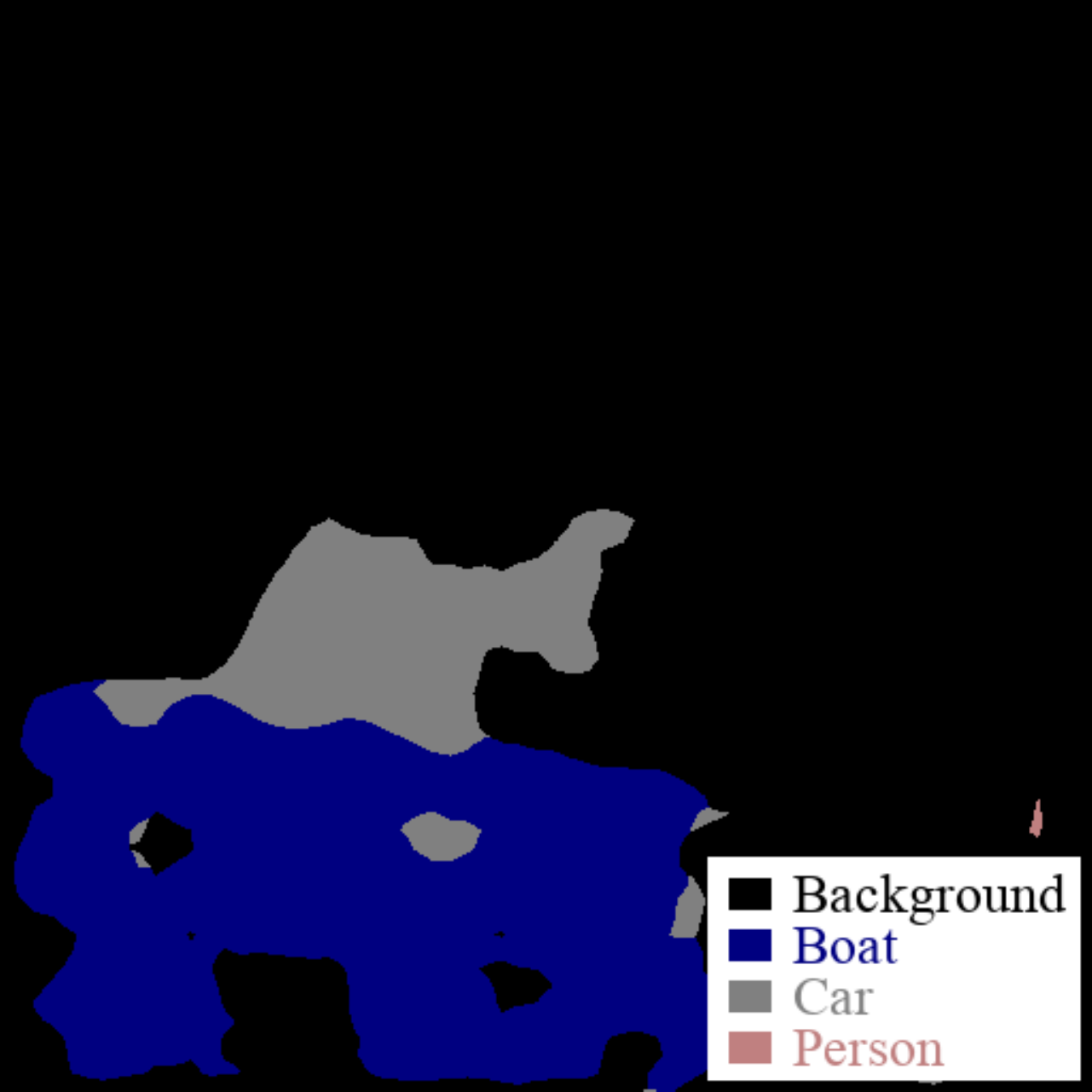}\vspace{0.51pt}\\
\includegraphics[width=0.31\linewidth]{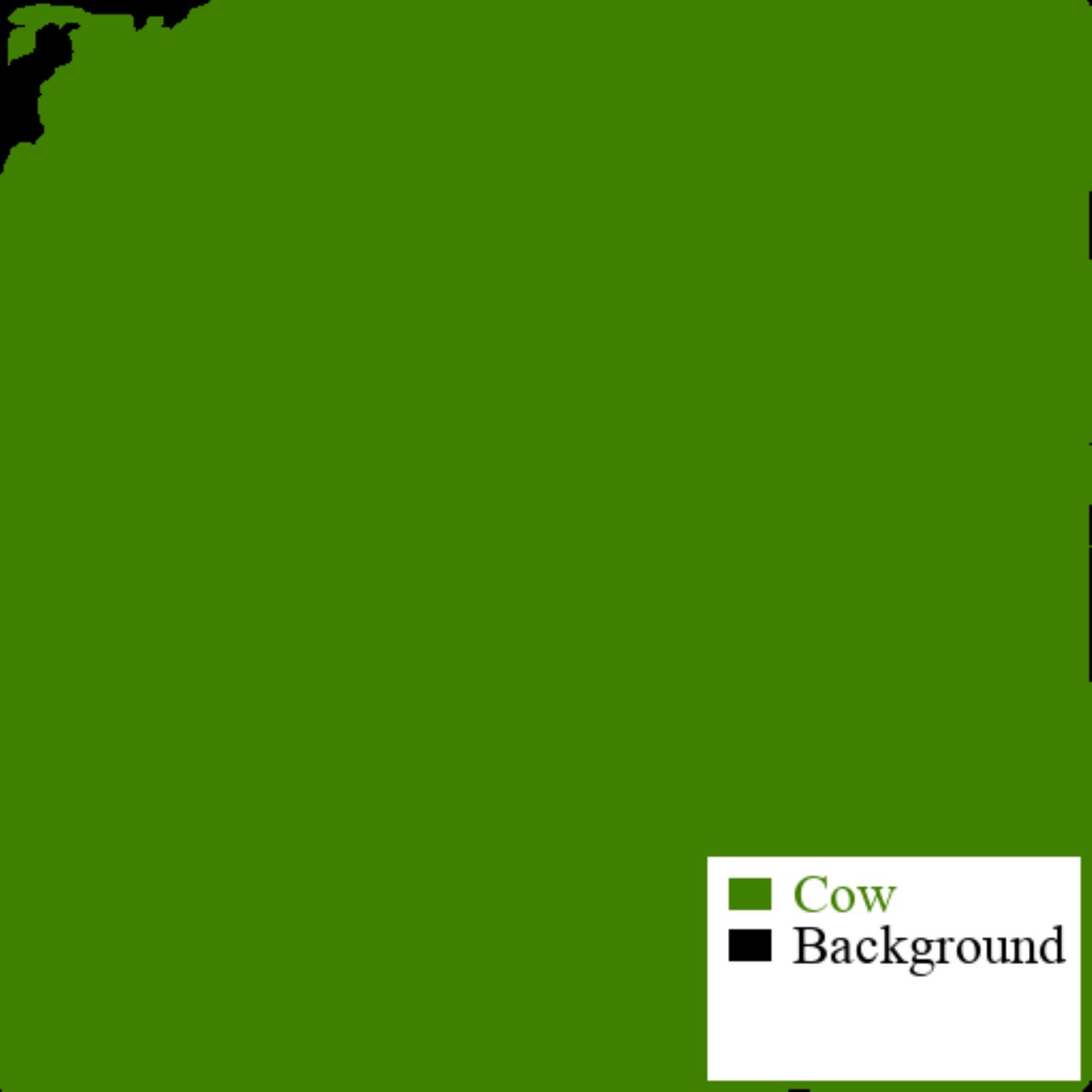}\vspace{0.51pt}
\includegraphics[width=0.31\linewidth]{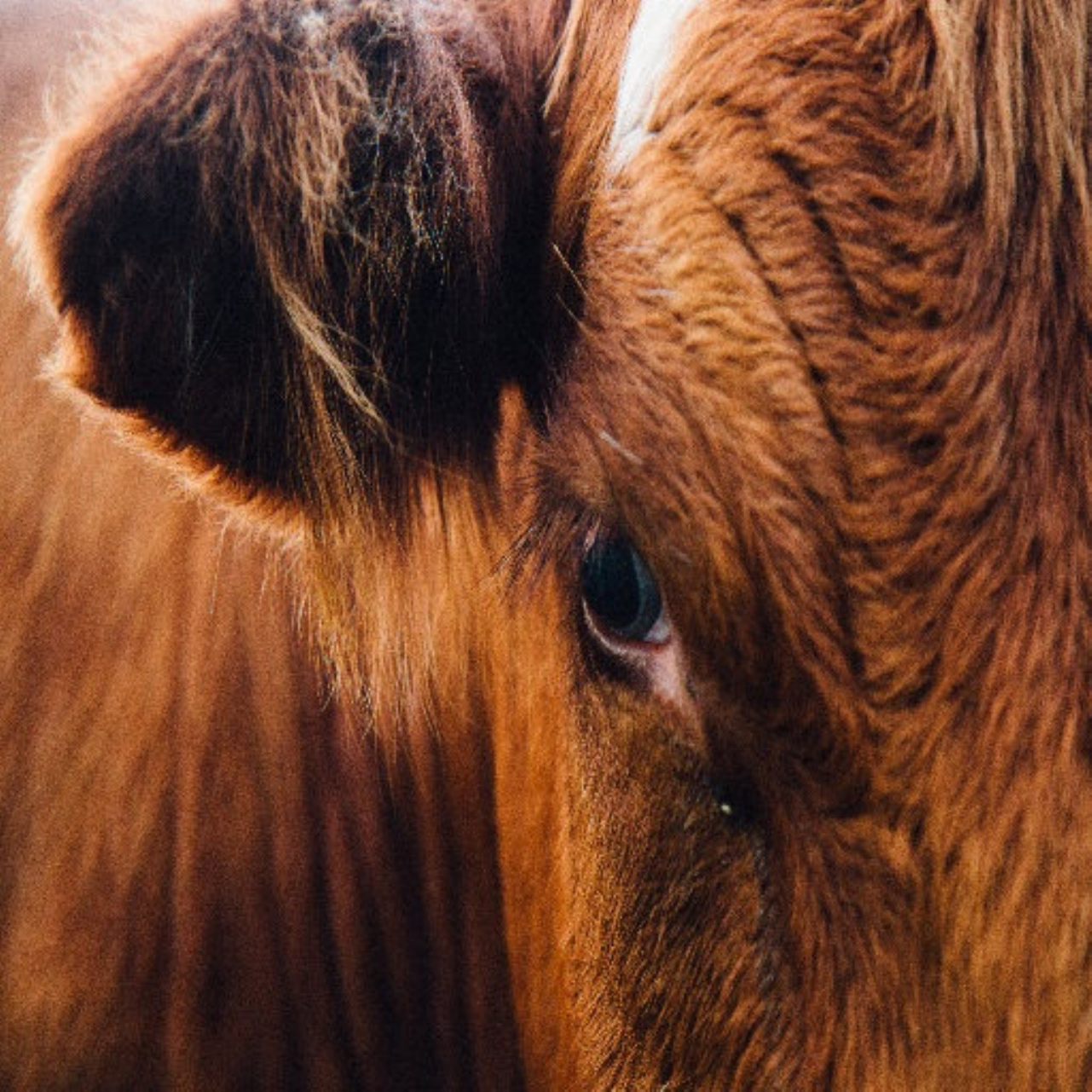}\vspace{0.51pt}
\includegraphics[width=0.31\linewidth]{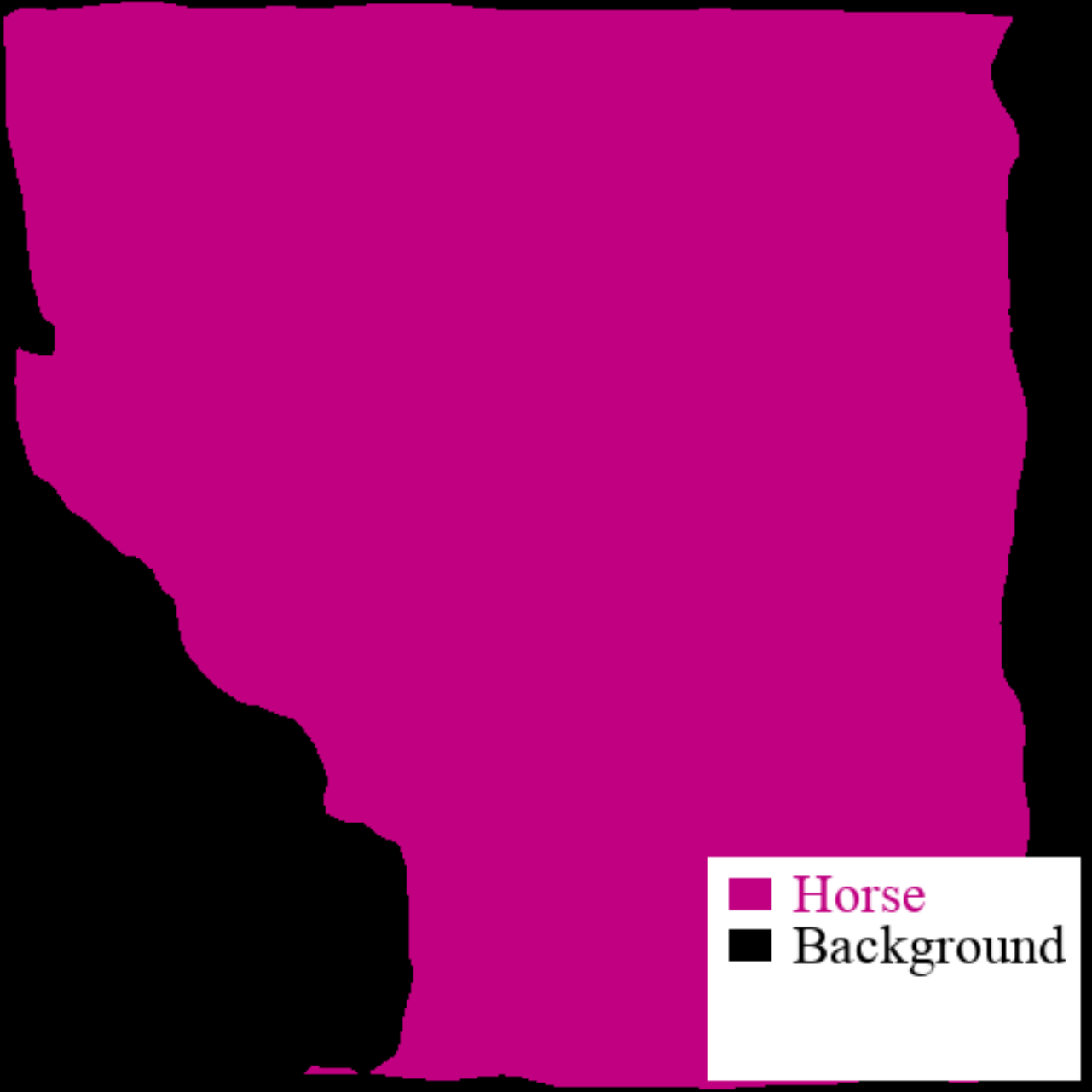}\vspace{0.51pt}
\end{minipage}
\begin{minipage}[]{0.48\linewidth}
\includegraphics[width=0.31\linewidth]{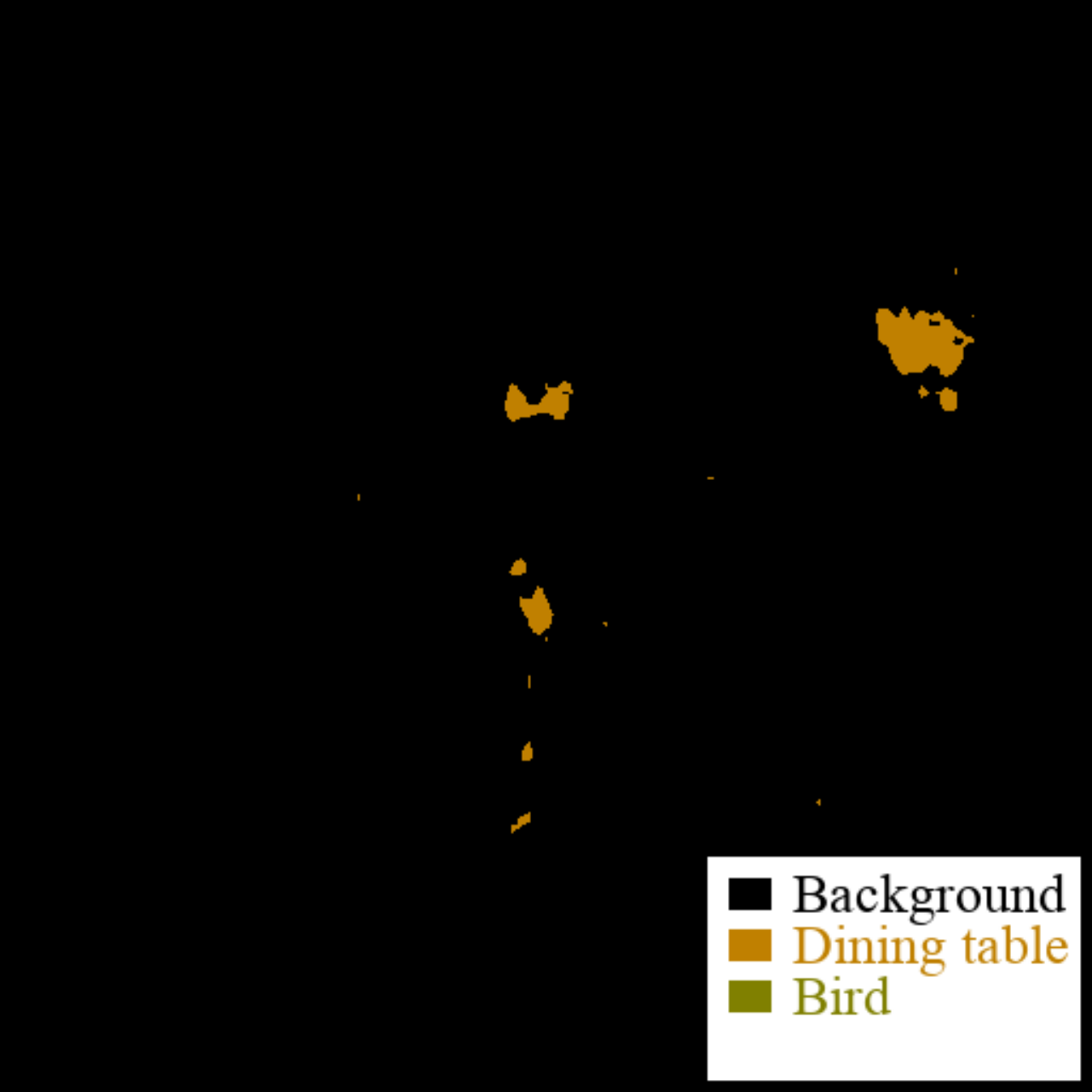}\vspace{0.51pt}
\includegraphics[width=0.31\linewidth]{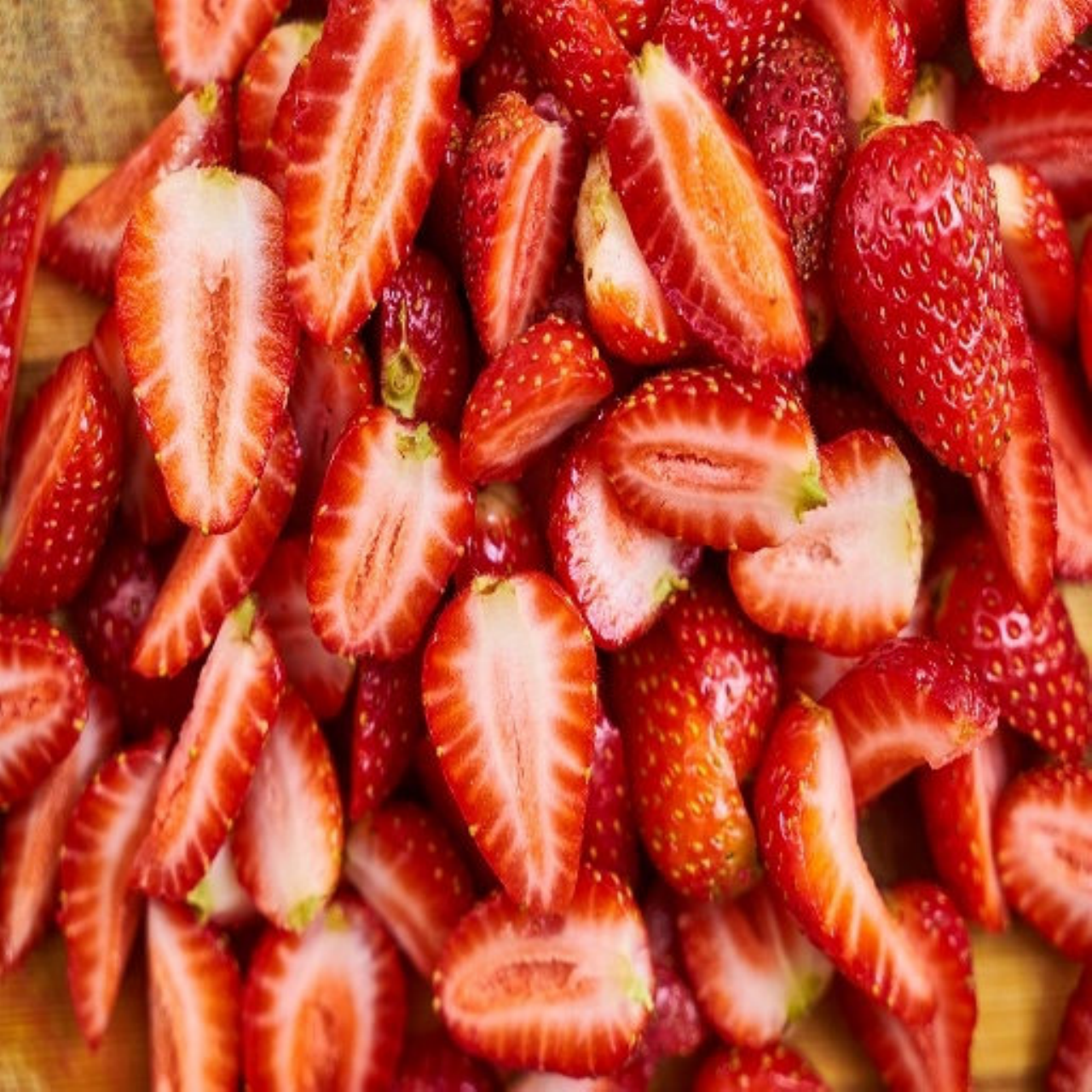}\vspace{0.51pt}
\includegraphics[width=0.31\linewidth]{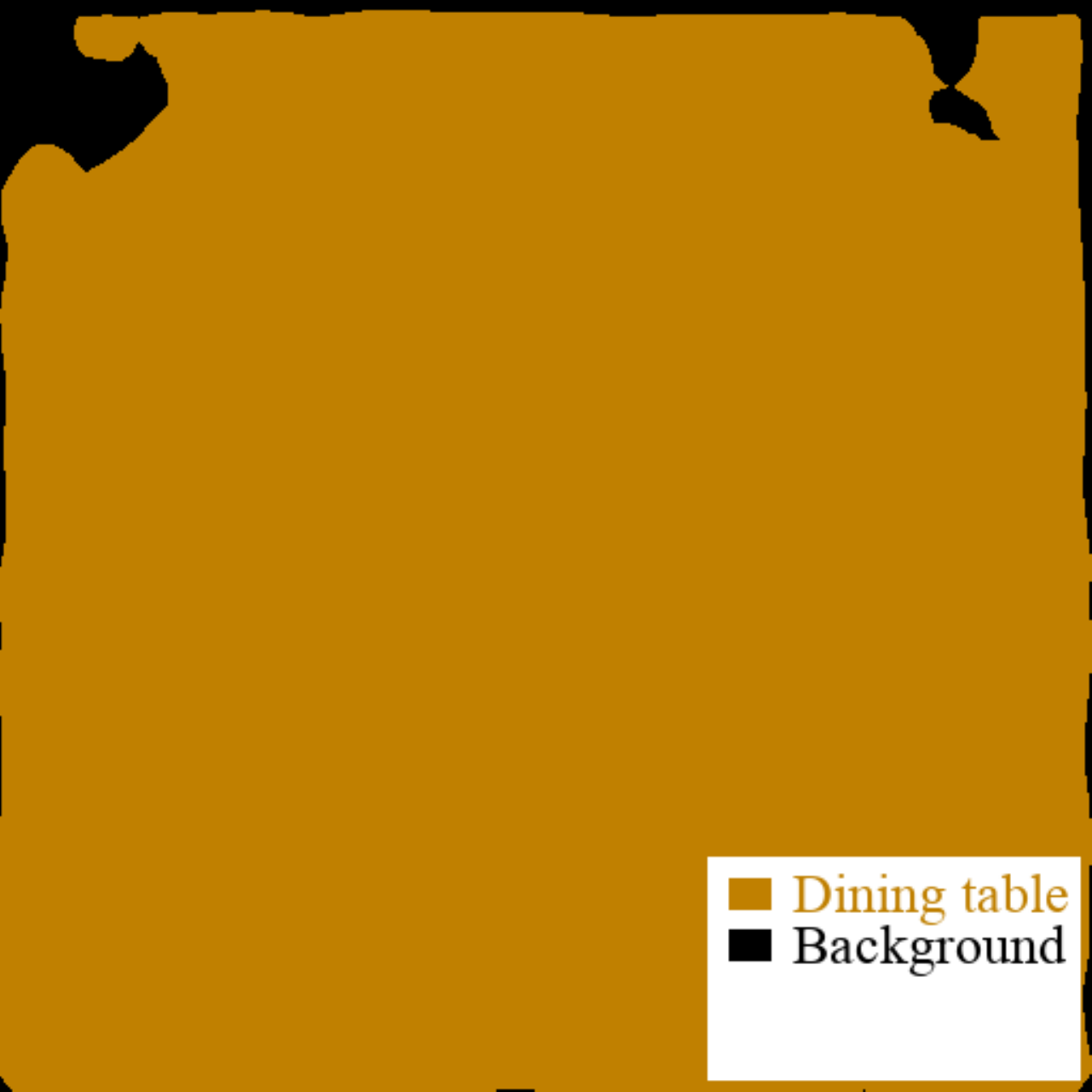}\vspace{0.51pt}\\
\includegraphics[width=0.31\linewidth]{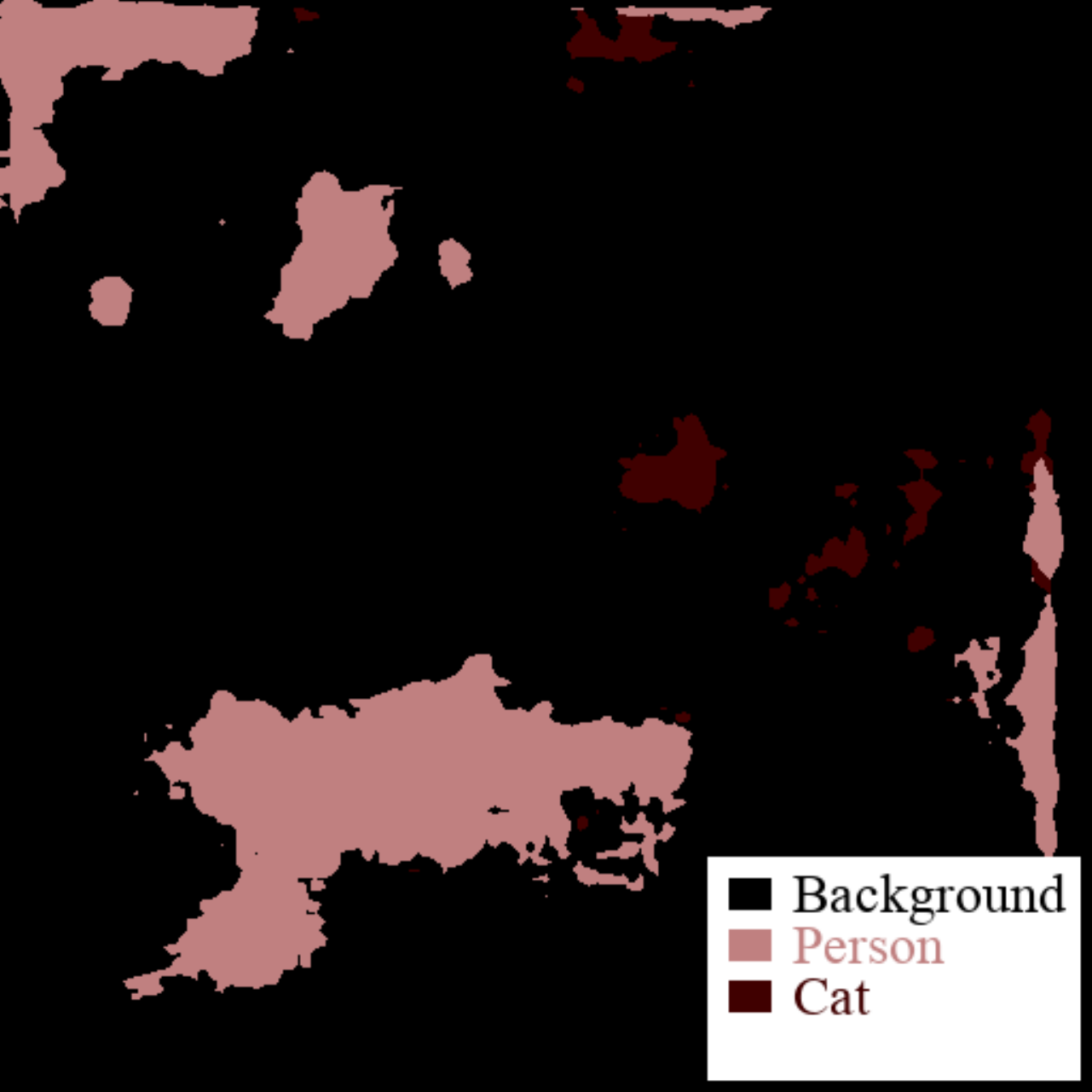}\vspace{0.51pt}
\includegraphics[width=0.31\linewidth]{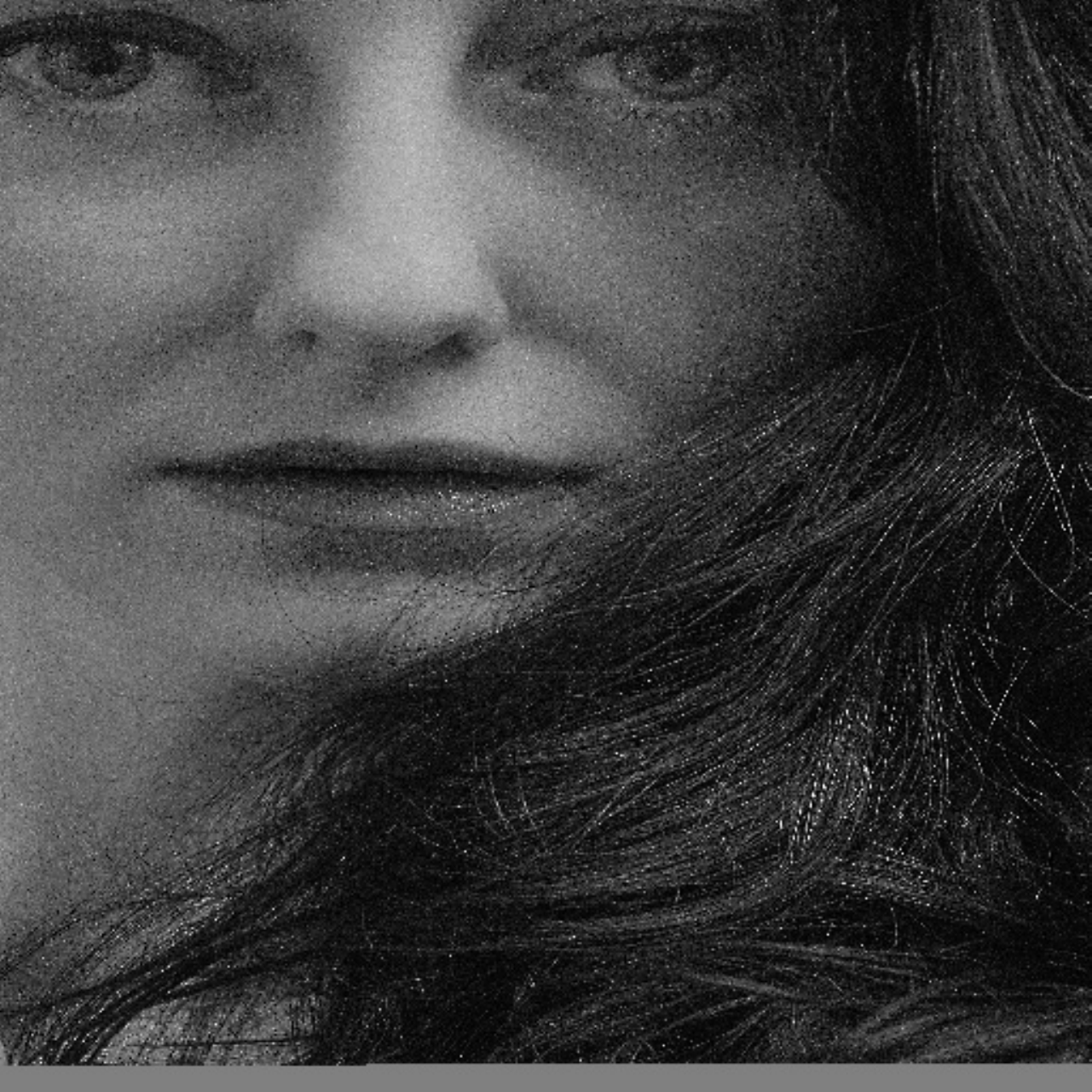}\vspace{0.51pt}
\includegraphics[width=0.31\linewidth]{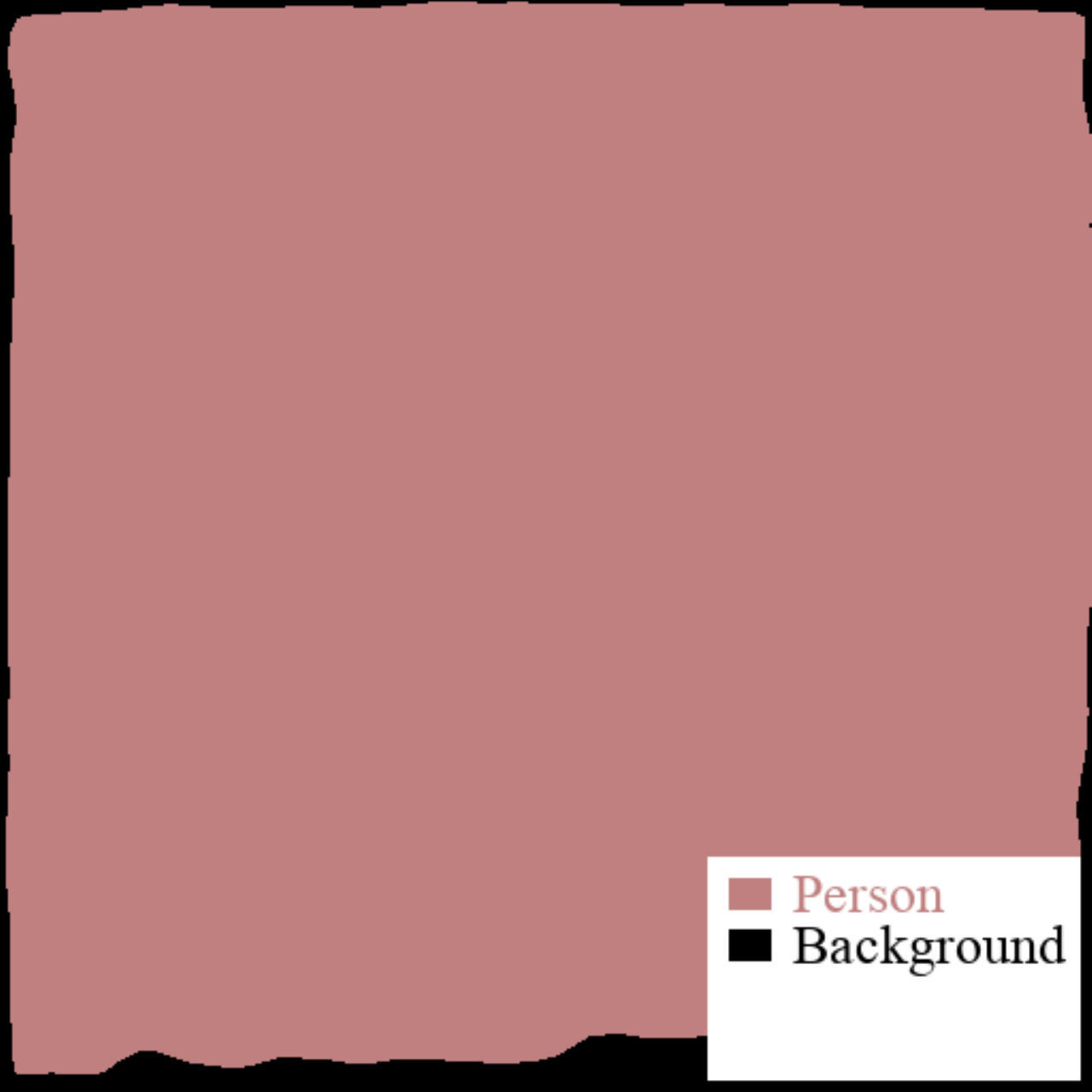}\vspace{0.51pt}
\end{minipage}}
\subfigure[]{
\begin{minipage}[]{0.48\linewidth}
\includegraphics[width=0.31\linewidth]{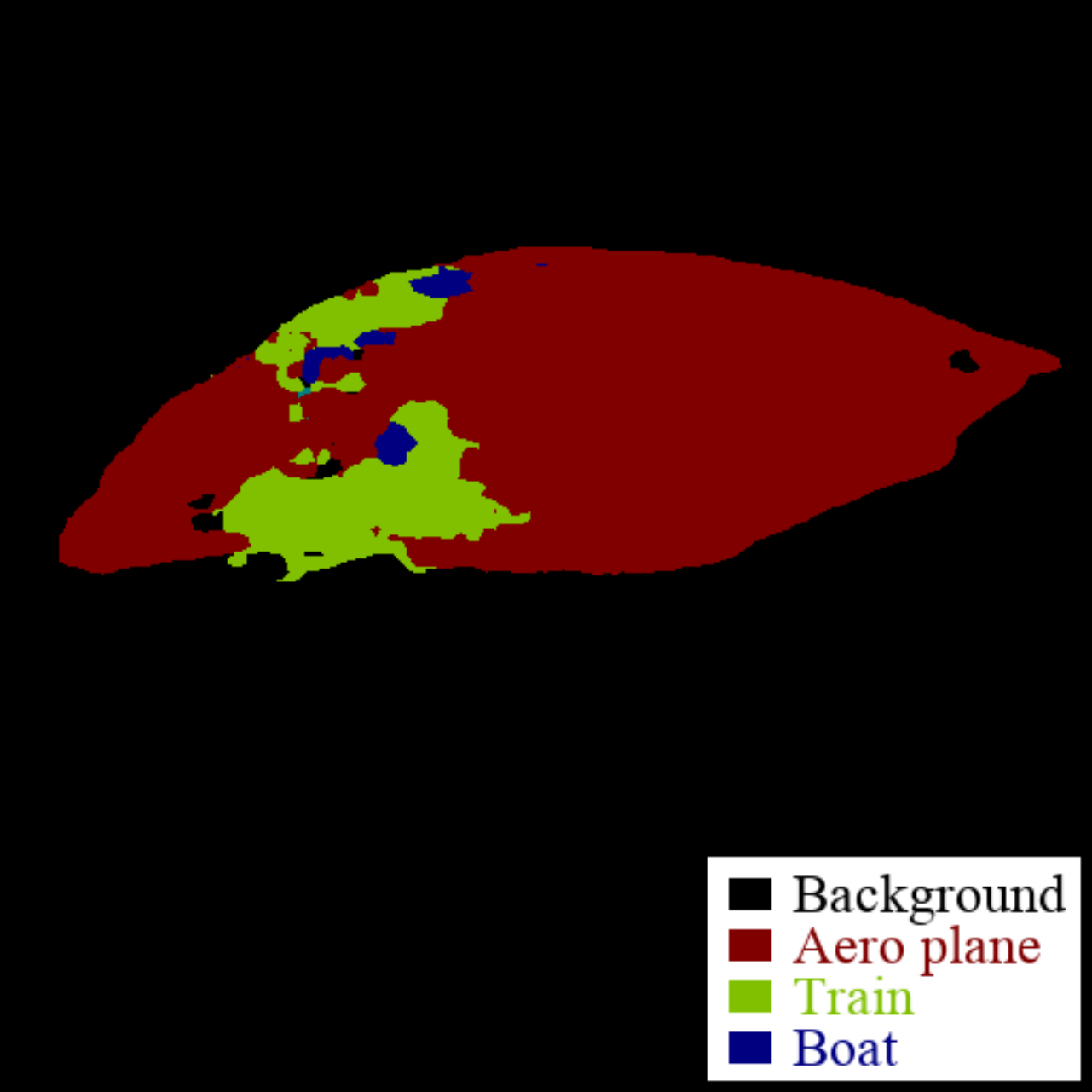}\vspace{0.51pt}
\includegraphics[width=0.31\linewidth]{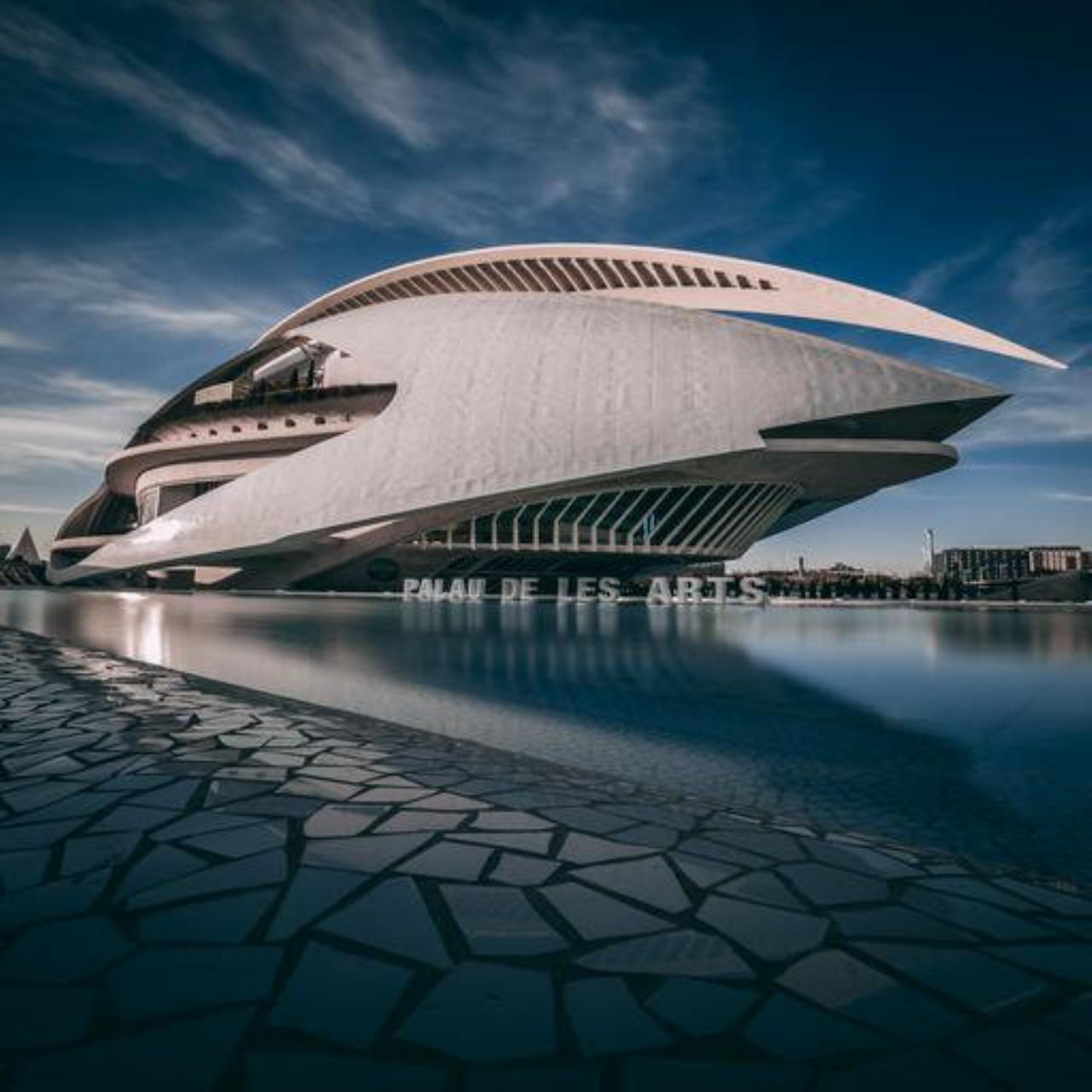}\vspace{0.51pt}
\includegraphics[width=0.31\linewidth]{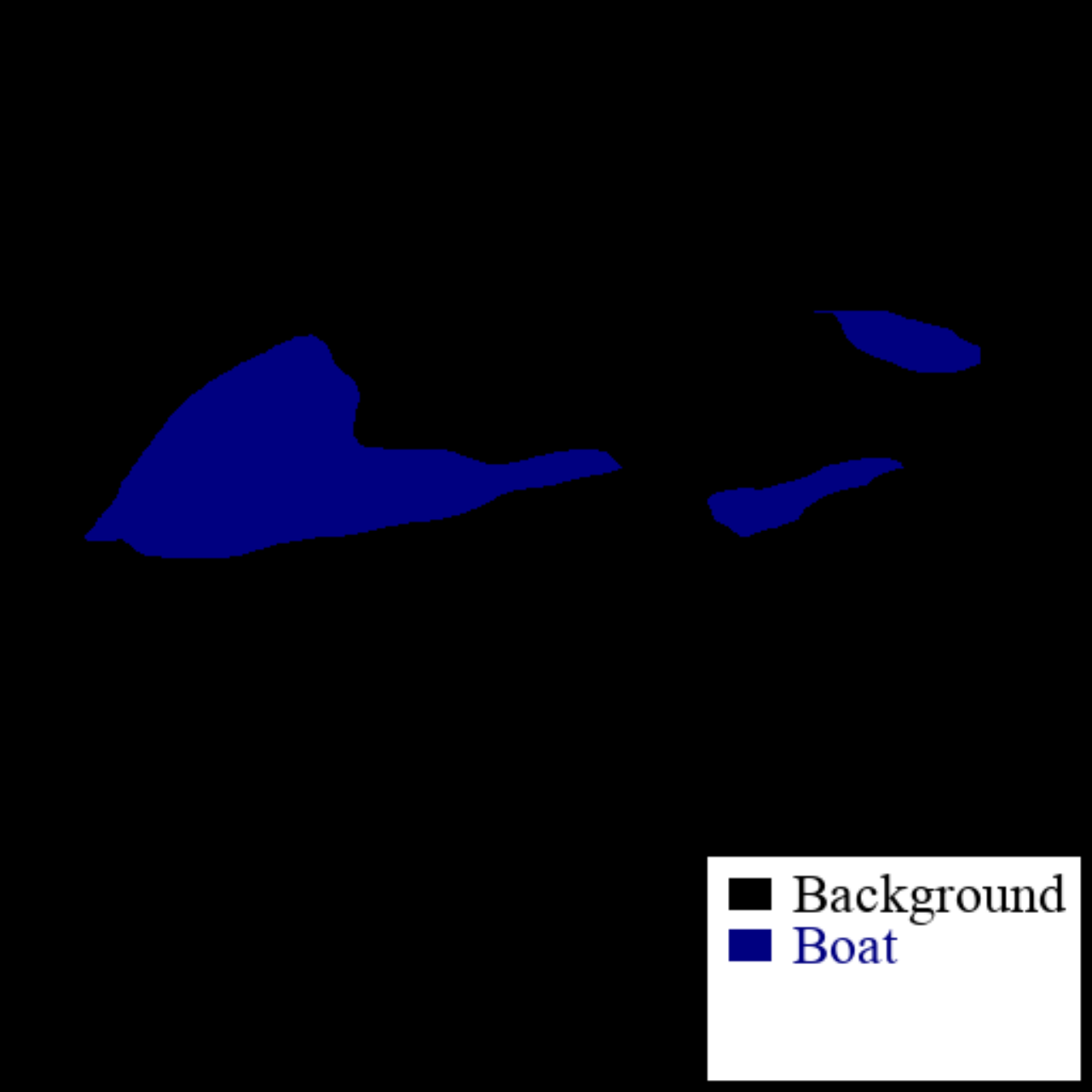}\vspace{0.51pt}\\
\includegraphics[width=0.31\linewidth]{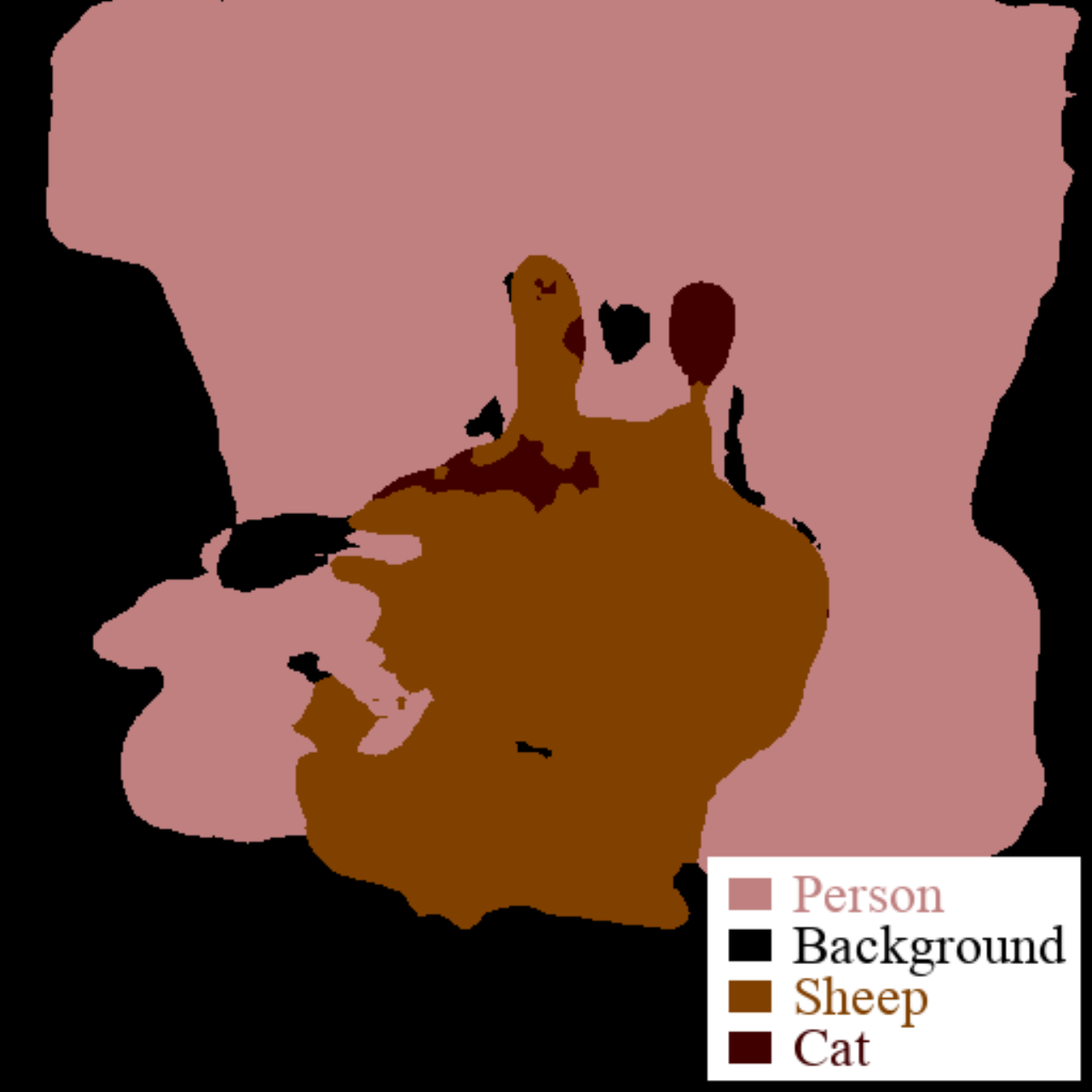}\vspace{0.51pt}
\includegraphics[width=0.31\linewidth]{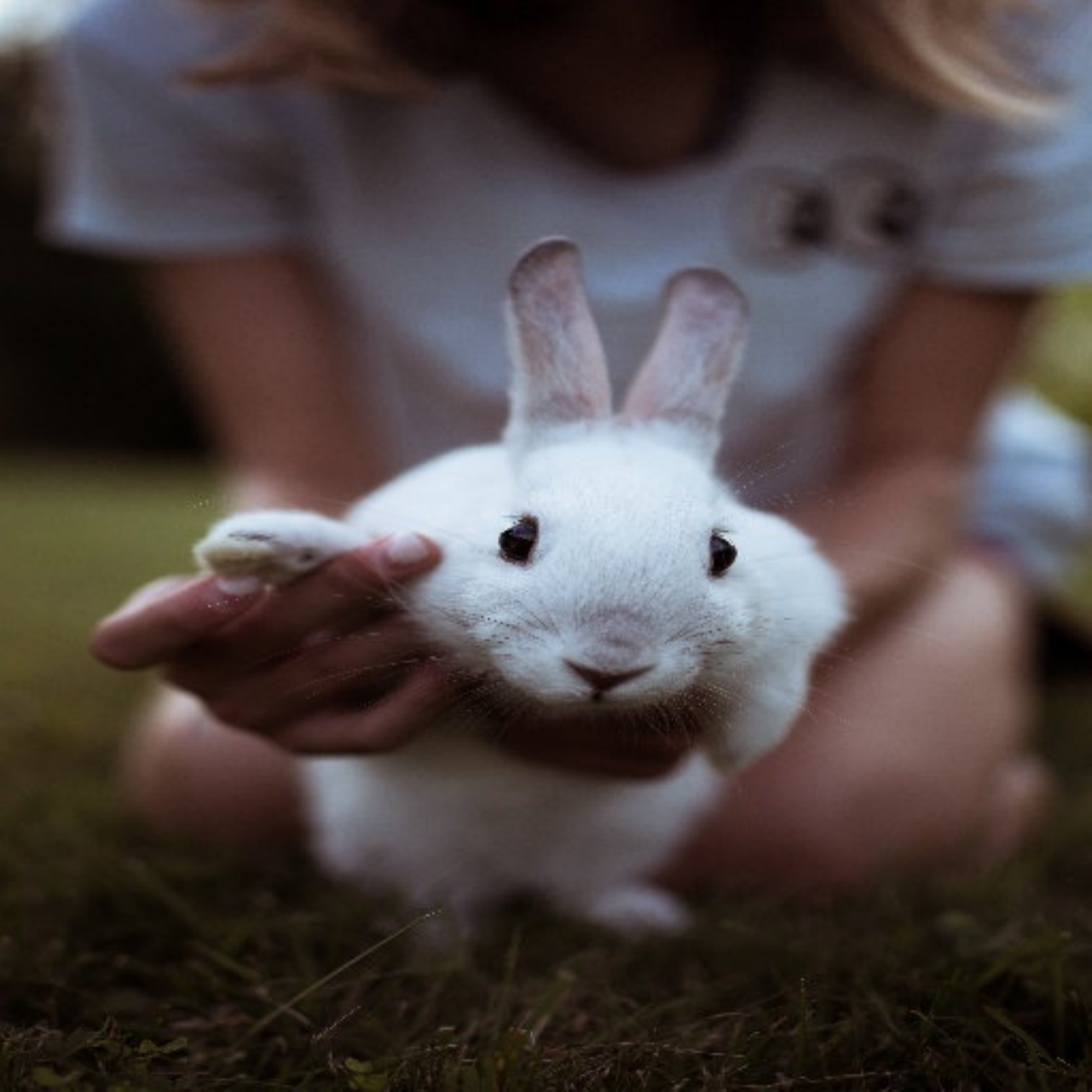}\vspace{0.51pt}
\includegraphics[width=0.31\linewidth]{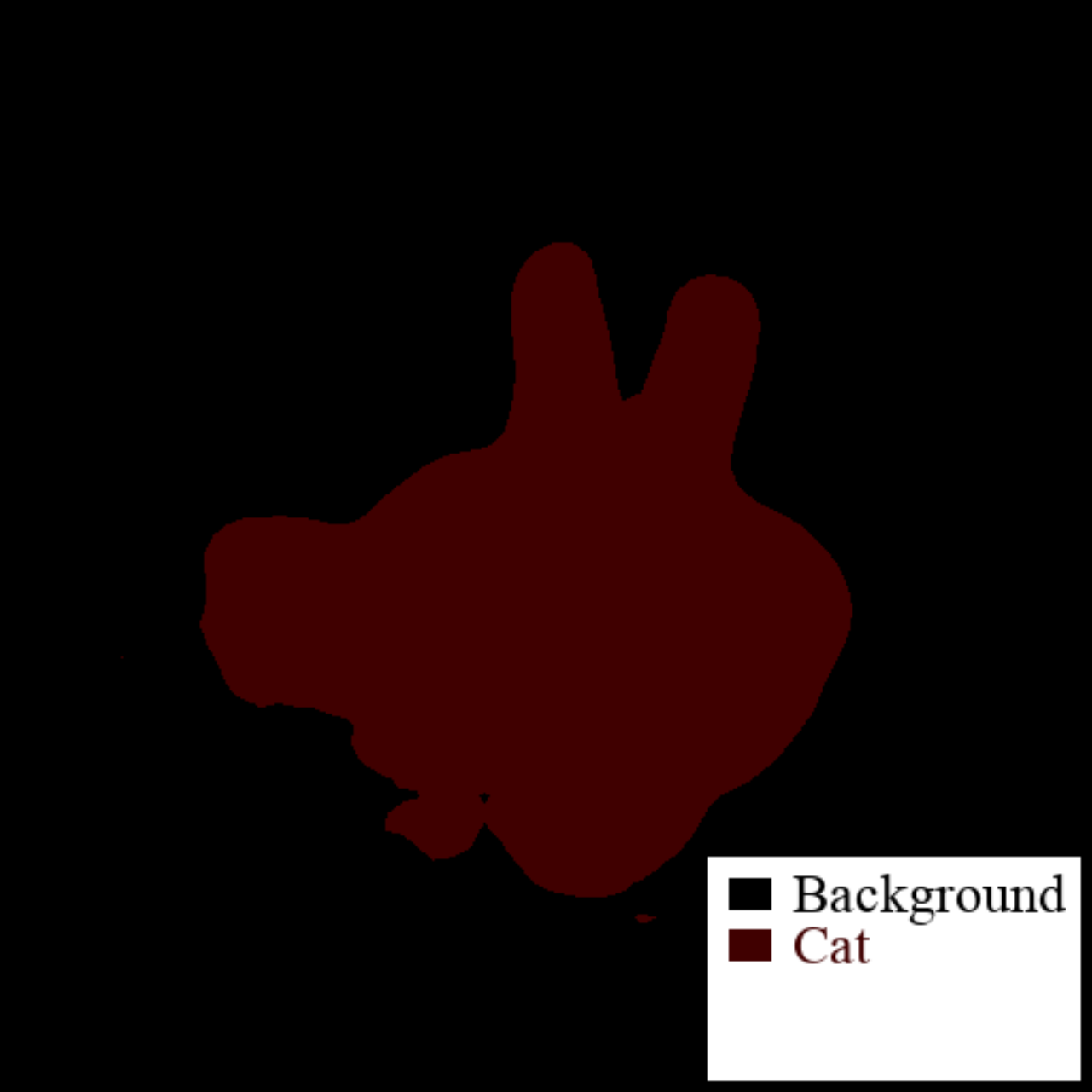}\vspace{0.51pt}
\end{minipage}
\begin{minipage}[]{0.48\linewidth}
\includegraphics[width=0.31\linewidth]{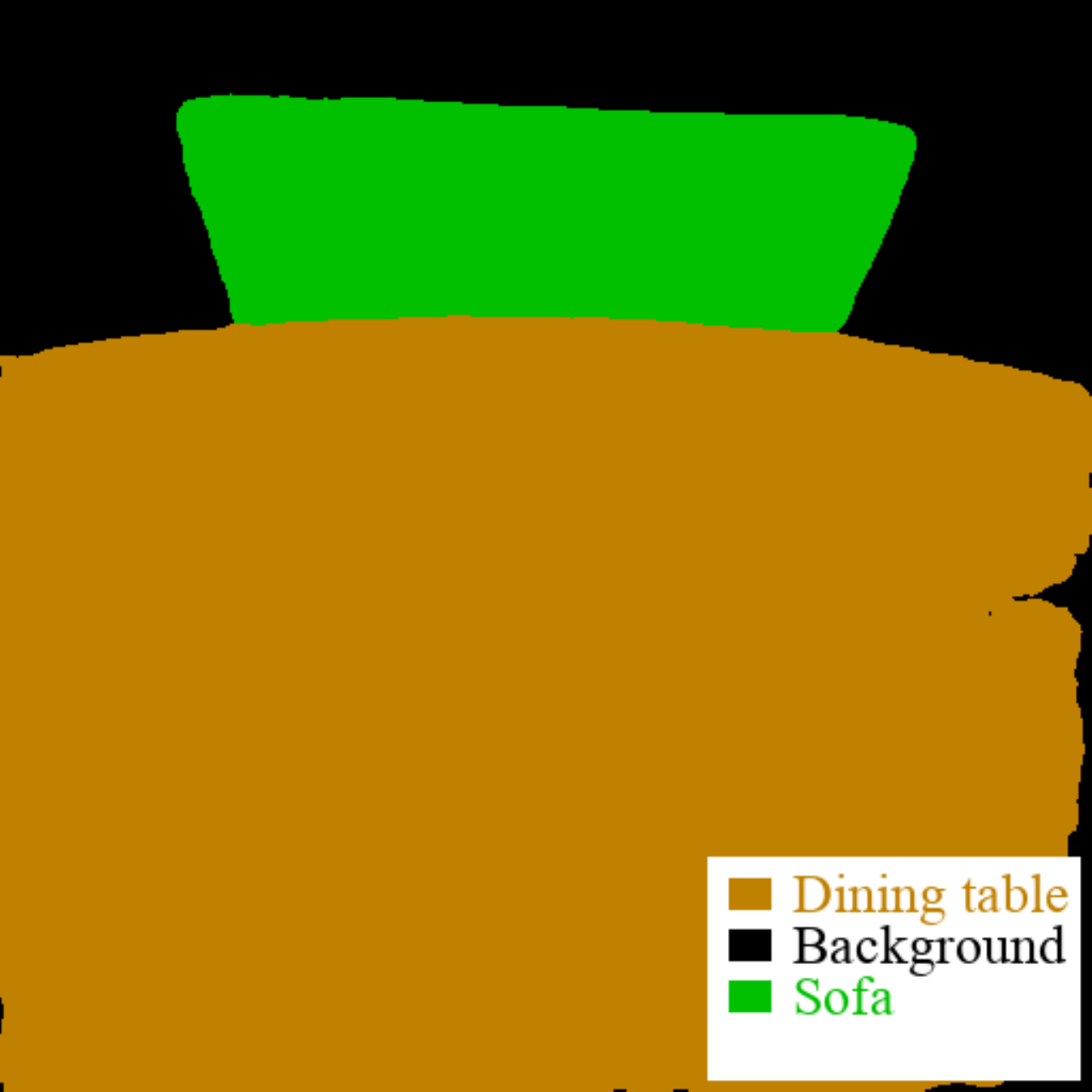}\vspace{0.51pt}
\includegraphics[width=0.31\linewidth]{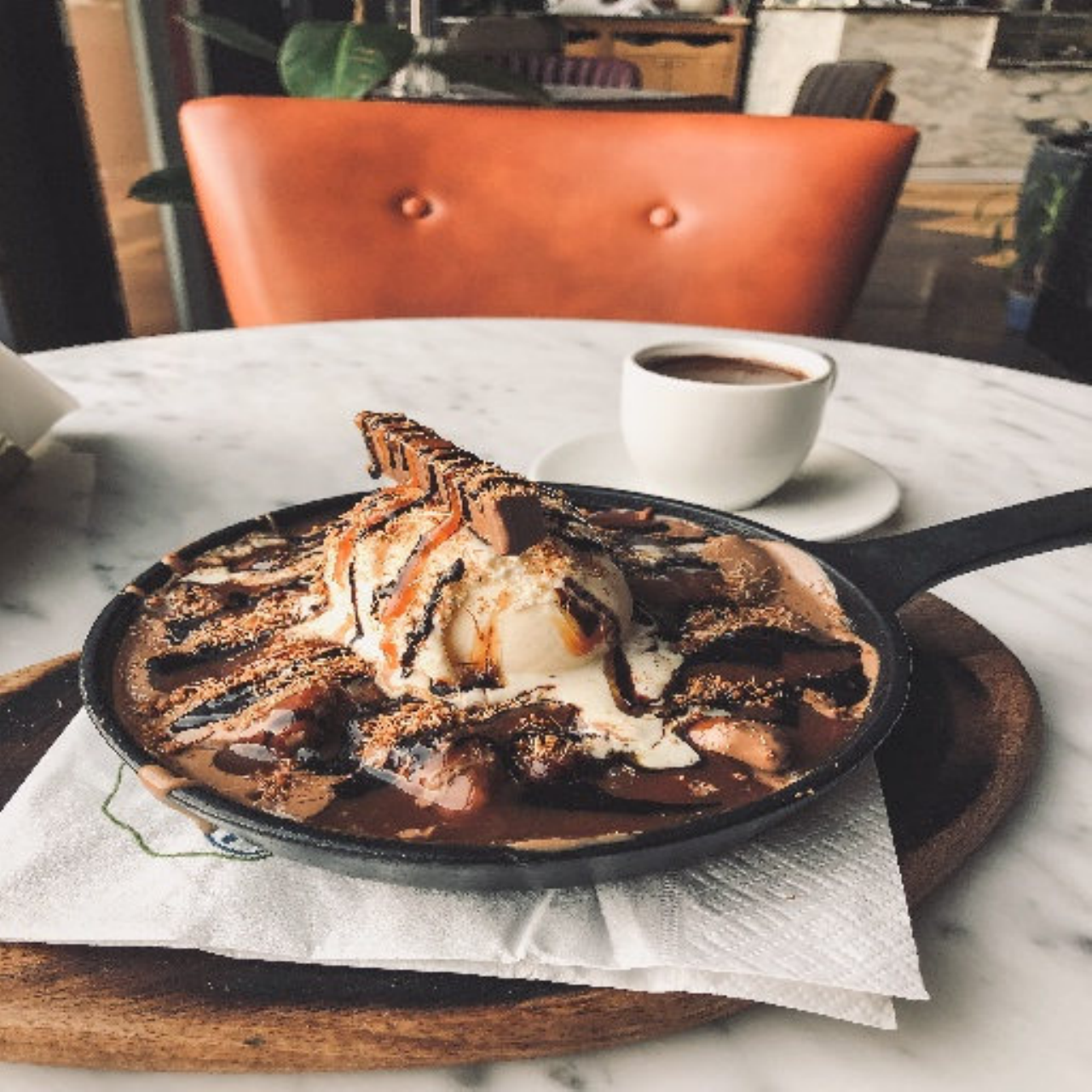}\vspace{0.51pt}
\includegraphics[width=0.31\linewidth]{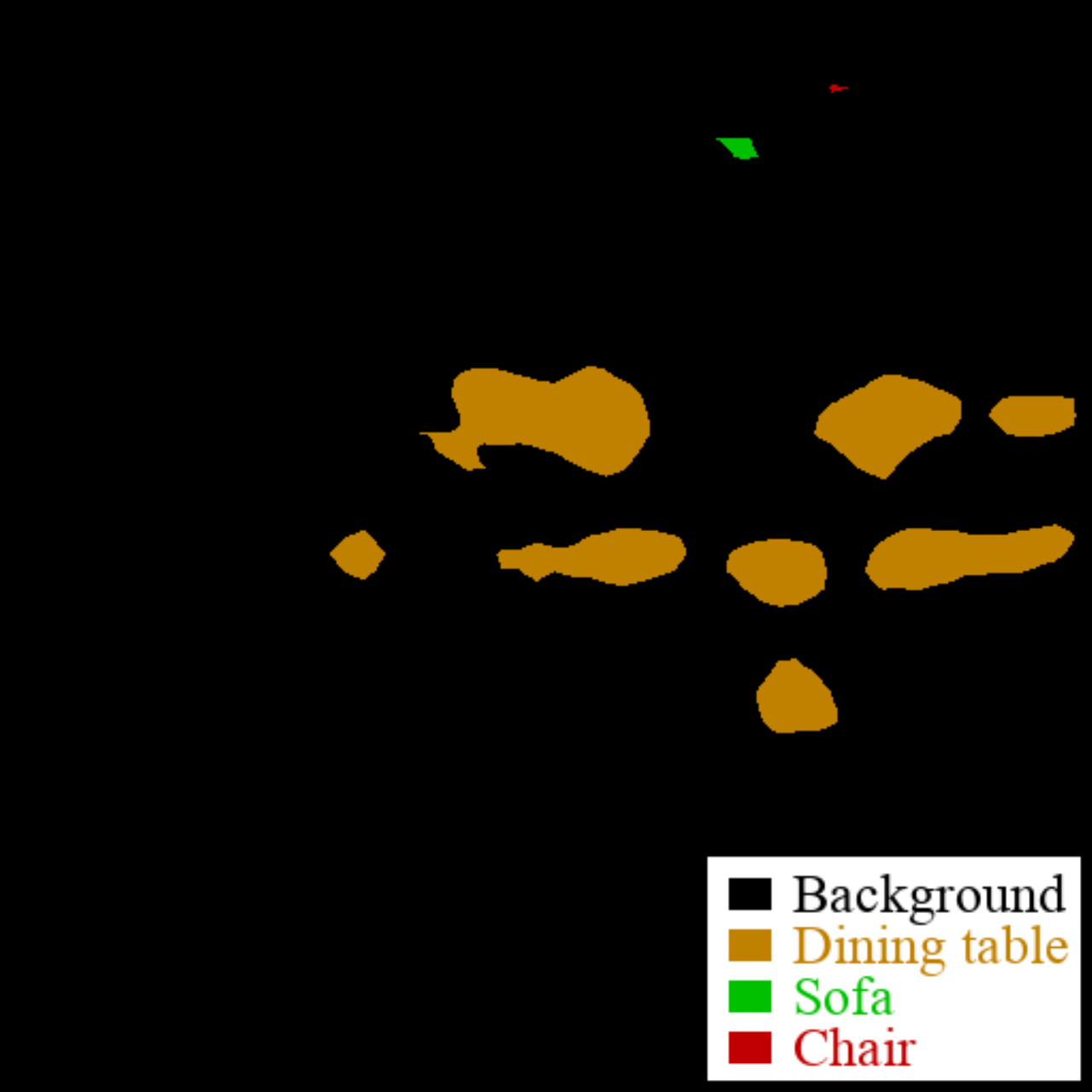}\vspace{0.51pt}\\
\includegraphics[width=0.31\linewidth]{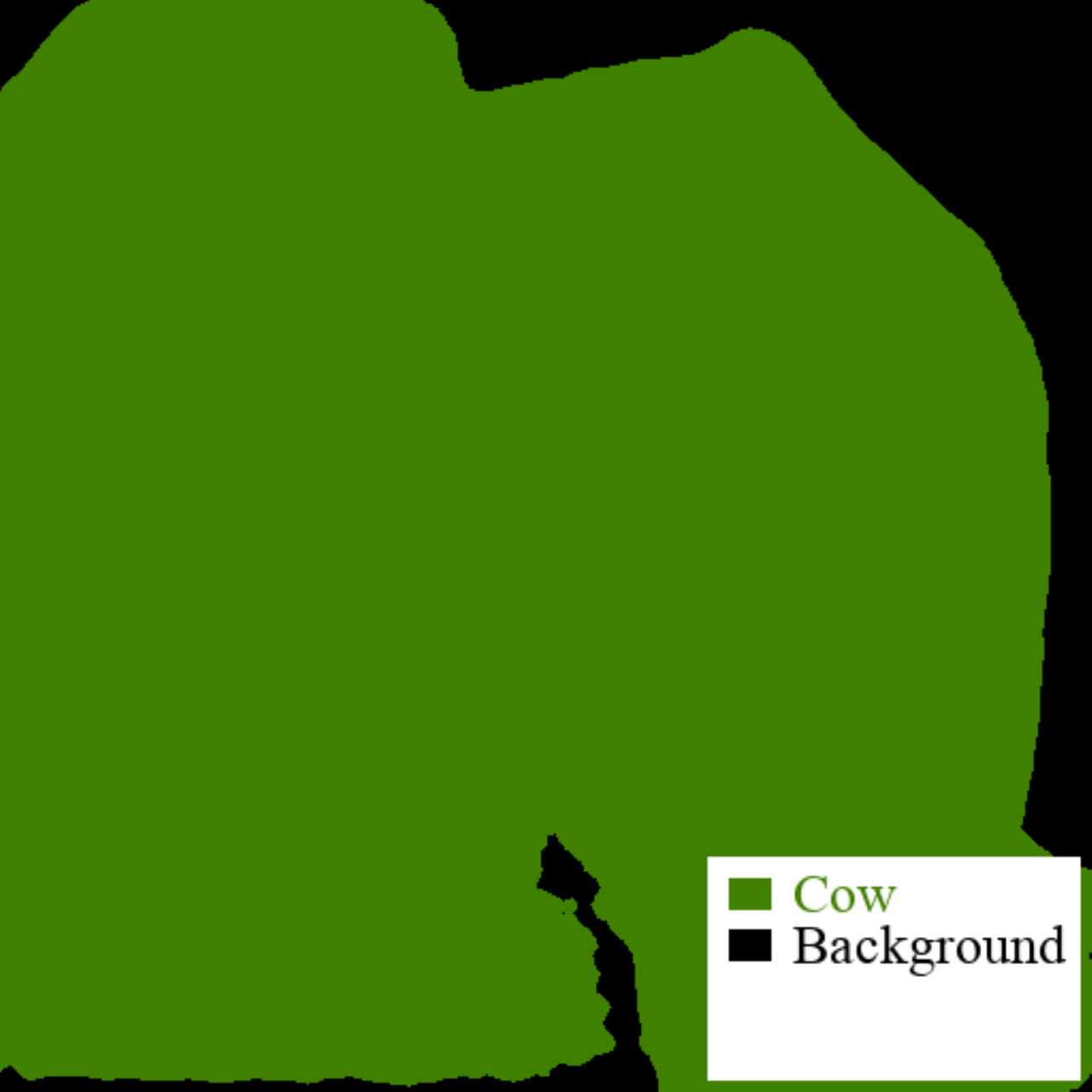}\vspace{0.51pt}
\includegraphics[width=0.31\linewidth]{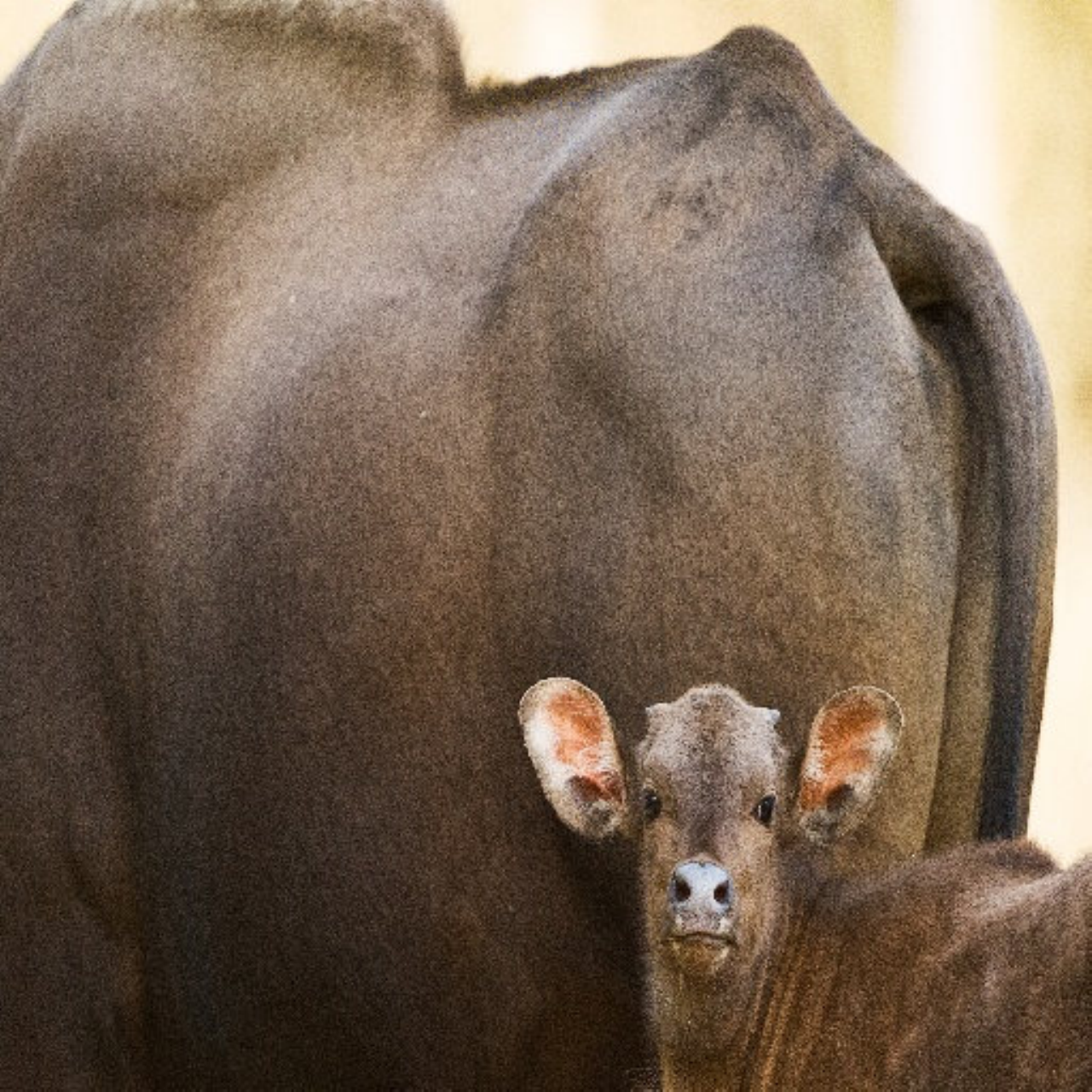}\vspace{0.51pt}
\includegraphics[width=0.31\linewidth]{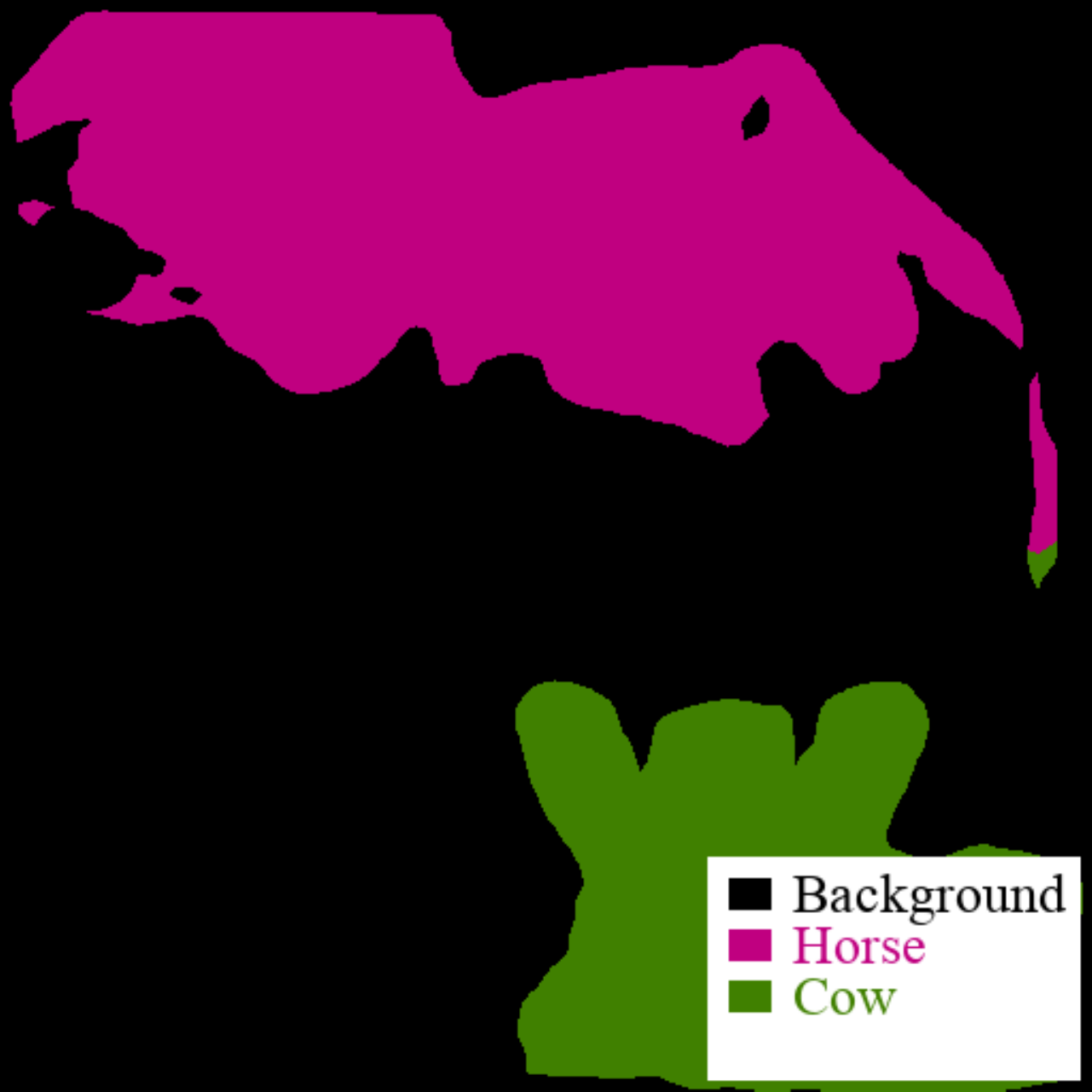}\vspace{0.51pt}
\end{minipage}}
\vspace{-.3cm}
\caption{An illustration of the set of empirical rules to encourage the diversity of the MAD-selected images in terms of scene content and error type. In this figure, we compare two top-performing semantic segmentation algorithms on PASCAL VOC \citep{Everingham2015} - DeepLabv3+ \citep{Chen2018v3+} and EMANet \citep{Lixia2019}. The $S(\cdot)$ in Eq. \eqref{eq:disc} is implemented by mIoU.  \textbf{(a)} The MAD samples without any constraints. It is not hard to observe that objects of the same categories and large scales appear much more frequently. \textbf{(b)} The MAD samples by adding the category constraint. The proposed method is refined to dig images of different categories, but still falls short in mining objects of different scales. \textbf{(c)} The MAD samples by further adding the scale constraint, which are more diverse in objective category and scale, therefore increasing the possibility of exposing different types of corner-case failures.}
\label{fig:constraint}
\end{figure*}

\subsection{The MAD Competition Procedure}
\label{subsec_flowchart}
We formulate the general problem of exposing failures in semantic segmentation as follows. We assume a natural image set $\mathcal{D} = \{x^{(m)}\}_{m=1}^M$ of an arbitrarily large size, $M$, whose purpose is to approximate the open visual world. For simplicity, each $x\in\mathcal{D}$ contains $N$ pixels, $x = \{x_n\}_{n=1}^N$. Also assumed is a subjective experimental environment, where we can associate each pixel with a class label $f(x^{(m)}_n)\in \mathcal{Y}$ for $n\in\{1,2\ldots, N\}$, $m\in\{1,2,\ldots, M\}$, and $\mathcal{Y}=\{1,2,\ldots, \vert\mathcal{Y}\vert\}$. Moreover, we are given $J$ semantic segmentation models $\mathcal{F}= \{f_j\}_{j=1}^J$, each of which accepts $x\in\mathcal{D}$ as input, and makes a dense prediction of $f(x) \in \mathcal{Y}^N$, collectively denoted by $\{f_j(x)\}_{j=1}^J$. The ultimate goal is to expose the failures of $\mathcal{F}$ and compare their relative performance under a minimum human labeling budget.

\subsubsection{Main Idea} We begin with the simplest scenario, where we plan to expose failures of two semantic segmentation methods $f_1$ and $f_2$. We are allowed to collect the ground-truth map $f(x)$ for a single image $x\in\mathcal{D}$ due to the most stringent labeling budget. The MAD competition suggests to select $\hat{x}$ by maximizing the discrepancy between the predictions of the two methods, or equivalently  minimizing their concordance:  
\begin{align}
\label{eq:disc}
\hat{x}= \mathop{\arg\min}_{x \in \mathcal{D}} S(f_1(x), f_2(x)),
\end{align}
where $S(\cdot)$ denotes a common metric in semantic segmentation, \eg, mIoU.
Comparison of $f_1(\hat{x})$ and $f_2(\hat{x})$ against the human-labeled $f(\hat{x})$ leads to three plausible outcomes:
\begin{itemize}\setlength{\itemsep}{12pt}
\item[] Case \uppercase\expandafter{\romannumeral1}: Both methods give satisfactory results (see Fig. \ref{fig:cases} (a)). This could happen in theory, for example if both $f_1$ and $f_2$ are nearly optimal, approaching the human-level performance. In practice, we would like to minimize this possibility since the selected $\hat{x}$ is not the counterexample of either method. This can be achieved by constantly increasing the size of the unlabeled set $\mathcal{D}$ while ensuring the design differences of $f_1$ and $f_2$.

\item[] Case \uppercase\expandafter{\romannumeral2}: One method (\eg, $f_1$) makes significantly better prediction than the other (\eg, $f_2$). That is, we have successfully exposed a strong failure of $f_2$ with the help of $f_1$ (see Figs. \ref{fig:cases} (b) and (c)).
The selected $\hat{x}$ is the most valuable in ranking their relative performance.

\item[] Case \uppercase\expandafter{\romannumeral3}: Both methods make poor predictions, meaning that we have automatically spotted a strong failure of both $f_1$ and $f_2$ (see Fig. \ref{fig:cases} (d)). Although the selected $\hat{x}$ offers useful information on how to improve the competing models, it contributes less to the relative performance comparison between $f_1$ and $f_2$.
\end{itemize}

\begin{algorithm}[t]
\caption{The MAD Competition for Semantic Segmentation}
\label{alg:Framwork}
\KwIn{A large-scale unlabeled image set $\mathcal{D}$, a list of semantic segmentation models $\mathcal{F}=\left\{f_{j}\right\}_{j=1}^{J}$, and a concordance measure $S$\\}
\KwOut{Two global ranking vectors $\mu^{(a)}, \mu^{(r)}\in \mathbb{R}^{J}$}
$\mathcal{S}\leftarrow\emptyset$, $A\leftarrow I$, and $R\leftarrow I$
\algorithmiccomment{$I$ is the $J\times J$ identity matrix}

\For{$i \gets 1$ \KwTo $J$}
{
Compute the segmentation results $\left\{f_{i}(x)\vert x \in \mathcal{D}\right\}$
}
\For{$i \gets 1$ \KwTo $J$}
{
\For{$j \gets 1$ \KwTo $J$}
{
\While{$j\ne i$}
{
Compute the concordance $\{S(f_i(x), f_j(x))\vert x\in\mathcal{D}\}$


Divide $\mathcal{D}$ into $\vert\mathcal{Y}\vert$ overlapping groups $\left\{\mathcal{D}^{(y)}_{ij}\right\}_{y=1}^{\vert\mathcal{Y}\vert}$ based on the predictions of $f_i$ (\ie, $f_i$ is the defender)
\label{alg:Framwork:com}

\For{$y \gets 1$ \KwTo $\vert\mathcal{Y}\vert$}
{
Filter $\mathcal{D}^{(y)}_{ij}$ according to the scale constraint

Create the MAD subset $\mathcal{S}^{(y)}_{ij}$ by selecting top-$K$ images in the filtered $\mathcal{D}^{(y)}_{ij}$ that optimize Eq. \eqref{eq:sub}

$\mathcal{S}\leftarrow\mathcal{S}\bigcup\mathcal{S}^{(y)}_{ij}$
\label{alg:Framwork:result}
}
}
}
}
Collect human segmentation maps for $\mathcal{S}$

Compute the aggressiveness matrix $A$ and the resistance matrix $R$ using Eq. \eqref{eq:agg} and Eq. \eqref{eq:res}, respectively

Aggregate paired comparison results into  two global ranking vectors $\mu^{(a)}$ and $\mu^{(r)}$ by maximum likelihood estimation
\end{algorithm}

\begin{algorithm}
\caption{Adding a New Segmentation Method into the MAD Competition}
\label{alg:AddFramwork}
\KwIn{The same unlabeled image set $\mathcal{D}$, a new segmentation model $f_{J+1}$, the aggressiveness and resistance matrices $A, R\in\mathbb{R}^{J\times J}$ for $\mathcal{F}\in\{f_j\}_{j=1}^J$, and a concordance measure $S$}
\KwOut{Two ranking vectors $\mu^{(a)}, \mu^{(r)}\in \mathbb{R}^{J+1}$}

$\mathcal{S}\leftarrow\emptyset$, $A'\leftarrow\begin{bmatrix}
A & 0 \\
      0^{T} & 1
\end{bmatrix}$, and $R'\leftarrow\begin{bmatrix}
 R & 0 \\
      0^{T} & 1
\end{bmatrix}$

Compute the segmentation results $\left\{f_{J+1}(x), x \in \mathcal{D}\right\}$

\For{$i \gets 1$ \KwTo $J$}
{
Compute the concordance $\{S(f_i(x),f_{J+1}(x))\vert x\in\mathcal{D}\}$

Select $f_{i}$ and $f_{J+1}$ as the defender and the attacker, respectively

Divide $\mathcal{D}$ into $\vert\mathcal{Y}\vert$ overlapping groups $\left\{\mathcal{D}^{(y)}_{i,J+1}\right\}_{y=1}^{\vert\mathcal{Y}\vert}$ based on the predictions of the defender
\label{alg:AddFramwork:com}

\For{$y \gets 1$ \KwTo $\vert\mathcal{Y}\vert$}
{
Filter $\mathcal{D}^{(y)}_{i,J+1}$ according to the scale constraint

Create the MAD subset $\mathcal{S}^{(y)}_{i,J+1}$ by selecting top-$K$ images in the filtered $\mathcal{D}^{(y)}_{i,J+1}$ that optimize Eq. \eqref{eq:sub}

$\mathcal{S}\leftarrow\mathcal{S}\bigcup\mathcal{S}^{(y)}_{i,J+1}$
\label{alg:AddFramwork:result}
}
Switch the roles of $f_{i}$ and $f_{J+1}$, and repeat Step \ref{alg:AddFramwork:com} to Step \ref{alg:AddFramwork:result}
}

Collect human segmentation maps for $\mathcal{S}$

\For{$i \gets 1$ \KwTo $J$}
{
 $a'_{i,J+1}\gets P(f_i;\mathcal{S}_{J+1,i})/P(f_{J+1};\mathcal{S}_{J+1,i})$

 $a'_{J+1,i}\gets P(f_{J+1};\mathcal{S}_{i,J+1})/P(f_i;\mathcal{S}_{i,J+1})$

 $r'_{i,J+1}\gets P(f_i;\mathcal{S}_{i,J+1})/P(f_{J+1};\mathcal{S}_{i,J+1})$

 $r'_{J+1,i}\gets P(f_{J+1};\mathcal{S}_{J+1,i})/P(f_i;\mathcal{S}_{J+1,i})$
}

Update the two global ranking vectors $\mu^{(a)}$ and $\mu^{(r)}$
\end{algorithm}

\subsubsection{Selection of Multiple Images}
\label{subsubsec:mi}
It seems trivial to extend the MAD competition  using multiple images by selecting top-$K$ images that optimize Eq. \eqref{eq:disc}. However, this na\"{i}ve implementation may simply expose different instantiations of the same type of errors. Fig. \ref{fig:constraint} (a) shows such an example, where we compare DeepLabv3+ \citep{Chen2018v3+} and EMANet \citep{Lixia2019}. The four samples in sub-figure (a) are representative among the top-$30$ images selected by MAD. We find that objects of the same categories (\eg, dining table and potted plant) and extremely large scales (\eg, occupying the entire image) frequently occur, which makes the comparison less interesting and useful. Here we enforce a category constraint to encourage more diverse images to be selected in terms of scene content and error type. Specifically, we set up a defender-attacker game, where the two segmentation models, $f_1$ and $f_2$, play the role of the defender and the attacker, respectively. We divide the unlabeled set $\mathcal{D}$ into $\vert\mathcal{Y}\vert$ overlapping groups $\{\mathcal D^{(y)}_{12}\}_{y=1}^{\vert\mathcal{Y}\vert}$  according to the predictions by the defender $f_1$: if $f_1(x_n) = y$, for at least one spatial location $n \in\{1, 2,\ldots, N\}$, then $x\in \mathcal D^{(y)}_{12}$. After that, we divide the optimization problem in \eqref{eq:disc} into $\vert\mathcal{Y}\vert$ subproblems:
\begin{align}
\label{eq:sub}
\hat{x}^{(k)}= \mathop{\arg\min}_{x \in \mathcal{D}^{(y)}_{12}\setminus \{\hat{x}^{(m)}\}_{m=1}^{k-1}} S(f_1(x), f_2(x)), \forall y\in\{1,\ldots, \vert\mathcal{Y}\vert\},
\end{align}
where $\{\hat{x}^{(m)}\}_{m=1}^{k-1}$ is the set of $k-1$ images that have already been identified. We may reverse the role of $f_1$ and $f_2$, divide $\mathcal{D}$ into $\{\mathcal D^{(y)}_{21}\}_{y=1}^{\vert\mathcal{Y}\vert}$ based on $f_2$ which is now the defender, and solve a similar set of optimization problems to find top-$K$ images in each subset $\mathcal D^{(y)}_{21}$. Fig. \ref{fig:constraint} (b) shows the MAD samples associated with DeepLabv3+ \citep{Chen2018v3+} and EMANet \citep{Lixia2019}, when we add the category constraint. As expected, objects of different categories begin to emerge, but the selection is still biased towards objects with relatively large scales. To mitigate this issue, we add a second scale constraint.
Suppose that all competing models use the same training set $\mathcal{L}$, which is divided into $\vert\mathcal{Y}\vert$ groups $\{\mathcal{L}^{(y)}\}_{y=1}^{\vert\mathcal{Y}\vert}$ based on the ground-truth segmentation maps.
For $x\in \mathcal{L}^{(y)}$, we count the proportion of pixels that belong to the $y$-th object category, and extract the first quartile $T^{(y)}_\mathrm{min}$ and the third quartile $T^{(y)}_\mathrm{max}$ as the scale statistics. For $x \in \mathcal D^{(y)}_{12} \bigcup \mathcal D^{(y)}_{21}$, if the proportion of pixels belonging to the $y$-th category is in the range $[T^{(y)}_\mathrm{min}, T^{(y)}_\mathrm{max}]$, $x$ is valid for comparison of $f_{1}$ and $f_{2}$. Otherwise, $x$ will be discarded. Fig. \ref{fig:constraint} (c) shows the MAD samples of  DeepLabv3+ and EMANet by further adding the scale constraint. As can be seen, objects of different categories and scales have been spotted, which may correspond to failures of different underlying root causes.



\subsubsection{Comparison of Multiple Models}
We further extend the MAD competition to include $J$ segmentation models. For each out of $J\times (J-1)$ ordered pairs of segmentation models, $f_i$ and $f_j$, where $i\ne j$ and $i,j\in\{1,2\ldots, J\}$, we follow the procedure described in Section \ref{subsubsec:mi} to obtain $\vert\mathcal{Y}\vert$ MAD subsets $\{\mathcal{S}^{(y)}_{ij}\}_{y=1}^{\vert\mathcal{Y}\vert} $, where $\mathcal{S}^{(y)}_{ij}$ contains top-$K$ images from $\mathcal{D}^{(y)}_{ij}$ with $f_i$ and $f_j$ being the defender and the attacker, respectively. The final MAD set $\mathcal{S} = \bigcup \mathcal{S}^{(y)}_{ij}$ has a total of $J\times (J-1)\times \vert\mathcal{Y}\vert\times K$ images, whose number is independent of the size of the large-scale unlabeled image set $\mathcal{D}$. This nice property of MAD encourages us to expand $\mathcal{D}$ to include as many natural images as possible, provided that the computational cost for dense prediction is negligible.

After collecting the ground-truth segmentation maps for $\mathcal{S}$ (details in Section~\ref{subsec_subjective}), we compare the segmentation models in pairs by introducing the notions of aggressiveness and resistance \citep{ma2018group}. The aggressiveness $a_{ij}$ quantifies how aggressive $f_i$ is to identify the failures of $f_{j}$ as the attacker:
\begin{align}\label{eq:agg}
    a_{ij} = \frac{P(f_i;\mathcal{S}_{ji})}{P(f_j;\mathcal{S}_{ji})},
\end{align}
where
\begin{align}\label{eq:pm}
    P(f_i;\mathcal{S}_{ji}) = \frac{1}{\vert\mathcal{Y}\vert}\sum_{y= 1}^{\vert\mathcal{Y}\vert}\left(\frac{1}{\left\vert\mathcal{S}^{(y)}_{ji}\right\vert}\sum_{x\in \mathcal{S}^{(y)}_{ji}}S(f_i(x), f(x))\right).
\end{align}
Note that a small positive constant $\epsilon$ may be added to both the numerator and the denominator of Eq. \eqref{eq:agg} as a form of Laplace smoothing to avoid potential division by zero. $a_{ij}$ is nonnegative with a higher value indicating stronger aggressiveness.  The resistance $r_{ij}$ measures how resistant $f_i$ is against the attack from $f_j$:
\begin{align}\label{eq:res}
    r_{ij} = \frac{P(f_i;\mathcal{S}_{ij})}{P(f_j;\mathcal{S}_{ij})}
\end{align}
with a higher value indicating stronger resistance. The pairwise aggressiveness and resistance statistics of all $J$ segmentation models form two $J\times J$ matrices $A$ and $R$, respectively, with ones on diagonals.
Next, we convert pairwise comparison results into global rankings using maximum likelihood estimation~\citep{Tsukida2011}. Letting $\mu^{(a)} = [\mu^{(a)}_1,\mu^{(a)}_2,\ldots,\mu^{(a)}_J]$ be the vector of global aggressive scores, we define the log-likelihood of the aggressive matrix $A$ as
\begin{align}\label{eq:ml}
    L(\mu^{(a)}|A) = \sum_{ij}^{}a_{ij}\log\left(\Phi\left(\mu^{(a)}_{i}-\mu^{(a)}_{j}\right)\right),
\end{align}
where $\Phi(\cdot)$ is the standard Gaussian cumulative distribution function. Direct optimization of $L(\mu^{(a)}|A)$ suffers from the translation ambiguity. To obtain a unique solution, one often adds an additional constraint that $\sum_i \mu^{(a)}_i = 1$ (or $\mu^{(a)}_1=0$). The vector of global resistance scores $\mu^{(r)} \in \mathbb{R}^J$ can be obtained in a similar way. We summarize the MAD competition procedure in Algorithm~\ref{alg:Framwork}.

Last, we  note that a new semantic segmentation algorithm $f_{J+1}$ can be easily added into the current MAD competition. The only additional work is to 1) select MAD subsets by maximizing the discrepancy between $f_{J+1}$ and $\mathcal{F}$, subject to the category and scale constraints, 2) collect ground-truth dense labels for the newly selected images, 3) expand $A$ and $R$ by one to accommodate new paired comparison results, and 4) update the two global ranking vectors (see Algorithm~\ref{alg:AddFramwork}).

\begin{table}[t]
\centering
\caption{
\label{tab:pascal}The object categories in the PASCAL VOC benchmark~\citep{Everingham2015}.}
\begin{tabular}{l|lllcc}
\toprule
 Category   & \\
\hline Person & person\\
         Animal & bird, cat, cow, dog, horse, sheep\\
         Vehicle & aero plane, bicycle, boat, bus, car, motorbike, train\\
         \multirow{2}{*}{\begin{tabular}[c]{@{}l@{}}Indoor\end{tabular}} & \multirow{2}{*}{\begin{tabular}[c]{@{}l@{}}bottle, chair, dining table, potted plant, sofa,\\ tv/monitor\end{tabular}}\\ \\
\bottomrule
\end{tabular}
\end{table}

\begin{table}[t]
\centering
\caption{
\label{tab:pascal_ratio}The first and third quartiles of the scale statistics for each category in PASCAL VOC~\citep{Everingham2015}.}
\begin{tabular}{l|ll|l|ll}
\toprule
Category &${T_\mathrm{min}}$& $T_\mathrm{max}$&Category& $T_\mathrm{min}$ & $T_\mathrm{max}$\\ \hline
aero plane  & 0.034    & 0.160  & bicycle     & 0.013   & 0.087   \\ \hline
bird        & 0.021    & 0.148  & boat        & 0.031   & 0.158   \\ \hline
bottle      & 0.006    & 0.148  & bus         & 0.167   & 0.450   \\ \hline
car         & 0.018    & 0.262  & cat         & 0.129   & 0.381   \\ \hline
chair       & 0.021    & 0.129  & cow         & 0.069   & 0.292   \\ \hline
potted plant & 0.079   & 0.286  & sheep       & 0.076   & 0.302   \\ \hline
sofa        & 0.082    & 0.284  & train       & 0.082   & 0.264   \\ \hline
tv/monitor  & 0.027    & 0.249  & dining table & 0.012  & 0.109   \\ \hline
dog         & 0.039    & 0.297  & horse       & 0.084   & 0.262   \\ \hline
motorbike   & 0.132    & 0.356  & person      & 0.028   & 0.199   \\
\bottomrule
\end{tabular}
\end{table}


\begin{figure*}[!ht]
  \includegraphics[width=1\linewidth]{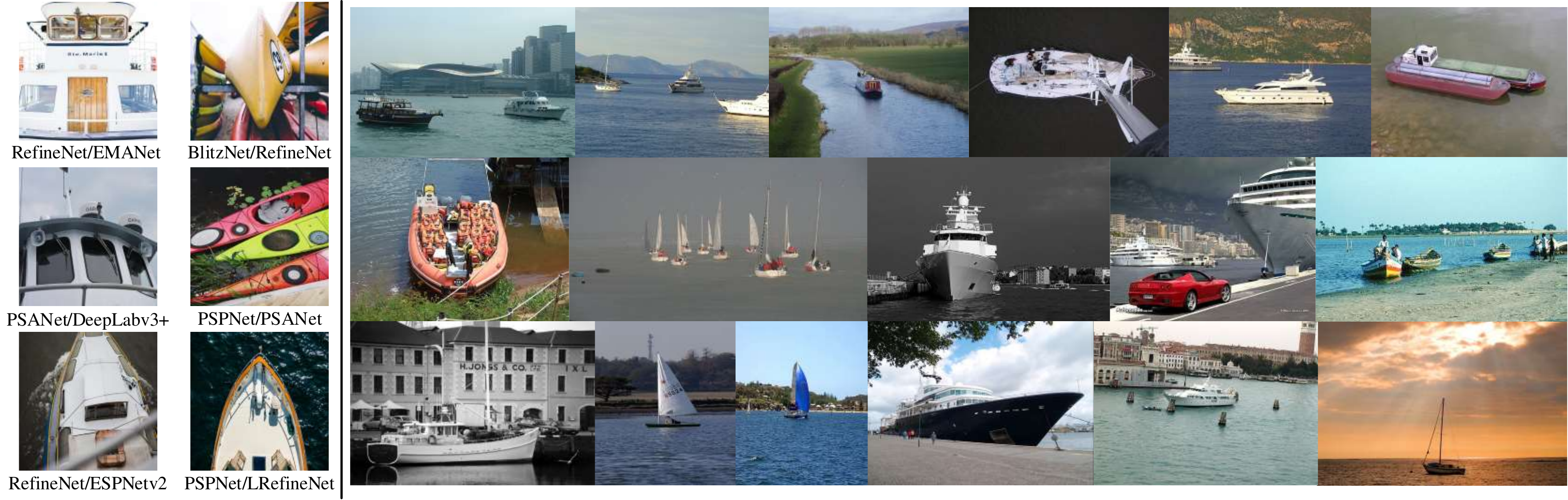}
\caption{``Boat'' images selected from  $\mathcal{D}$ by MAD (left panel) and from PASCAL VOC (right panel). Shown below each MAD image are the two associated  segmentation methods.}
\label{fig:boat}
\end{figure*}


\begin{figure*}[!ht]
  \includegraphics[width=1\linewidth]{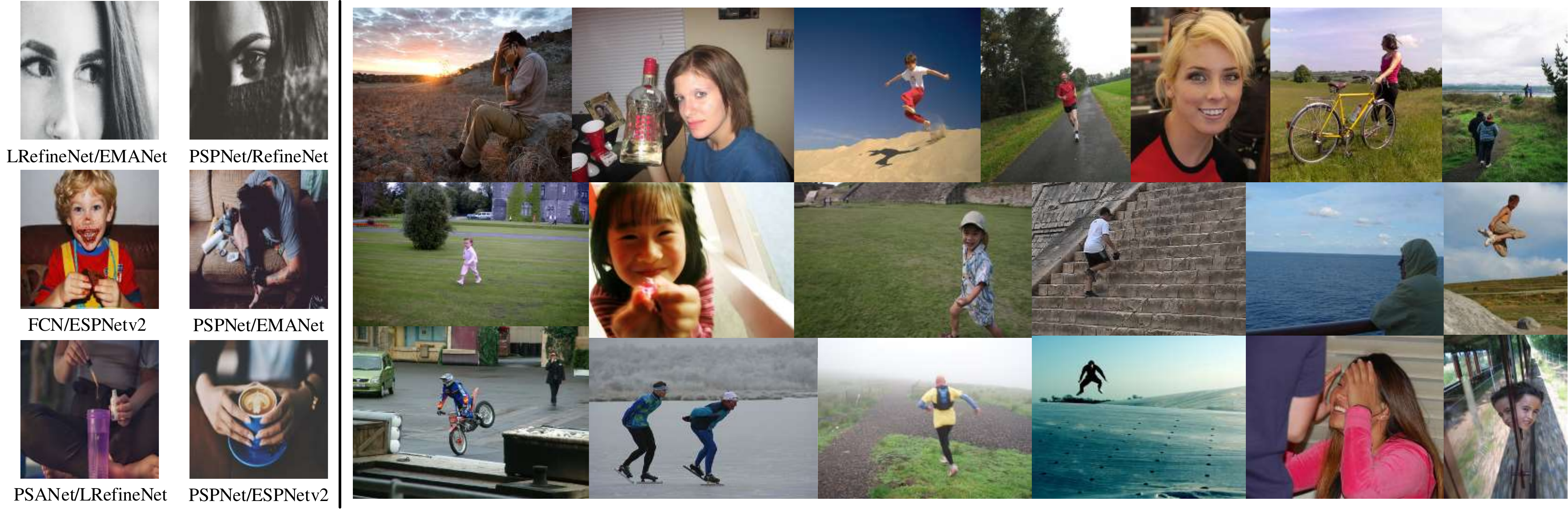}
\caption{``Person'' images selected from  $\mathcal{D}$ by MAD (left panel) and  from PASCAL VOC (right panel).}
\label{fig:person}
\end{figure*}

\section{Experiments}
In this section, we apply the MAD competition to compare ten semantic segmentation methods trained on PASCAL VOC \citep{Everingham2015}. We focus primarily on PASCAL VOC because 1) it is a well-established benchmark accompanied by many state-of-the-art methods with different design modules, therefore providing a comprehensive pictures of the segmentation methods' pros and cons; and 2) the performance on this benchmark seems to have saturated, therefore questioning demanding finer discrimination between segmentation methods.

\label{sec_experi}

\subsection{Experimental Setup}
\subsubsection{Construction of $\mathcal{D}$}
\label{subsec_data}
Before introducing the unlabeled image set $\mathcal{D}$, we first briefly review the PASCAL VOC benchmark~\citep{Everingham2015}. It contains $20$ object categories as listed in Table \ref{tab:pascal}. The first and third quartiles of the scale statistics for each category are given in Table \ref{tab:pascal_ratio}.  The number of training and validation images  are $1,464$ and $1,149$, respectively. We follow the construction of PASCAL VOC, and crawl web images using the same category keywords, their synonyms, as well as keyword combinations (\eg, person + bicycle). No data screening is needed during data crawling. In our experiment, we build $\mathcal{D}$ with more than $100,000$ images, which is substantially larger than the training and validation sets of PASCAL VOC. Some MAD images from $\mathcal{D}$ are shown in Figs.~\ref{fig:boat} and \ref{fig:person}, in contrast to images of the same categories in PASCAL VOC.

\subsubsection{Construction of $\mathcal{F}$}
\label{subsec_models}

In this study, we test ten CNN-based semantic segmentation models trained on PASCAL VOC~\citep{Everingham2015}, including FCN~\citep{Long2015}, PSPNet~\citep{Zhao2017}, RefineNet~\citep{LinGS2017}, BlitzNet~\citep{dvornik2017blitznet}, DeepLabv3+~\citep{Chen2018v3+}, LRefineNet~\citep{nekrasov2018light}, PSANet~\citep{zhao2018psanet}, EMANet~\citep{Lixia2019}, DiCENet~\citep{Mehta2019dicenet} and ESPNetv2~\citep{mehta2019espnetv2}, which are created from 2015 to 2019. Among the test methods, FCN is a pioneering work in deep semantic segmentation. LRefineNet is a faster version of RefineNet. Therefore, it is interesting to see the trade off between speed and accuracy under the MAD setting. It is also worth investigating the effects of several core modules, such as dilated convolution, skip connection, multi-scale computation, and attention mechanism to enhance model robustness when facing the open visual world. For all segmentation models, we use the publicly available implementations with pre-trained weights to guarantee the reproducibility of the results on PASCAL VOC (see Table~\ref{tab:summary}). In all experiments,  we resize the input image to $512\times512$ for inference.

\begin{figure}[t]
	\includegraphics[width=1\linewidth]{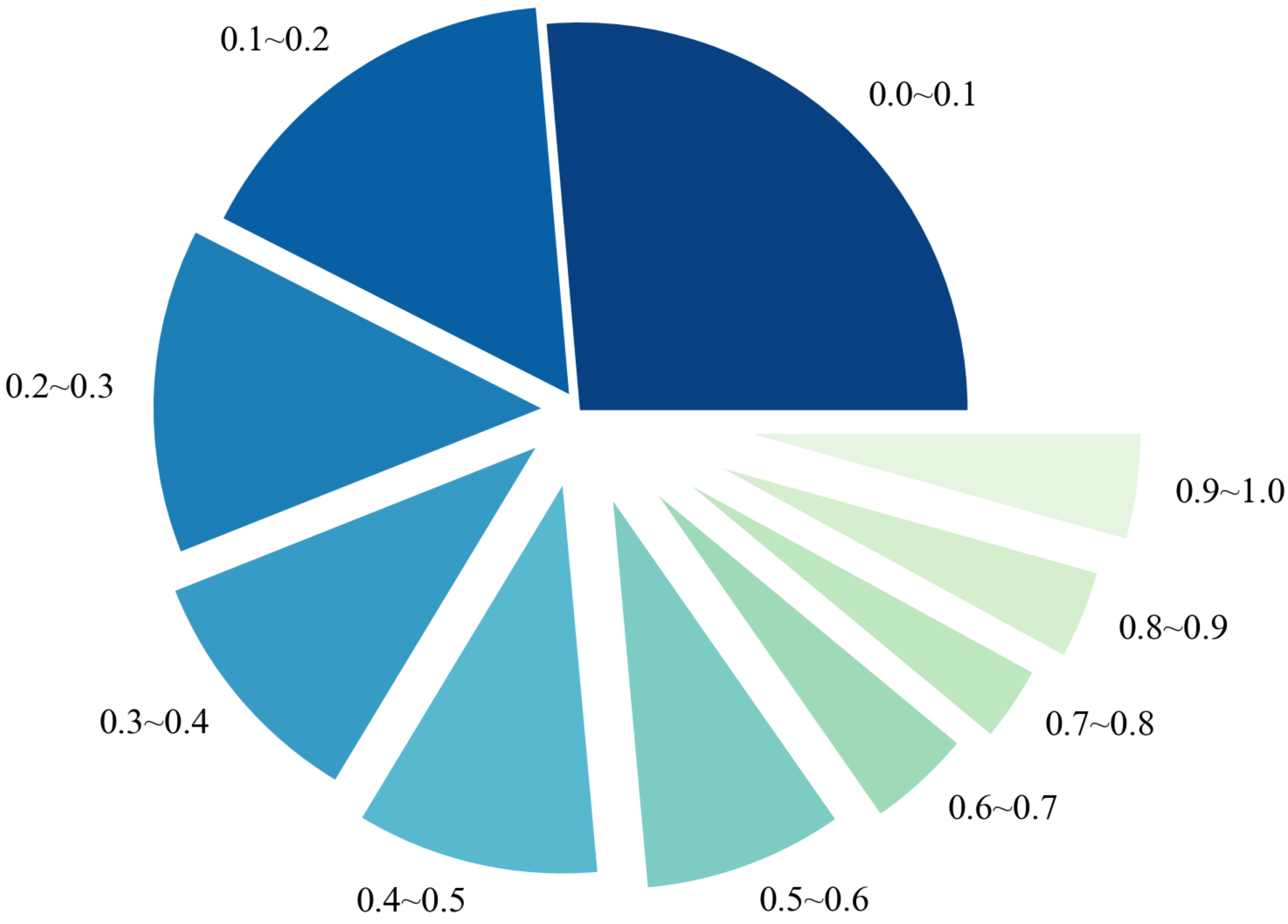}
	\caption{The pie chart of the mIoU values of all ten segmentation methods on the MAD set $\mathcal{S}$.}
	\label{fig:pie_mIoU}
\end{figure}

\subsubsection{Construction of $\mathcal{S}$}\label{subsec:sss}
To construct $\mathcal{S}$, we need to specify the concordance metric in Eq. \eqref{eq:disc}. In the main experiments, we adopt mIoU, which is defined as
\begin{align}
	\label{eq:mIoU}
	\mathrm{mIoU} = \frac{1}{\vert\mathcal{Y}\vert+1}\sum_{y=0}^{\vert\mathcal{Y}\vert}{\frac{N_{yy}}{\sum_{y'=0}^{\vert\mathcal{Y}\vert}N_{yy'}+\sum_{y'=0}^{\vert\mathcal{Y}\vert}N_{y'y}-N_{yy}}},
\end{align}
where the mean is taken over all object categories, and
\begin{align}\label{eq:Nyy}
    N_{yy'} = \sum_{n=1}^N \mathbb{I}[f_i(x_n) = y \cap f_j(x_n) = y'],
\end{align}
where $\mathbb{I}[\cdot]$ denotes the indicator function. In Section \ref{subsec_ablation}, we will investigate the sensitivity of the results when we switch to other metrics commonly used in semantic segmentation, such as mean accuracy.

\begin{figure}[!ht]
	\includegraphics[width=1\linewidth]{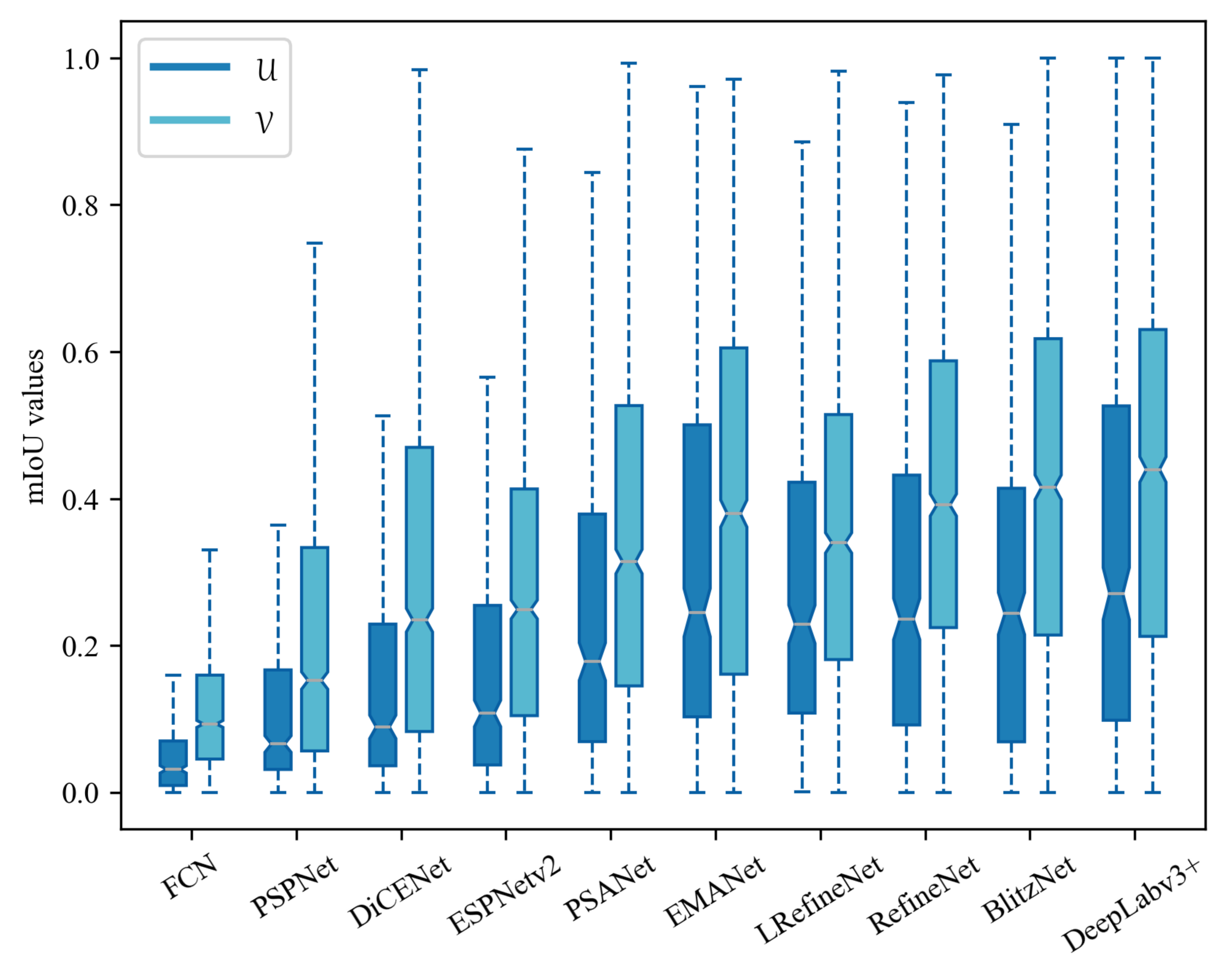}
	\caption{The box-plot of the mIoU values of individual segmentation methods. We evaluate $f_i$ on two disjoint and complementary subsets $\mathcal{U}_i$ and $\mathcal{V}_i$ of $\mathcal{S}$, where $\mathcal{U}_i$ and $\mathcal{V}_i$ include images that are associated with and irrelevant to $f_i$, respectively. The gray bar represents the median mIoU. The lower and upper boundaries of the box indicate the first and third quartiles, respectively. The minimum and maximum data are also shown as the endpoints of the dashed line.}
	\label{fig:boxplot}
\end{figure}

As no data cleaning is involved in the construction of $\mathcal{D}$, it is possible that the selected MAD images  fall out of the $20$ object categories (\eg, elephants), or are unnatural (\eg, cartoons). In this paper, we restrict our study to test the generalizability of semantic segmentation models to subpopulation shift \citep{santurkar2020breeds}, and only include natural photographic images containing the same $20$ object categories. This is done in strict adherence to the data preparation guideline of PASCAL VOC~\citep{Everingham2015}. As will be clear in Section \ref{subsec_diagnosis}, this seemingly benign setting already poses a grand challenge to the competing models. Considering the expensiveness of subjective testing, we set $K=1$, which corresponds to selecting the optimal image to discriminate between each ordered pair of segmentation models in each object category. After removing repetitions,  $\mathcal{S}$ includes a total of $833$ images.

\subsubsection{Subjective Annotation}
\label{subsec_subjective}
 We invite seven participants to densely label the MAD images with the help of the graphical image annotation tool - LabelMe\footnote{\url{https://github.com/wkentaro/labelme}}. To familiarize participants with the labeling task, we include a training session to introduce general knowledge about semantic segmentation and the manual of LabelMe. As suggested in PASCAL VOC, we perform two rounds of subjective experiments. The first round of experiment is broken into multiple sessions. In each session, participants are required to annotate $20$ images, followed by a short break to reduce the fatigue effect.  Participants can discuss with each other freely, which contributes to the annotation consistency. In the second round, we carry out cross validation to further improve the consistency of the annotated data. Each participant takes turn to check the segmentation maps completed by others, marking  positions and types of possible annotation errors. During this process,  participants can also communicate with each other. After sufficient discussions among participants, disagreement on a small portion of annotations is thought to be aligned.

\begin{figure*}[!ht]
  \includegraphics[width=1\linewidth]{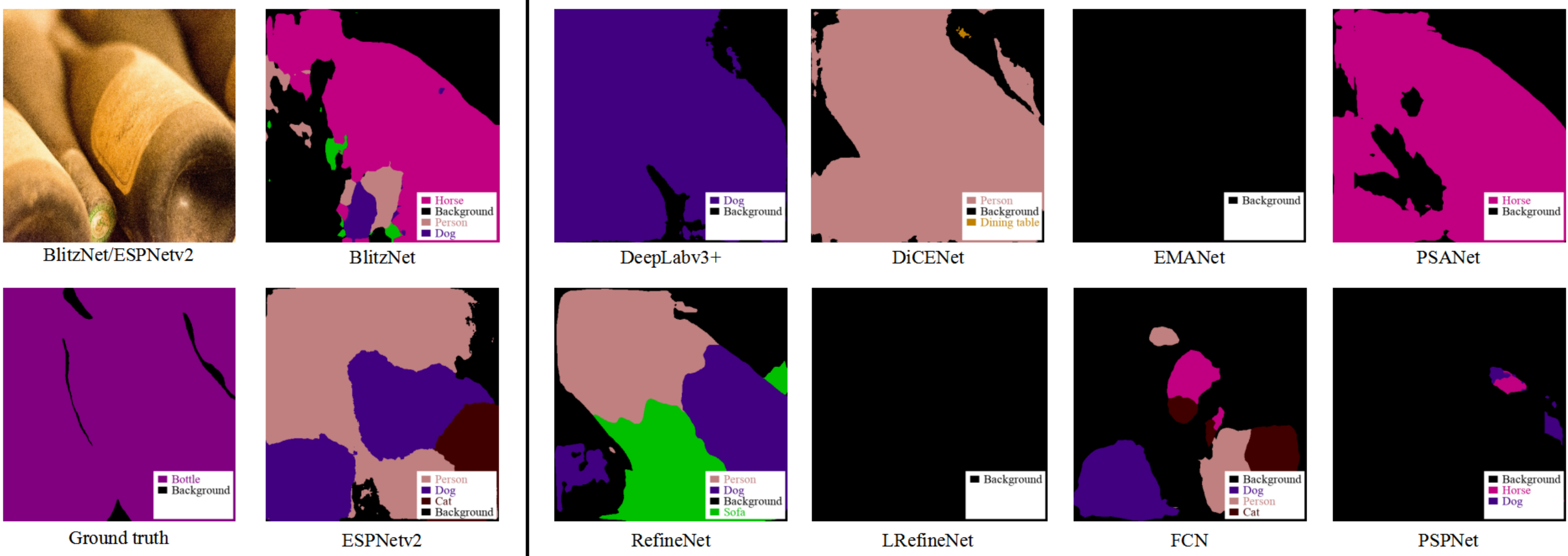}
\caption{An illustration of MAD failure transferability. The natural photographic image in the left panel is selected by MAD to best differentiate BlitzNet \citep{dvornik2017blitznet} and ESPNetv2 \citep{mehta2019espnetv2}. Interestingly, it is also able to falsify other competing models in dramatically different ways, as shown in the right panel.}
\label{fig:transferability}
\end{figure*}

\begin{figure}[!ht]
\includegraphics[width=1\linewidth]{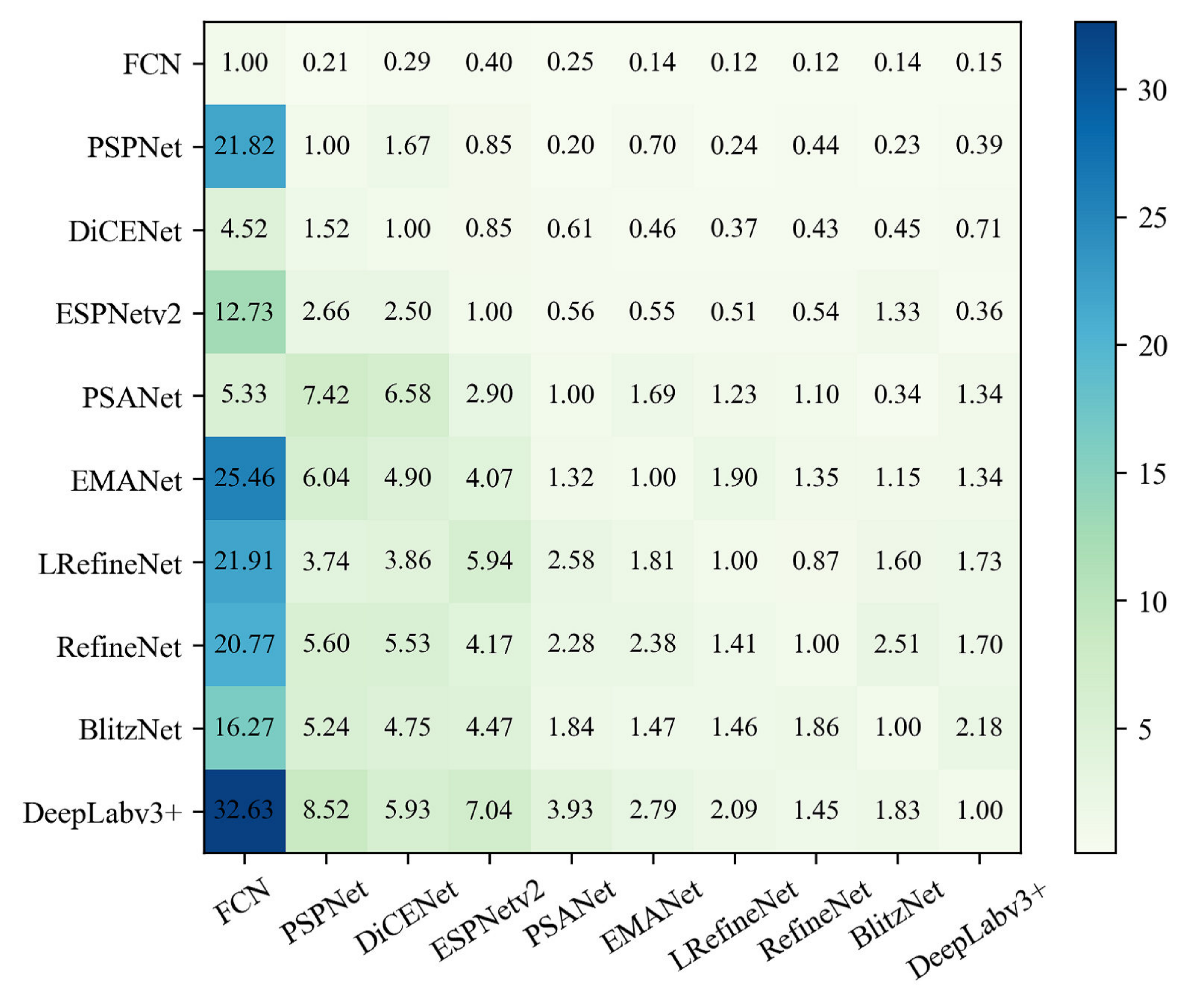}
\caption{The aggressiveness matrix $A$ computed by Eq. \eqref{eq:agg} to indicate how aggressive the row model is to falsify the column model. A larger value with cooler color indicates stronger aggressiveness.}
\label{fig:aggressiveness}
\end{figure}

\begin{figure}[!ht]
\includegraphics[width=1\linewidth]{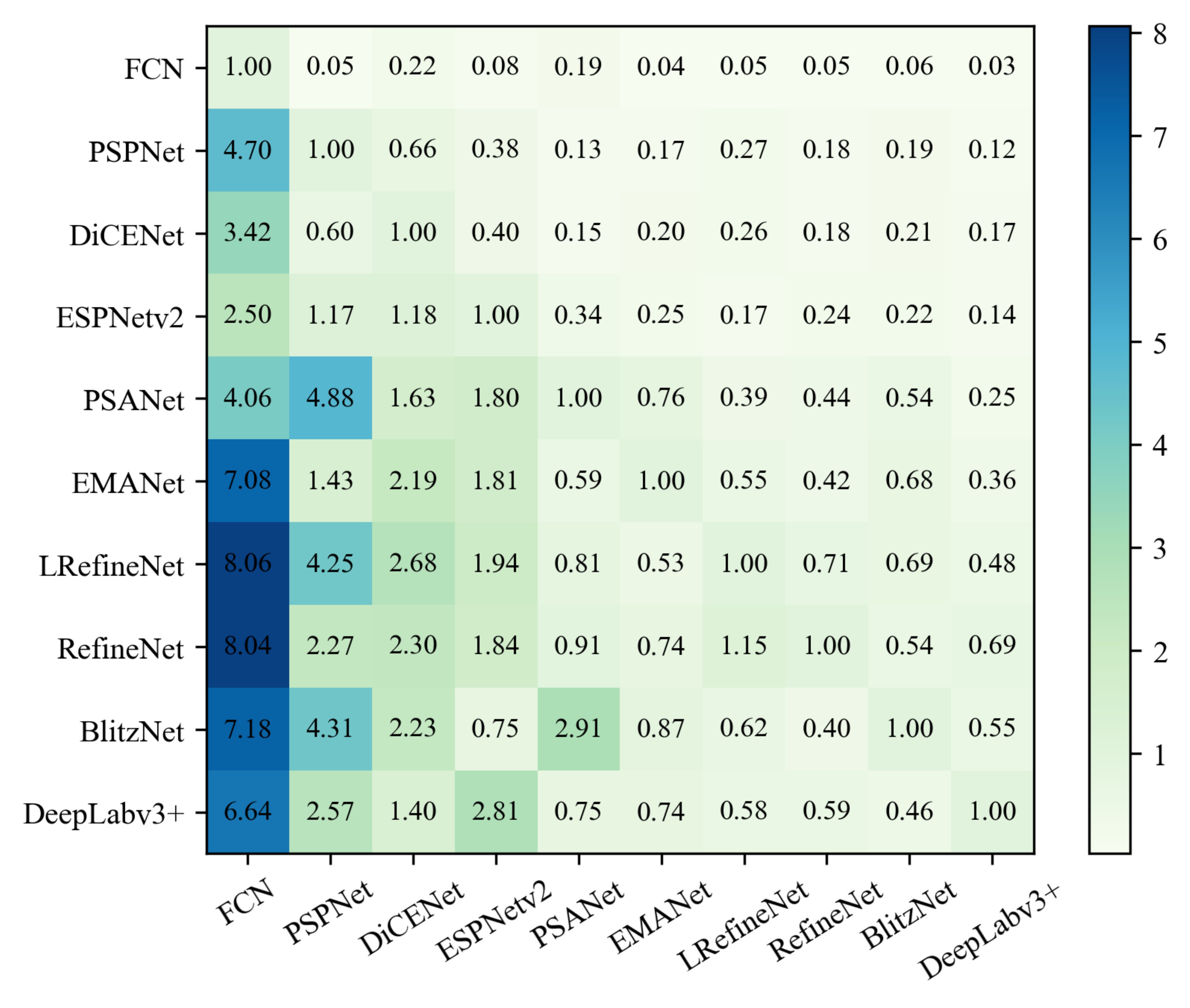}
\caption{The resistance matrix $R$ computed by Eq. \eqref{eq:res} to indicate how resistant the row model is under attack by the column model. A larger value with cooler color indicates stronger resistance.}
\label{fig:resistance}
\end{figure}

\subsection{Main Results}
\label{subsec_diagnosis}
We first get an overall impression on  the model generalizability by putting together the mIoU values obtained by all ten segmentation methods on the MAD set $\mathcal{S}$ in Fig. \ref{fig:pie_mIoU}. It is clear that  MAD  successfully exposes failures of these CNN-based methods, leading to a substantial performance drop on $\mathcal{S}$. More precisely, over $84.7\%$ of the mIoU values are smaller than $0.6$, while less than $8\%$ exceed $0.8$ (see Fig.~\ref{fig:pie_mIoU}). This is in stark contrast to the pronounced results on PASCAL VOC, where all ten methods achieve an mIoU value greater than $0.6$ (see Table~\ref{tab:summary}). We next take a closer look at individual performance. For each segmentation method $f_i$, we divide $\mathcal{S}$ into two disjoint subsets $\mathcal{U}_i$ and $\mathcal{V}_i$, where images in $\mathcal{U}_i$ and $\mathcal{V}_i$ are associated with and irrelevant to $f_i$, respectively. We evaluate $f_i$ on the two subsets, and summarize the mIoU results in Fig. \ref{fig:boxplot}, from which we have two insightful observations. First, the performance of $f_i$ on $\mathcal{U}_i$ for all $i = \{1,2,\ldots, J\}$ is marginal, indicating that a significant portion of the MAD images are double-failure cases of the two associated segmentation methods (\ie, falling in Case \uppercase\expandafter{\romannumeral3}). Second, the performance of $f_i$ on $\mathcal{V}_i$ is only slightly better than that on $\mathcal{U}_i$, suggesting that the MAD images are highly transferable to falsify other segmentation models (see Fig. \ref{fig:transferability}). This coincides with the strong transferability of adversarial examples (often with imperceptible  perturbations) found in image classification. In summary, MAD has identified automatically a set of naturally occurring hard examples from the open visual world that  significantly degrade segmentation accuracy.

\begin{table*}[!h]
	\centering
	\caption{
		\label{tab:summary}The performance comparison of ten semantic segmentation models. $A$ and $R$ represent aggressiveness and resistance, respectively.}
	\begin{tabular}{l|l|cccccclllll}
		\toprule Model                              & Backbone & mIoU  & mIoU rank & $A$ rank & $\Delta A$ rank & $R$ rank & $\Delta R$ rank\\
		\hline
		DeepLabv3+~\citep{Chen2018v3+}        & Xception   & 0.890 & 1         & 1 & 0            & 1 & 0 \\
		EMANet~\citep{Lixia2019}             & ResNet-152 & 0.882 & 2         & 5 & -3           & 5 & -3 \\
		PSANet~\citep{zhao2018psanet}        & ResNet-101 & 0.857 & 3         & 6 & -3           & 6 & -3 \\
		PSPNet~\citep{Zhao2017}              & ResNet-101 & 0.854 & 4         & 8 & -4           & 9 & -5 \\
		LRefineNet~\citep{nekrasov2018light} & ResNet-152 & 0.827 & 5         & 4 & +1           & 4 & +1 \\
		RefineNet~\citep{LinGS2017}          & ResNet-101 & 0.824 & 6         & 2 & +4           & 3 & +3 \\
		BlitzNet~\citep{dvornik2017blitznet} & ResNet-50  & 0.757 & 7         & 3 & +4           & 2 & +5 \\
		ESPNetv2~\citep{mehta2019espnetv2}   &      -     & 0.680 & 8         & 7 & +1           & 7 & +1 \\
		DiCENet~\citep{Mehta2019dicenet}     &      -     & 0.673 & 9         & 9 & 0            & 8 & +1 \\
		FCN~\citep{Long2015}                 & VGG16      & 0.622 & 10        & 10 & 0           & 10 & 0 \\
		\bottomrule
	\end{tabular}
\end{table*}

We continue to analyze the aggressiveness and resistance of the test methods when they compete with each other in MAD. We first show the aggressiveness and resistance matrices, $A$ and $R$, in Figs.~\ref{fig:aggressiveness} and~\ref{fig:resistance}, respectively. A higher value in each entry with cooler color represents stronger aggressiveness/resistance of the corresponding row model against the column model. From the figures, we make two interesting observations. First, it is relatively easier for a segmentation method to attack other methods than to survive the attack from others, as indicated by the larger values in $A$ than those in $R$. Second, a segmentation method with stronger aggressiveness generally shows stronger resistance, which is consistent with the finding in the context of image quality assessment \citep{ma2018group}. We next aggregate paired comparisons into global rankings, and show the results in Table~\ref{tab:summary}. We have a number of interesting findings, which are not apparently drawn from the high mIoU numbers on PASCAL VOC.

First, although BlitzNet \citep{dvornik2017blitznet} focuses on improving the computational efficiency of semantic segmentation with below-the-average mIoU performance on PASCAL VOC, it performs remarkably well in comparison to other models in MAD. A distinct characteristic of BlitzNet is that it employs object detection as an auxiliary task for joint multi-task learning. It is widely acceptable that object detection and semantic segmentation are closely related. Early researchers \citep{leibe2004combined} used detection to provide a good initialization for segmentation. In the context of end-to-end optimization, we conjecture that the detection task would regularize the segmentation task to learn better localized features for more accurate segmentation boundaries. Another benefit of this multi-task setting is that BlitzNet can learn from a larger number of images through the detection task, with labels much cheaper to obtain than the segmentation task.

\begin{figure*}[!h]
	\centering
	\subfigure[Appearance.]{
		\begin{minipage}[]{0.96\linewidth}
			\includegraphics[width=0.12\linewidth]{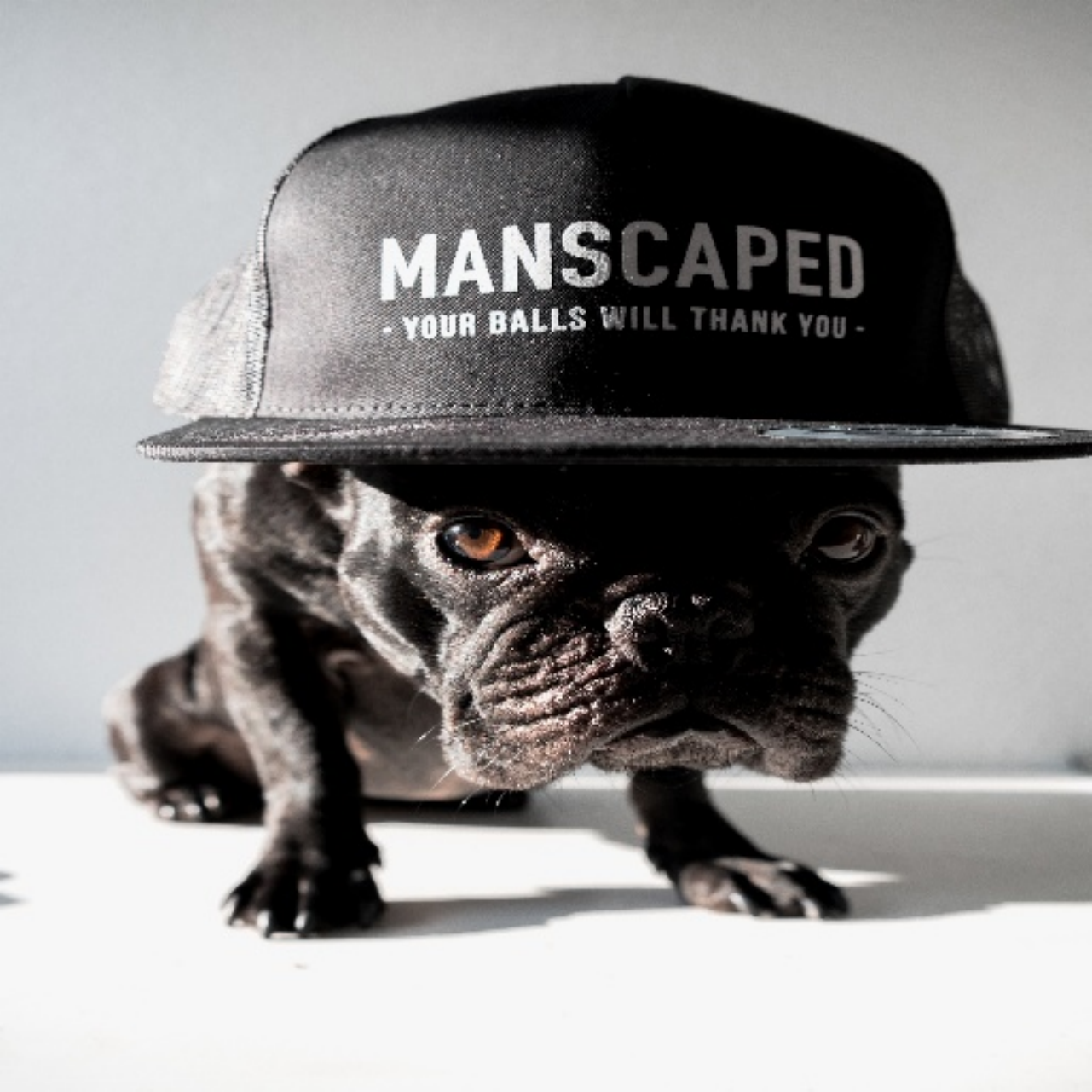}\vspace{0.51pt}
			\includegraphics[width=0.12\linewidth]{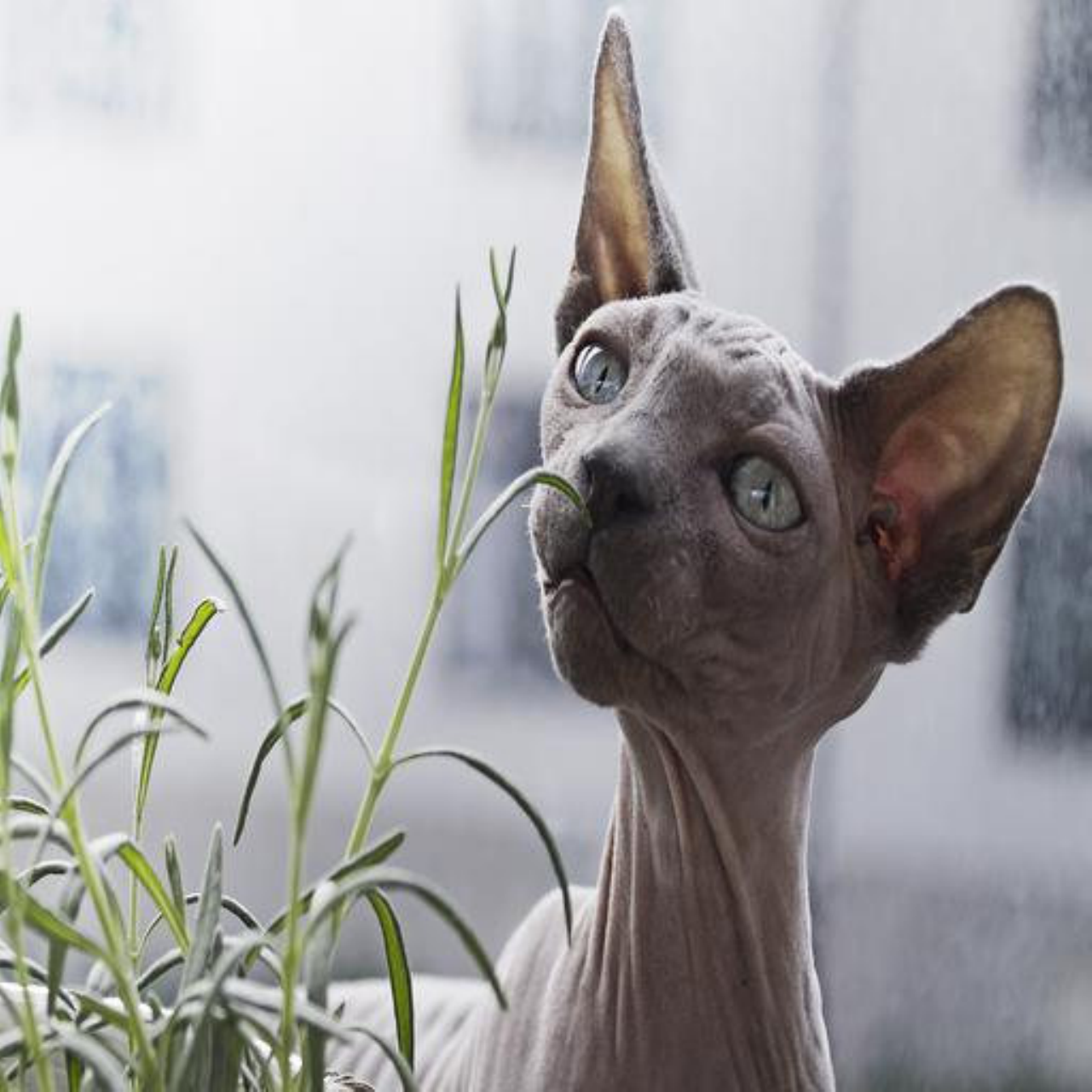}\vspace{0.51pt}
			\includegraphics[width=0.12\linewidth]{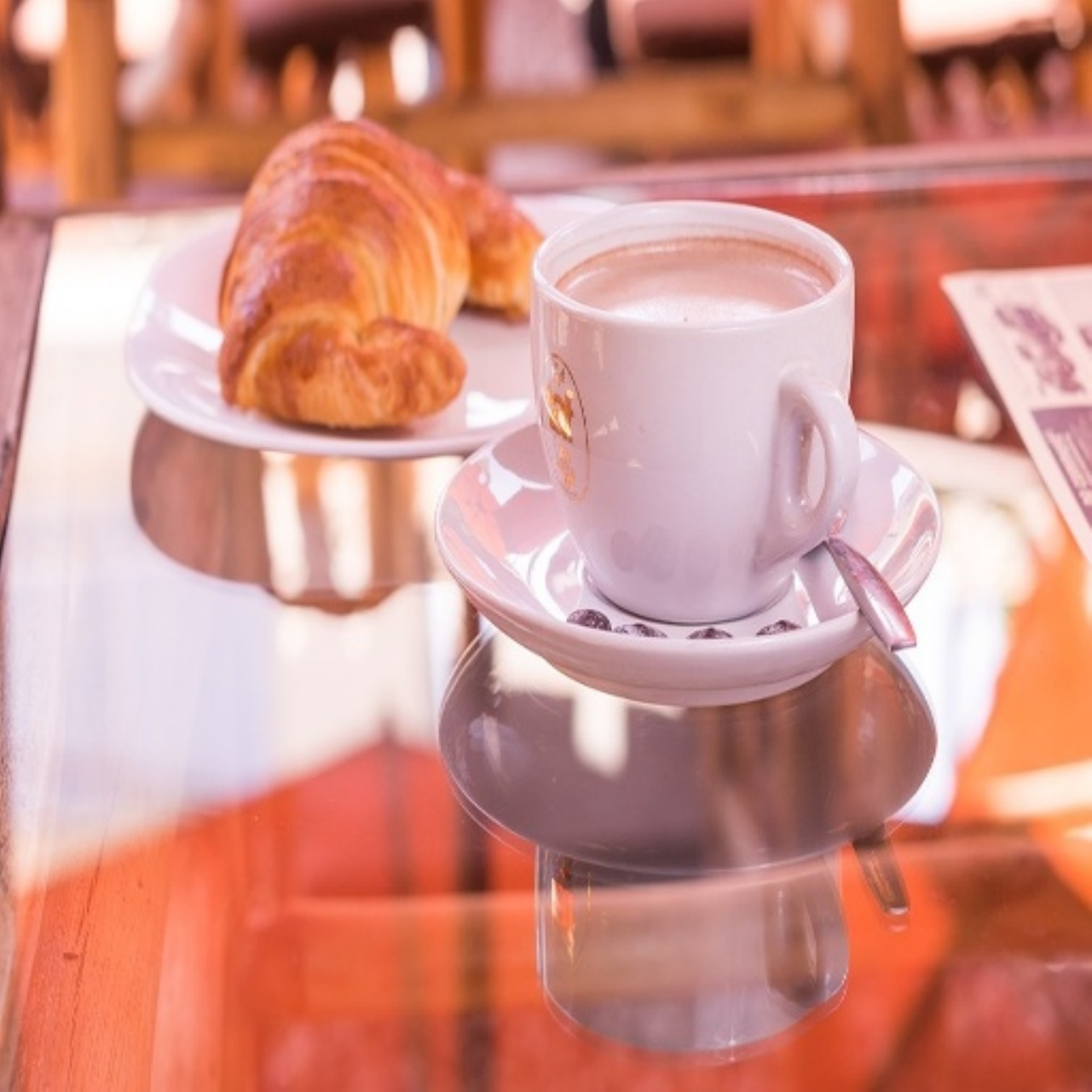}\vspace{0.51pt}
			\includegraphics[width=0.12\linewidth]{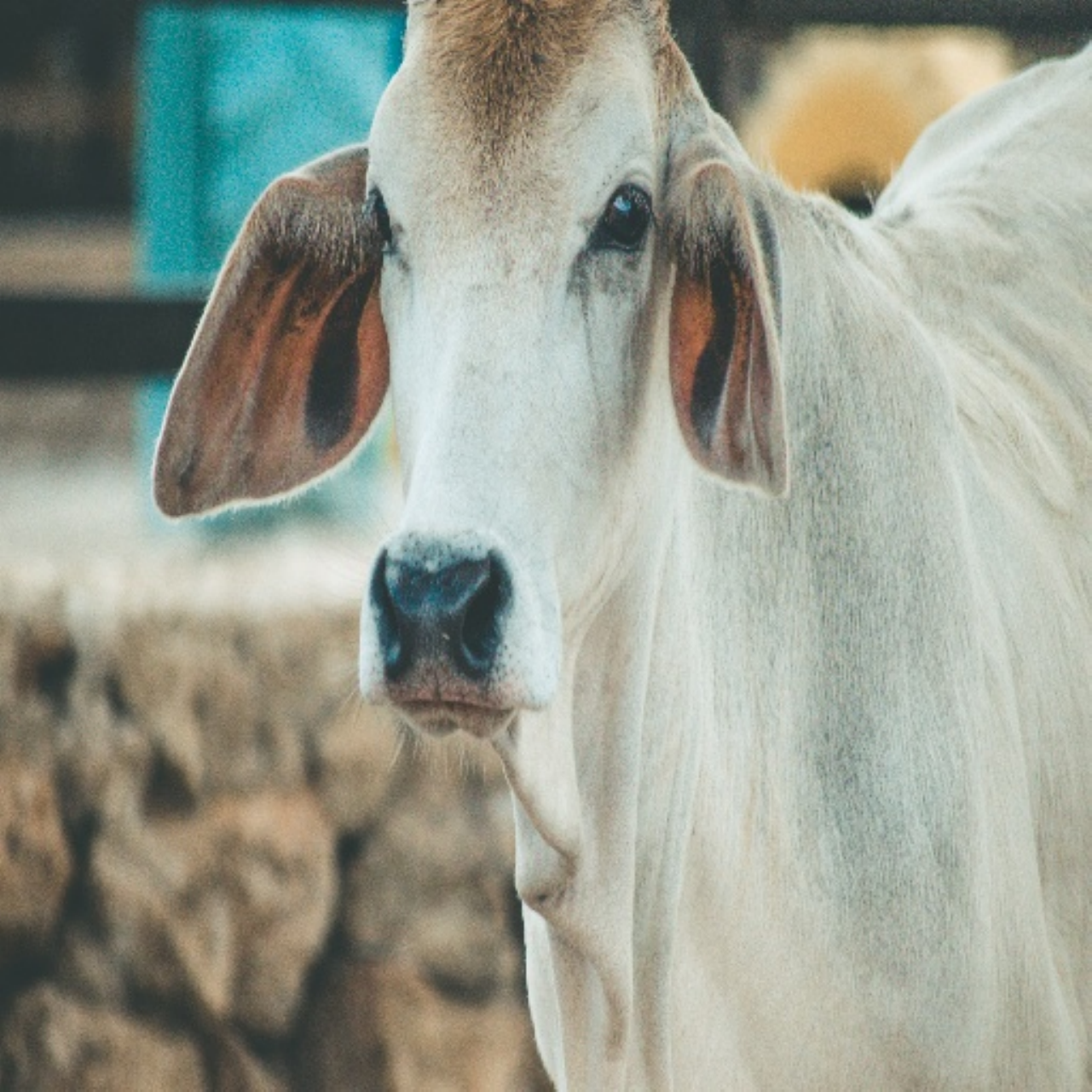}\vspace{0.51pt}
			\includegraphics[width=0.12\linewidth]{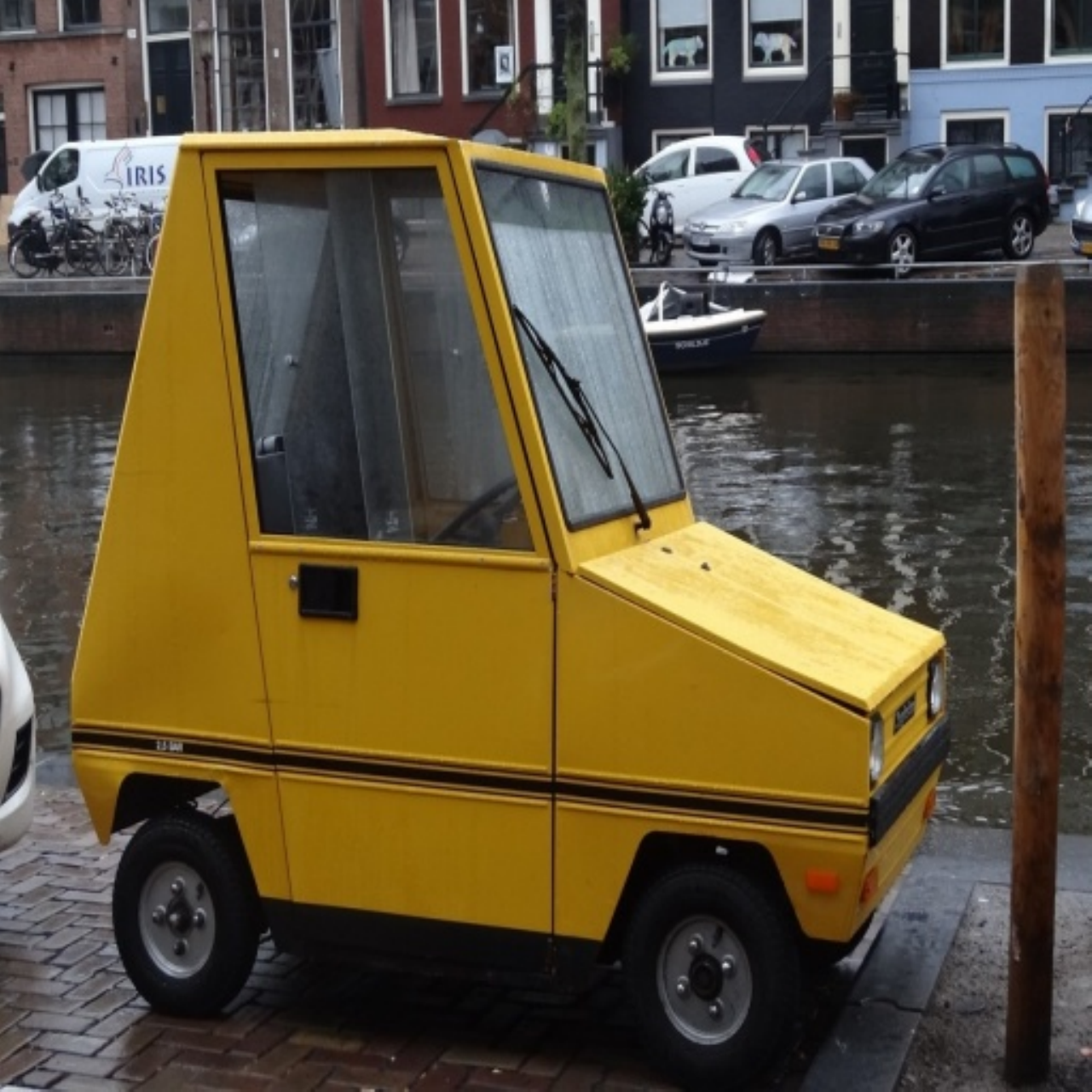}\vspace{0.51pt}
			\includegraphics[width=0.12\linewidth]{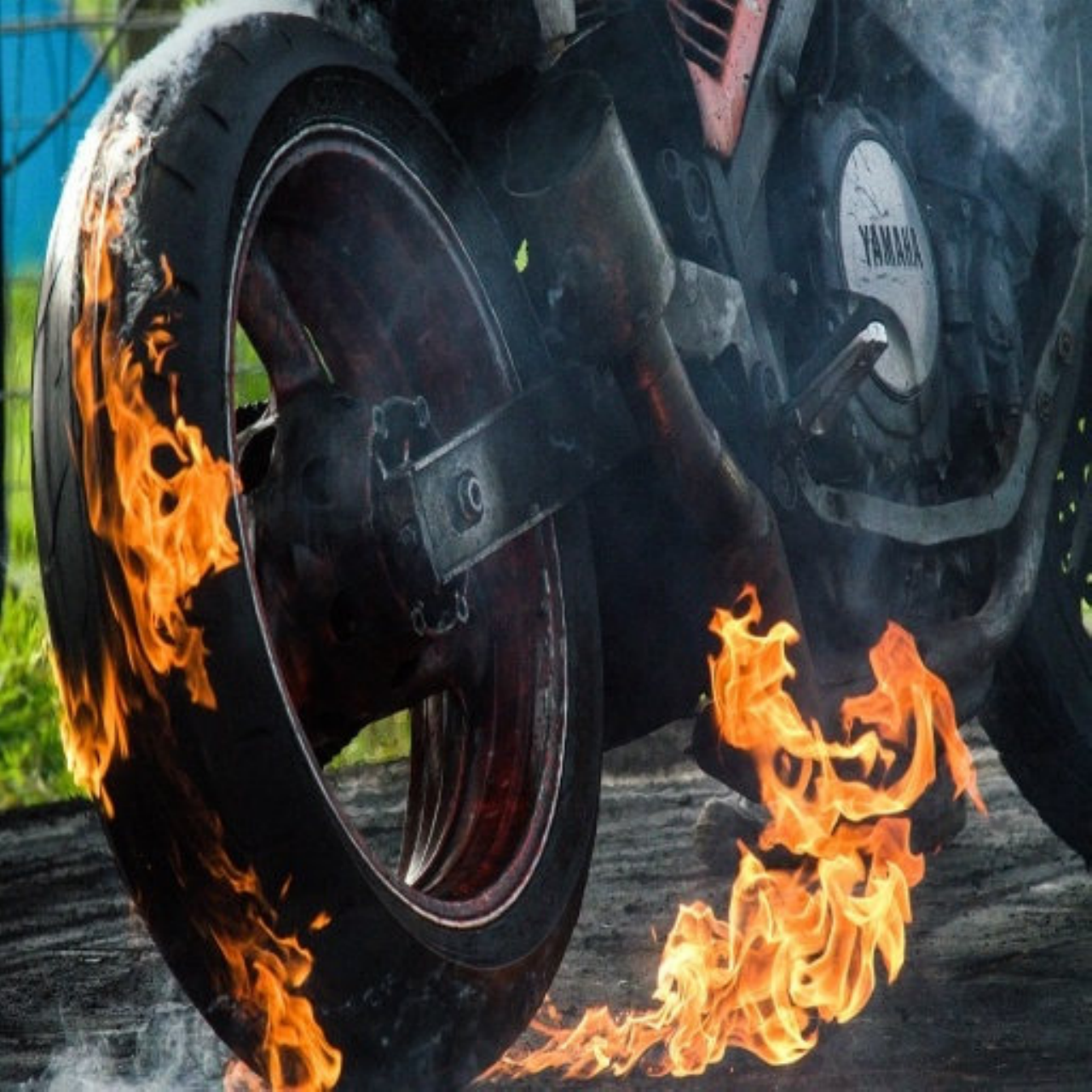}\vspace{0.51pt}
			\includegraphics[width=0.12\linewidth]{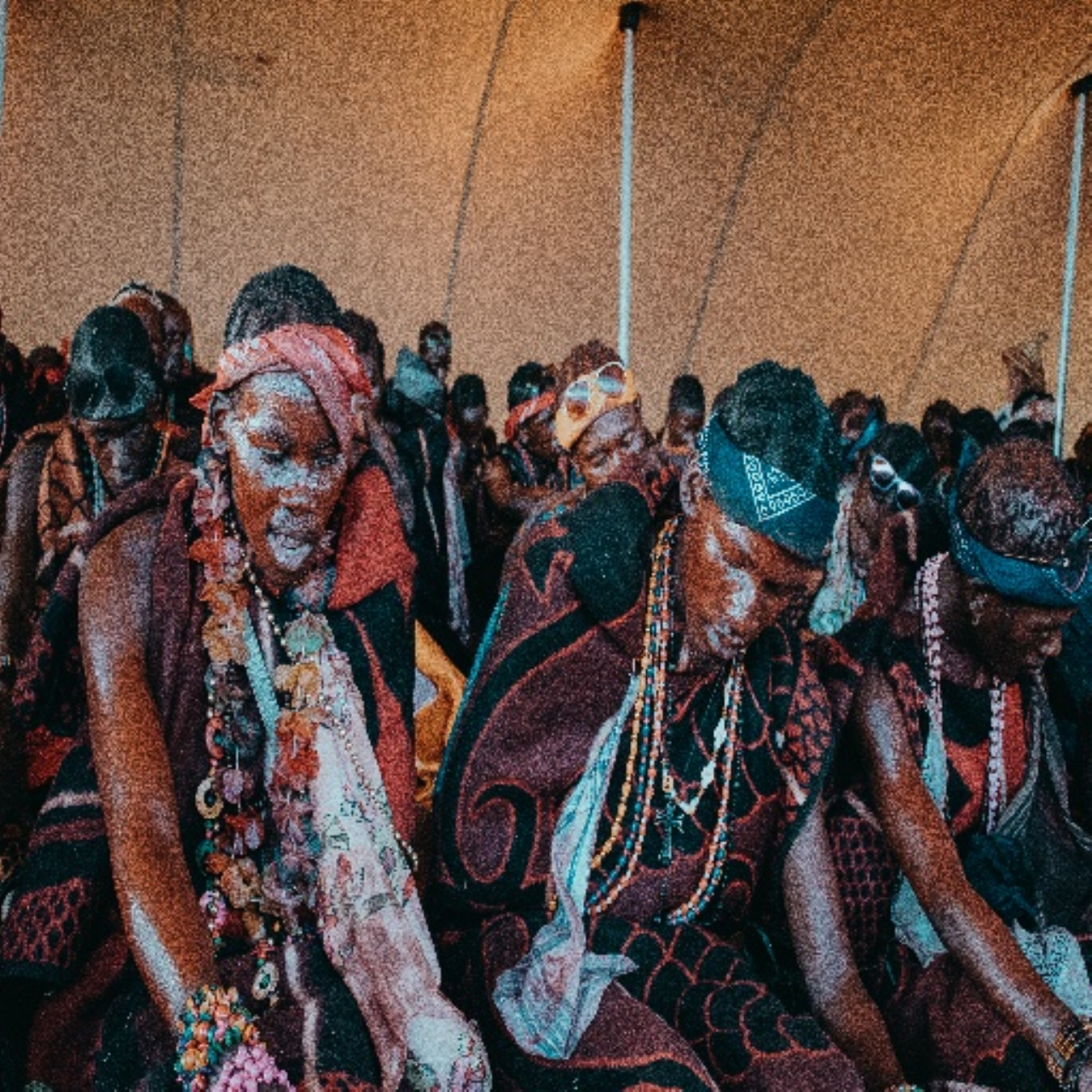}\vspace{0.51pt}
			\includegraphics[width=0.12\linewidth]{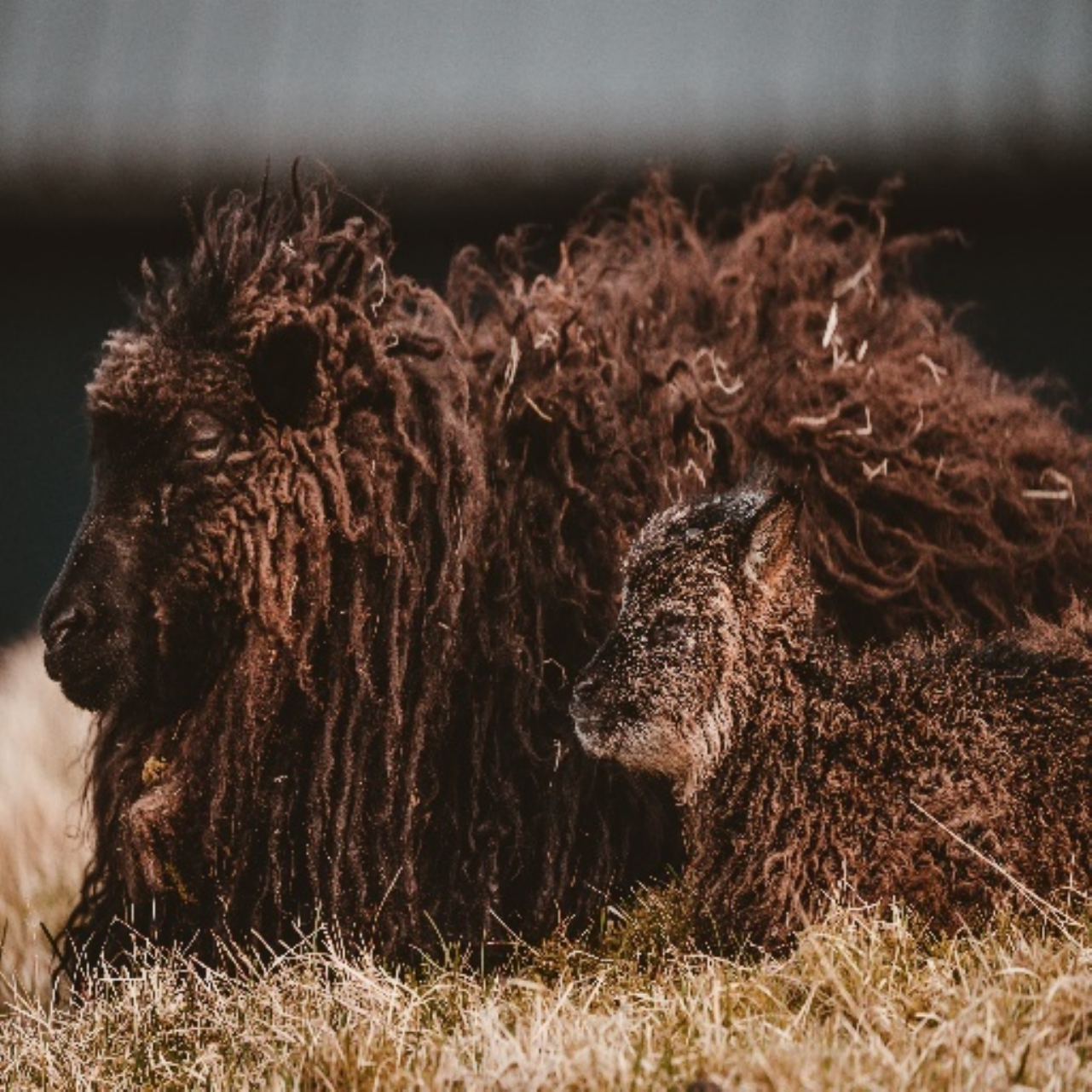}\vspace{0.51pt}
	\end{minipage}}
	\subfigure[Viewpoint.]{
		\begin{minipage}[]{0.96\linewidth}
			\includegraphics[width=0.12\linewidth]{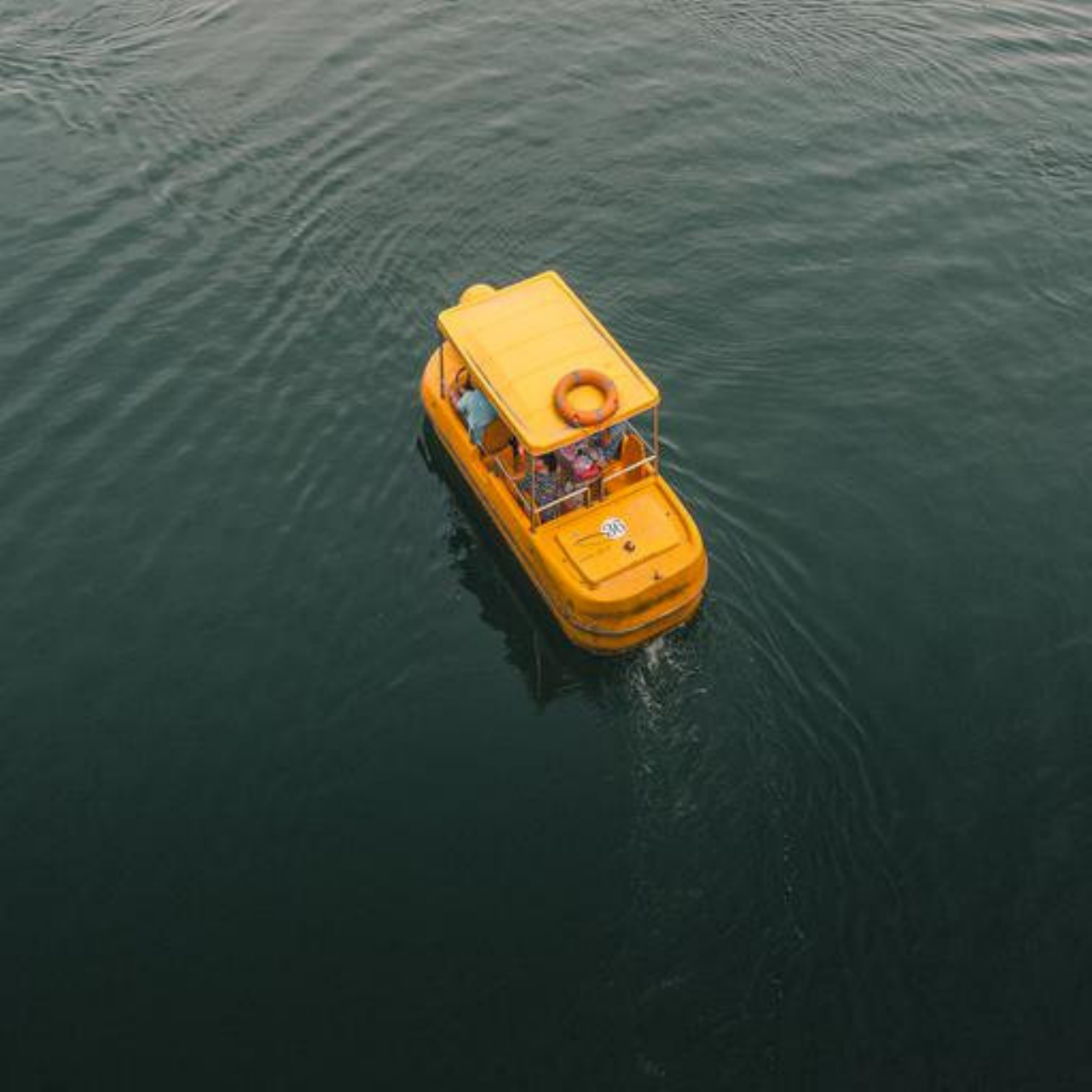}\vspace{0.51pt}
			\includegraphics[width=0.12\linewidth]{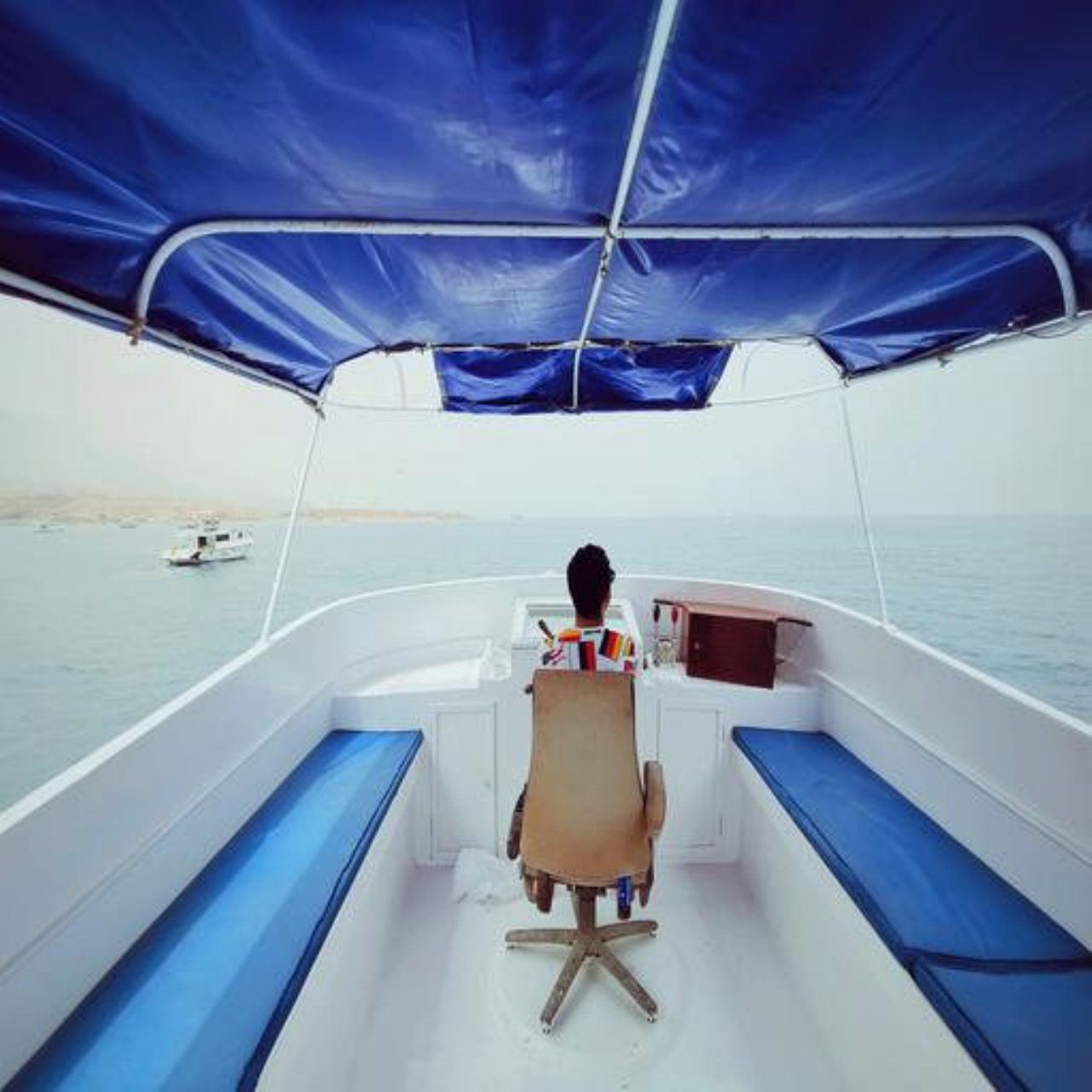}\vspace{0.51pt}
			\includegraphics[width=0.12\linewidth]{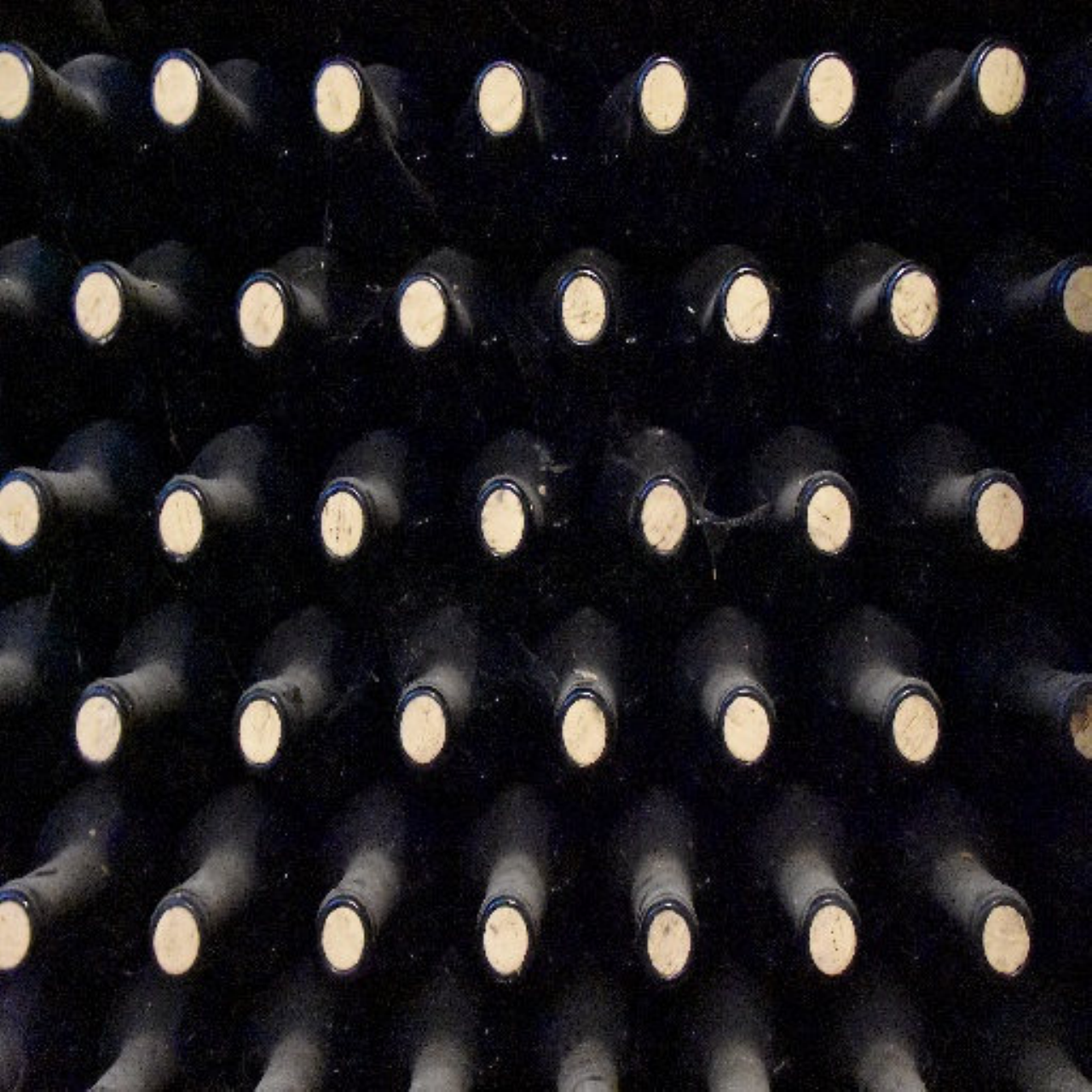}\vspace{0.51pt}
			\includegraphics[width=0.12\linewidth]{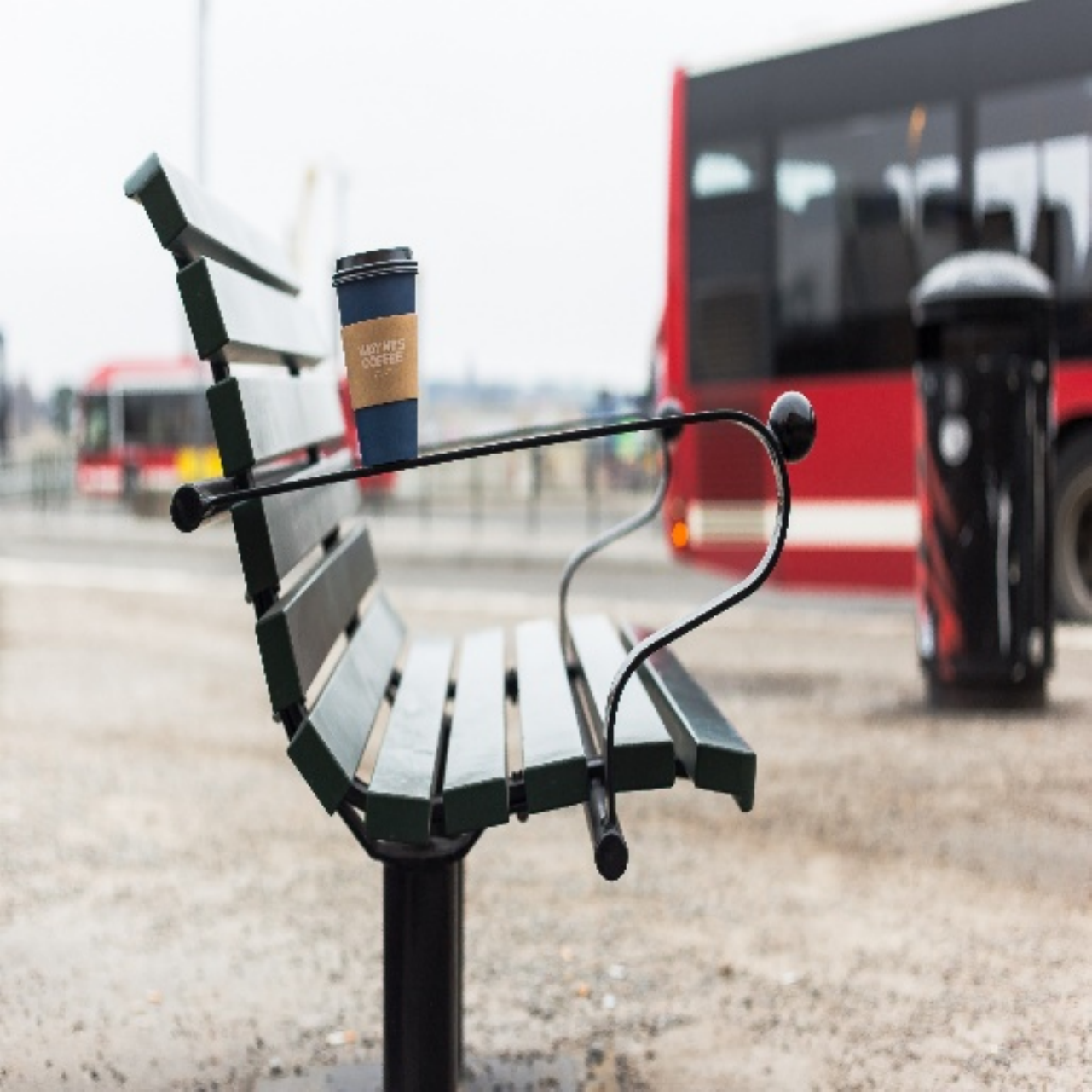}\vspace{0.51pt}
			\includegraphics[width=0.12\linewidth]{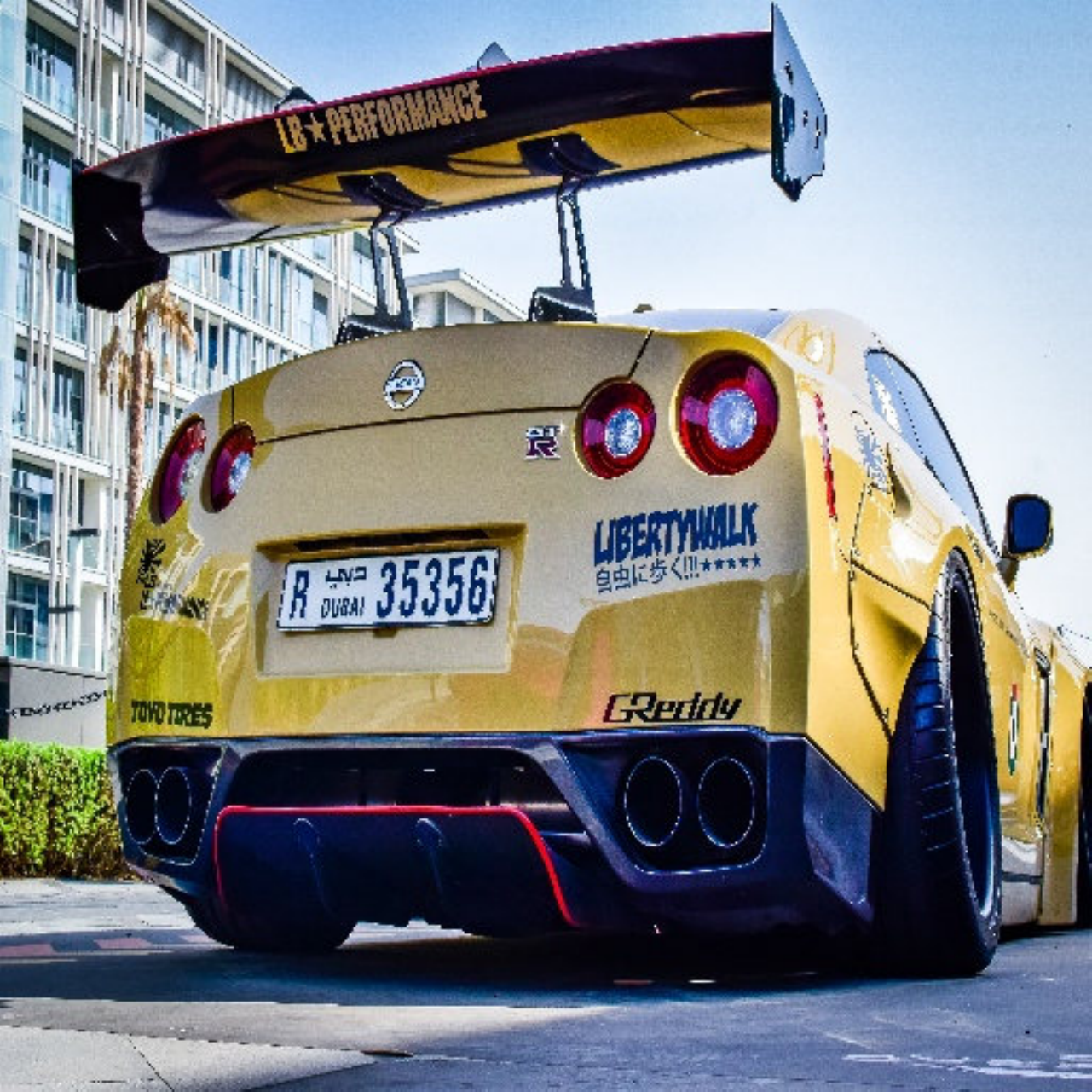}\vspace{0.51pt}
			\includegraphics[width=0.12\linewidth]{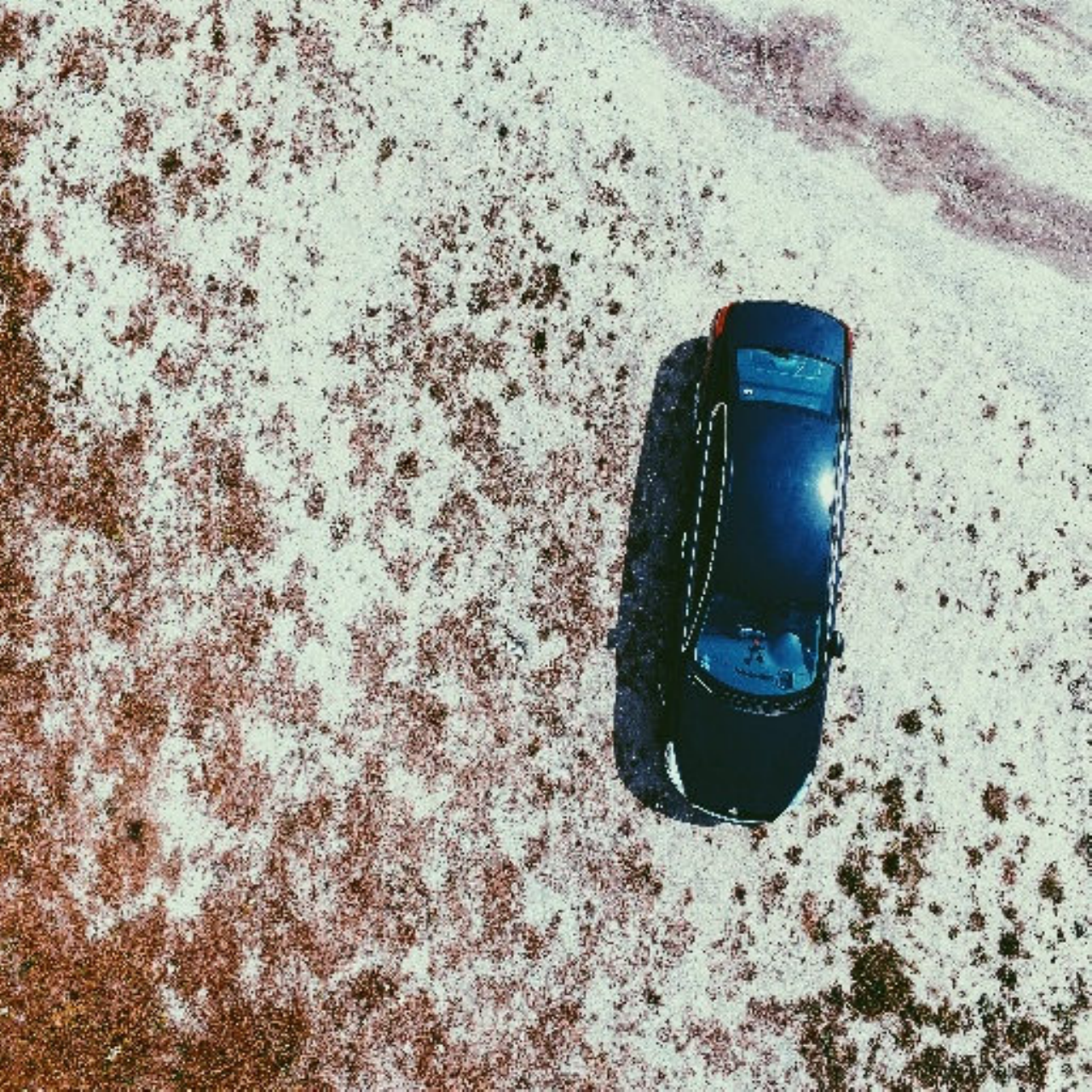}\vspace{0.51pt}
			\includegraphics[width=0.12\linewidth]{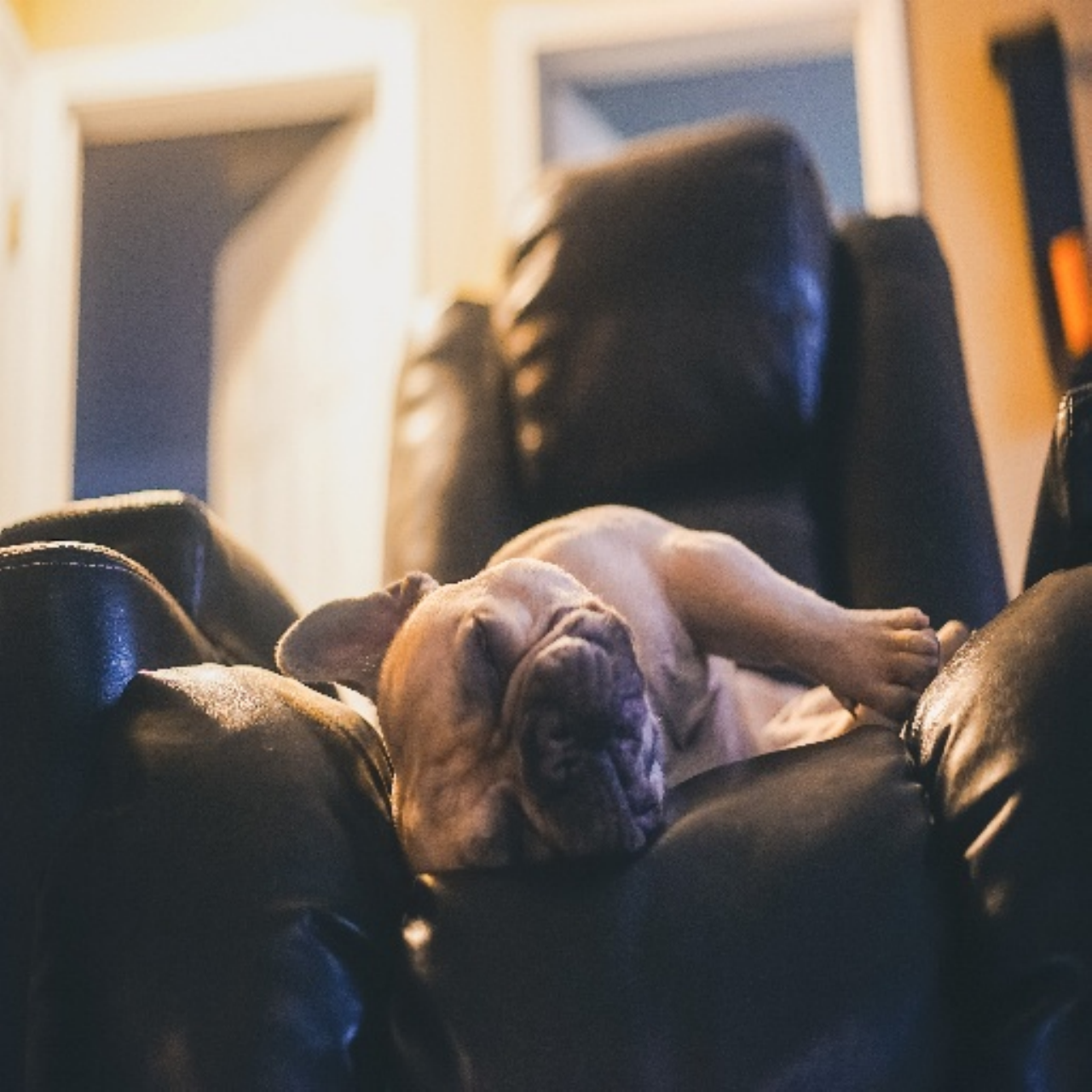}\vspace{0.51pt}
			\includegraphics[width=0.12\linewidth]{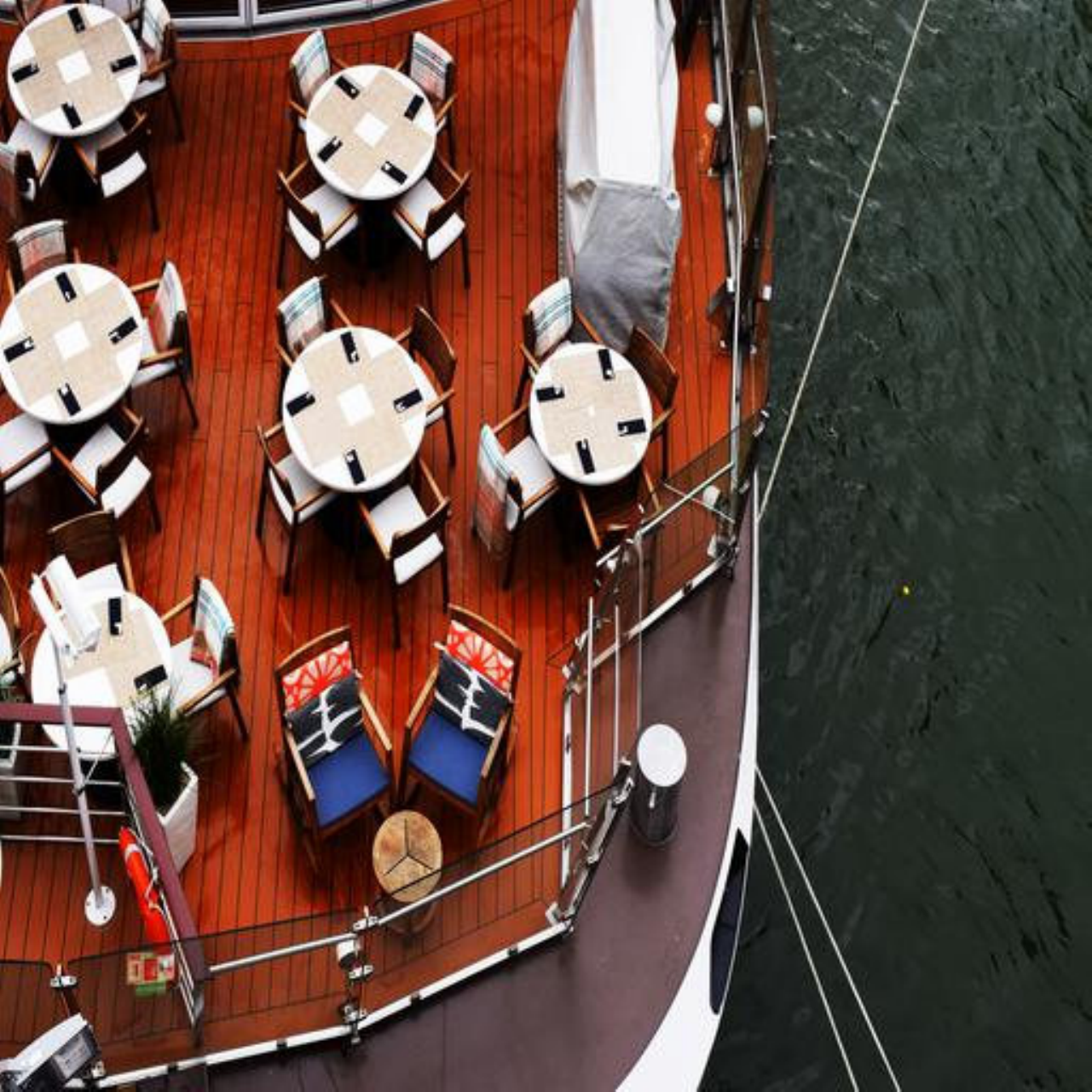}\vspace{0.51pt}
	\end{minipage}}\\
	\subfigure[Object scale.]{
		\begin{minipage}[]{0.96\linewidth}
			\includegraphics[width=0.12\linewidth]{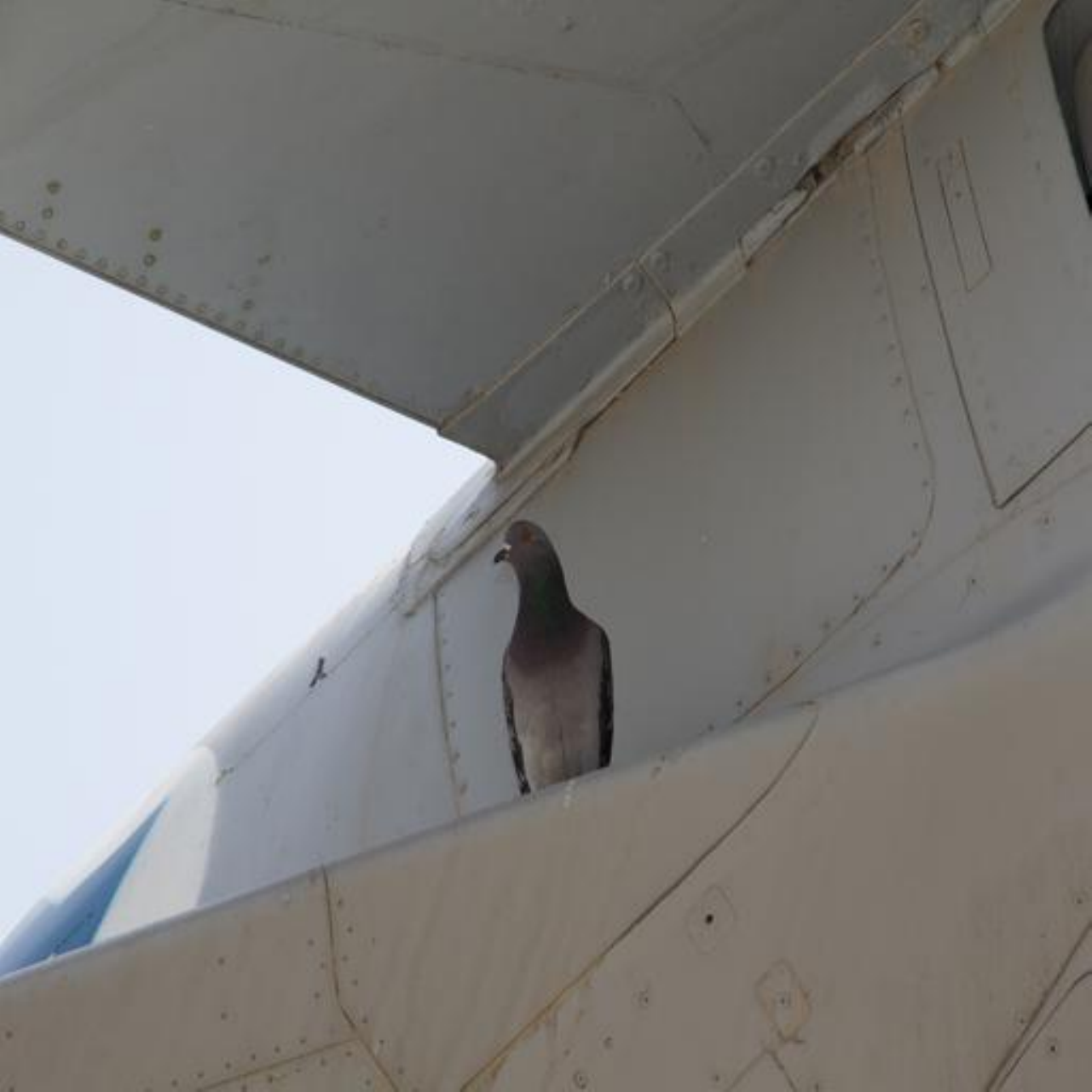}\vspace{0.51pt}
			\includegraphics[width=0.12\linewidth]{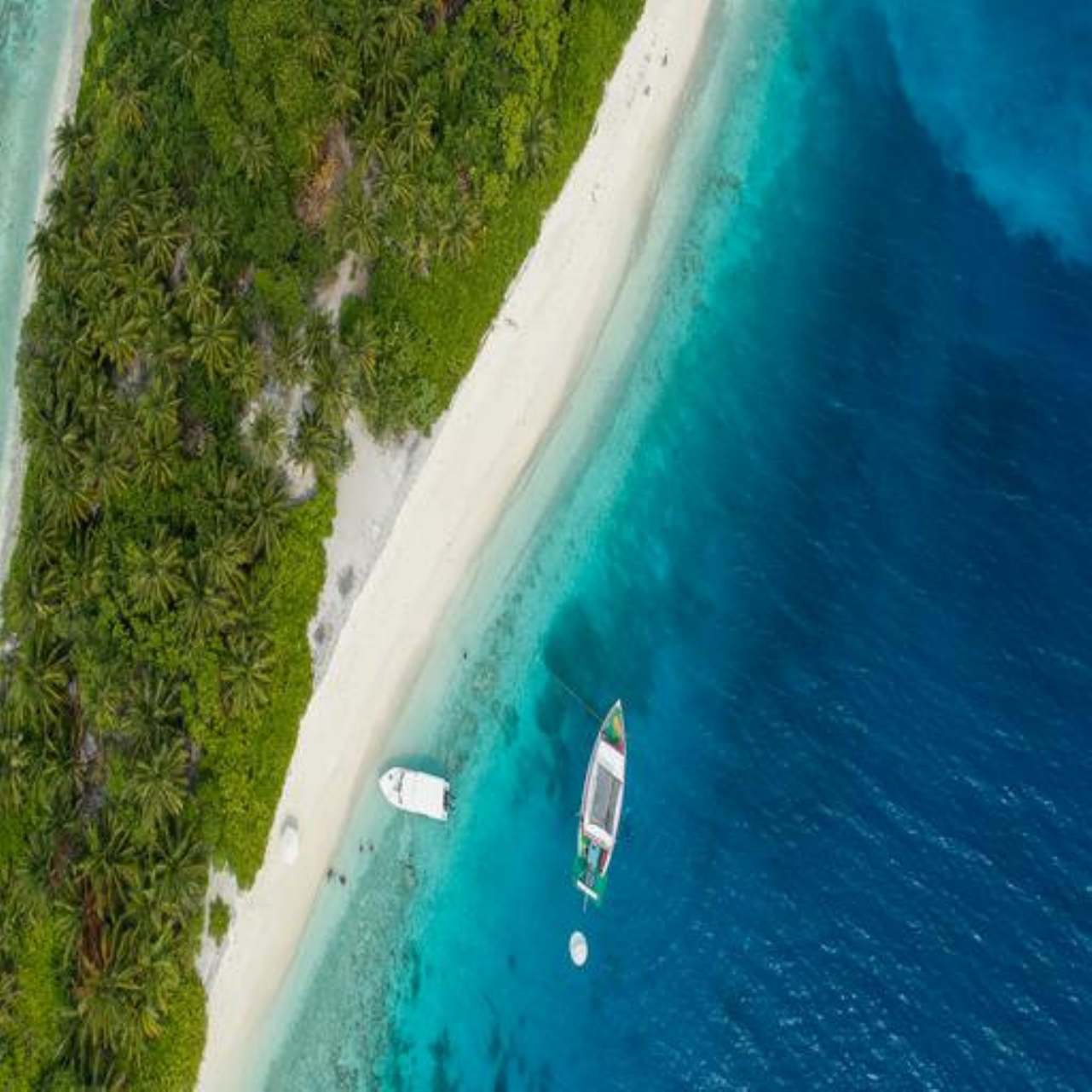}\vspace{0.51pt}
			\includegraphics[width=0.12\linewidth]{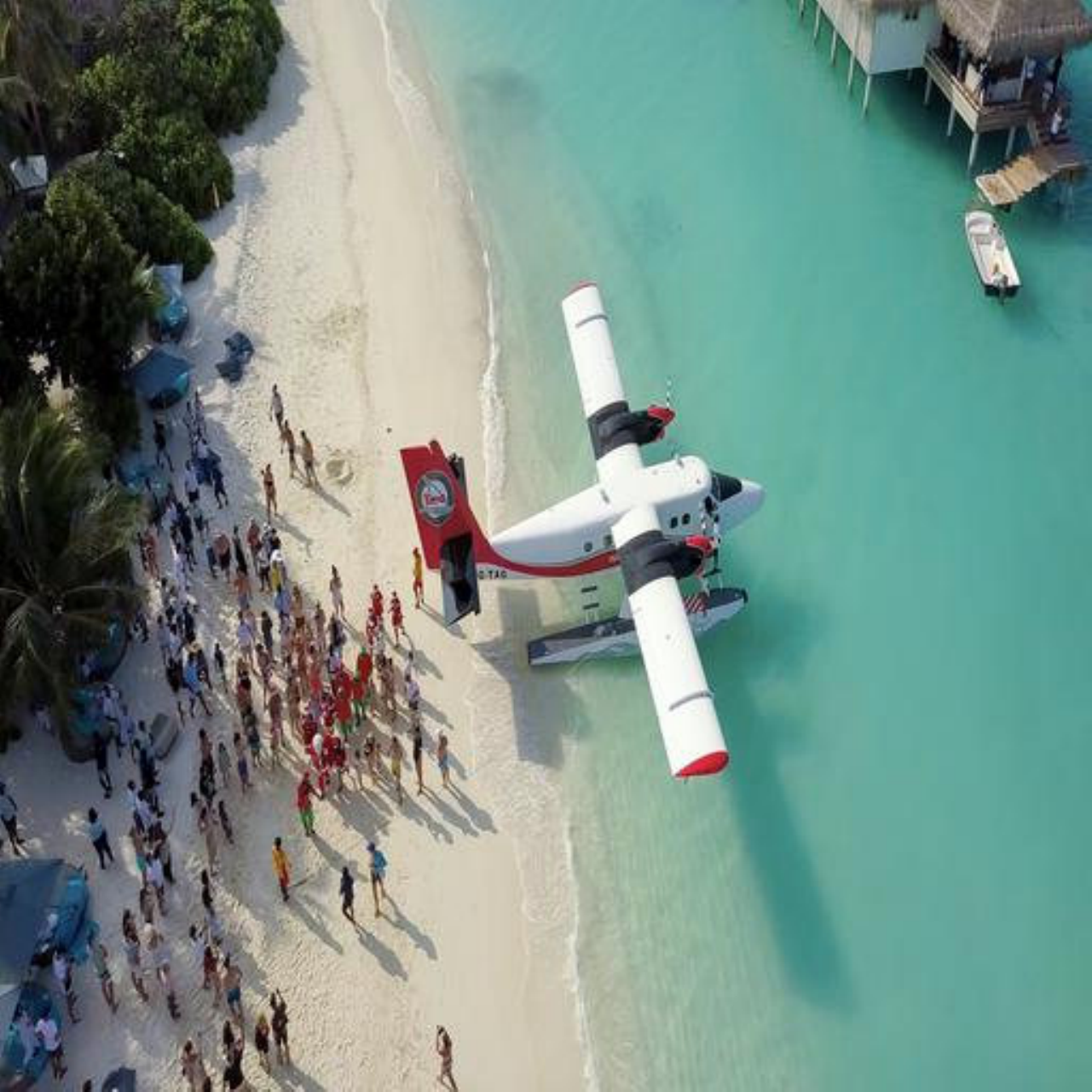}\vspace{0.51pt}
			\includegraphics[width=0.12\linewidth]{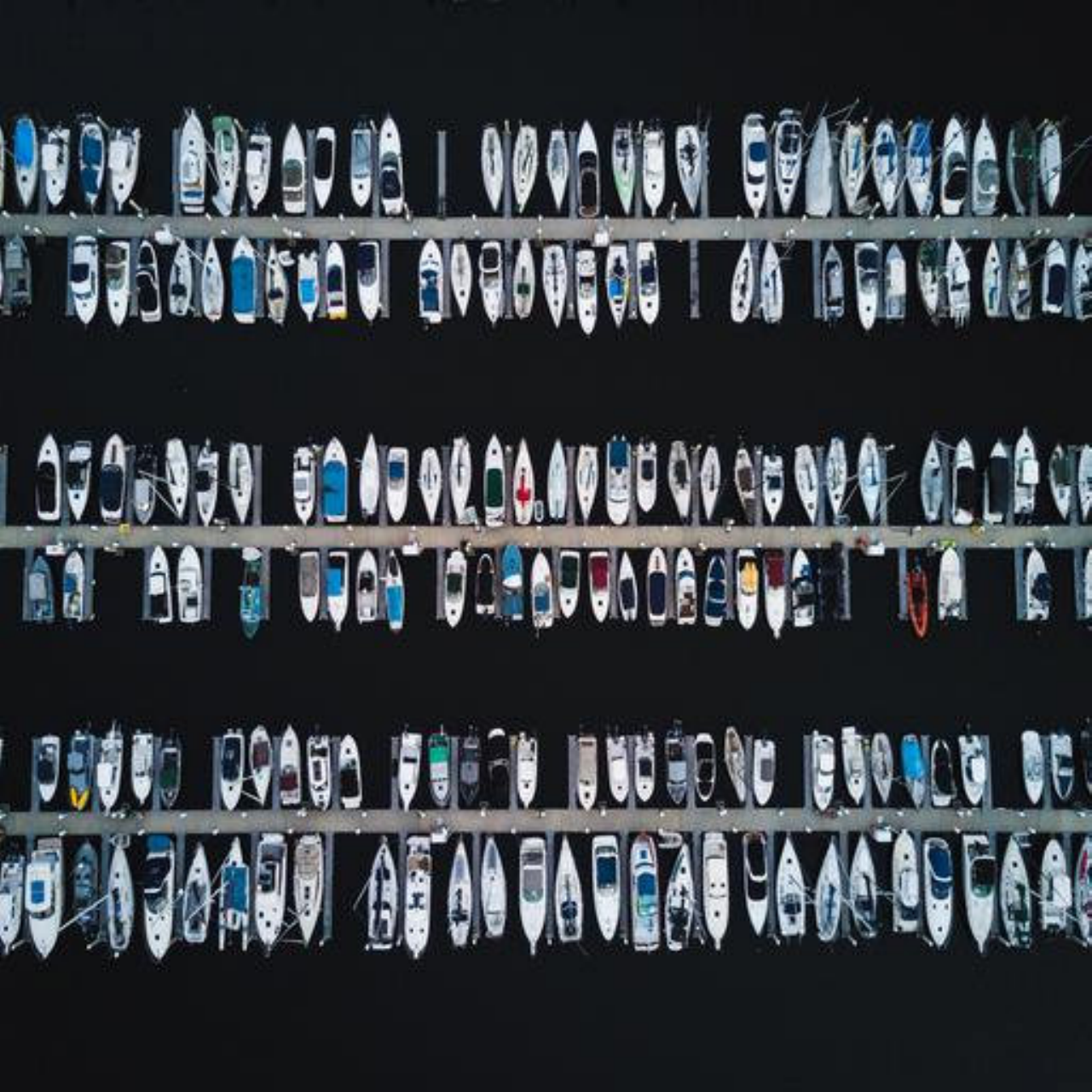}\vspace{0.51pt}
			\includegraphics[width=0.12\linewidth]{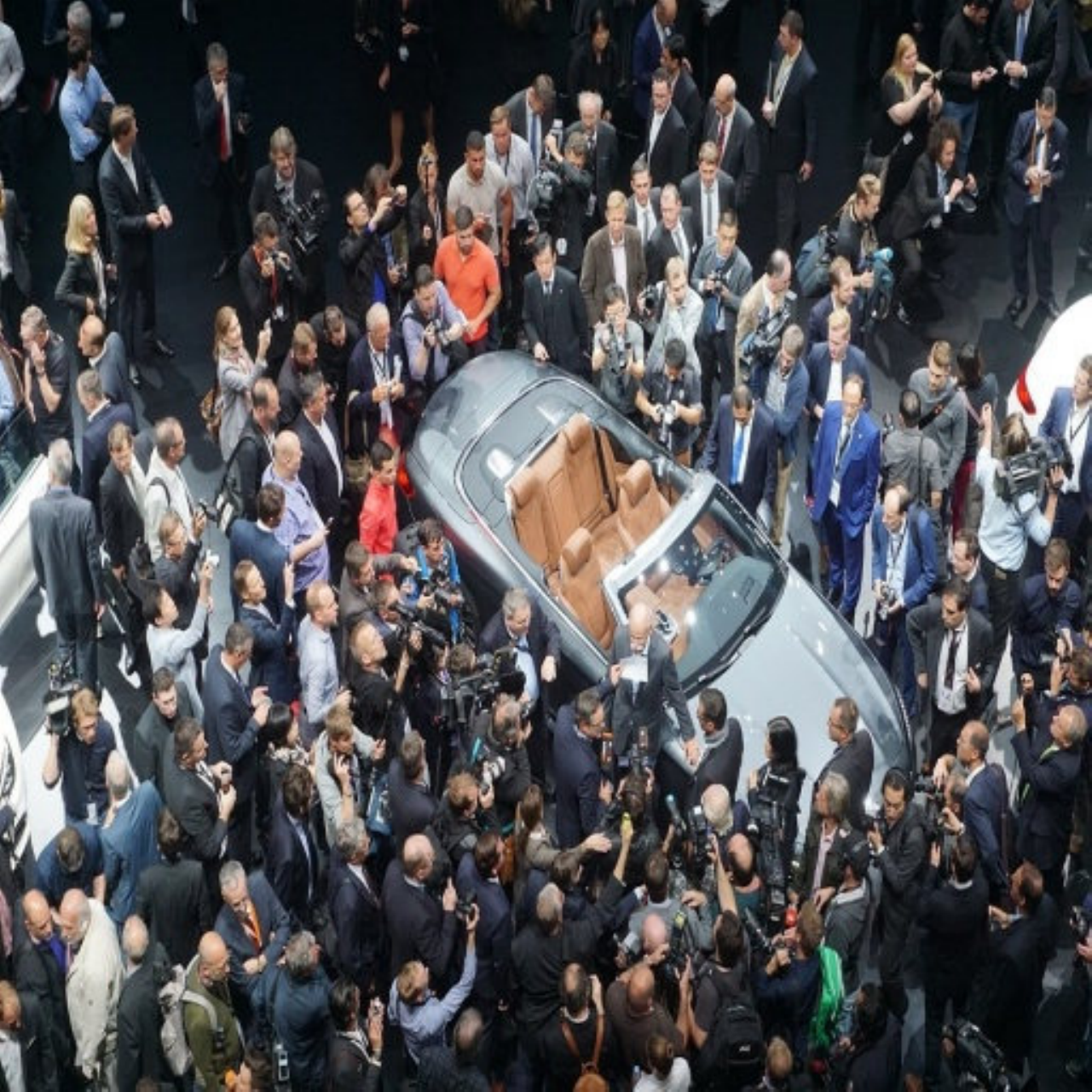}\vspace{0.51pt}
			\includegraphics[width=0.12\linewidth]{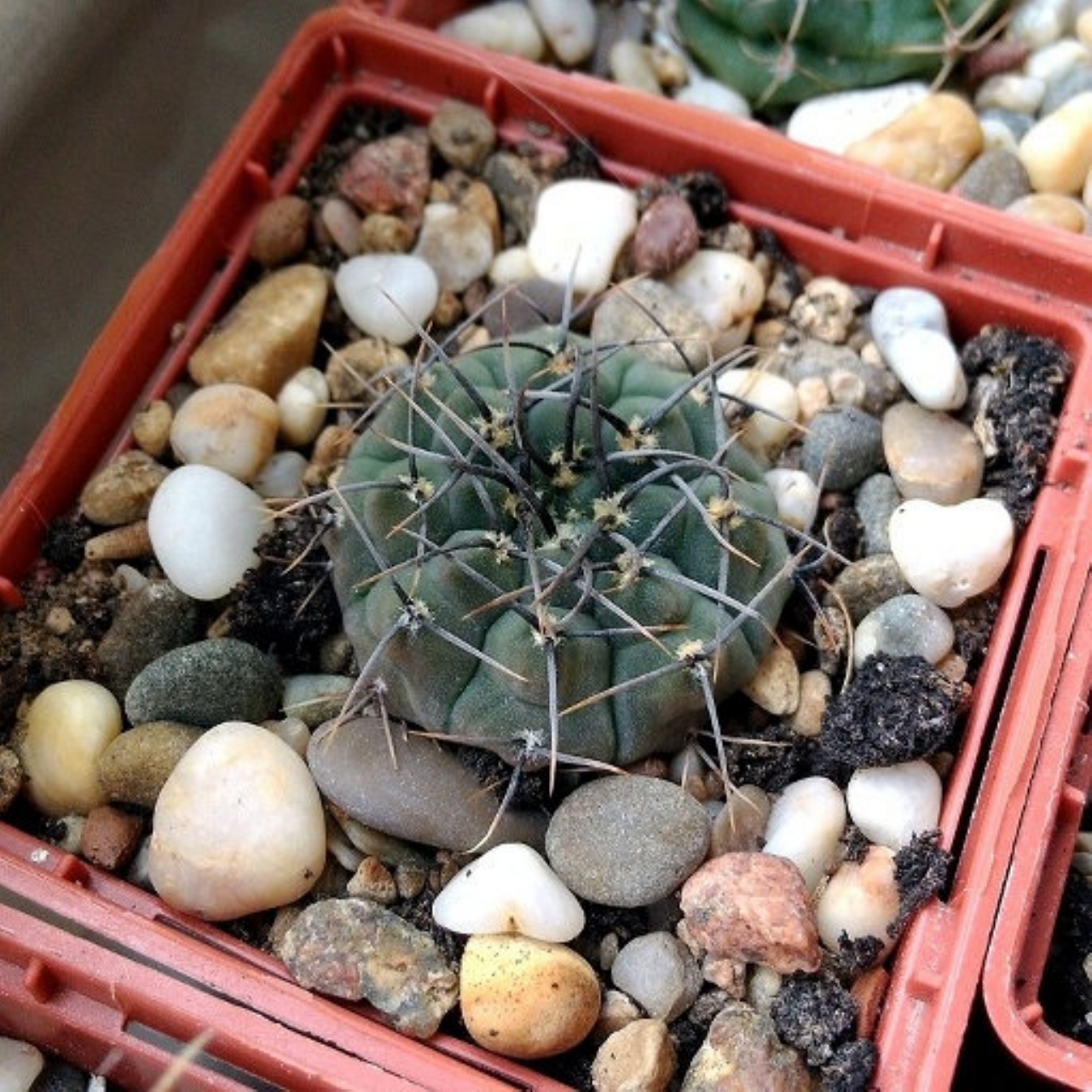}\vspace{0.51pt}
			\includegraphics[width=0.12\linewidth]{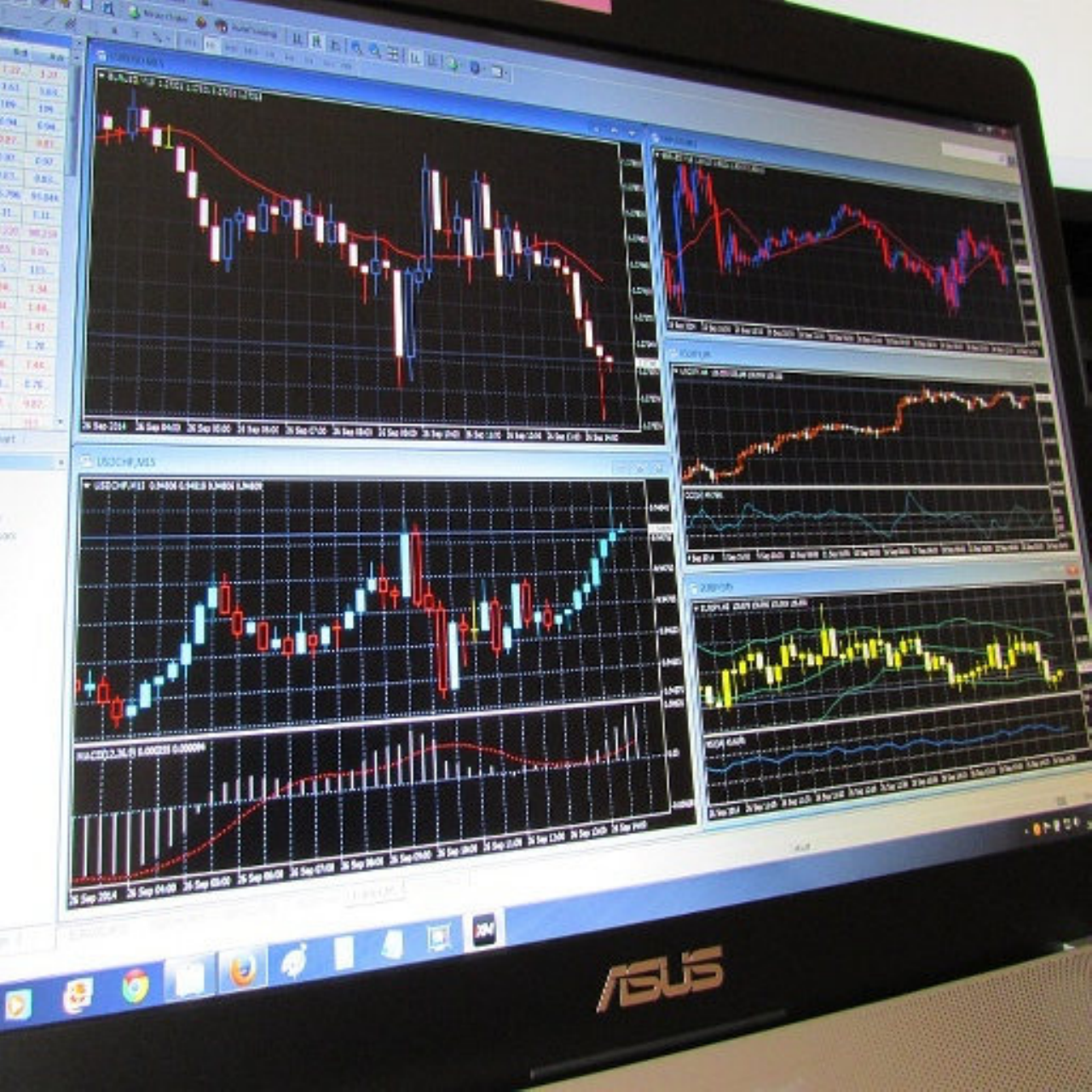}\vspace{0.51pt}
			\includegraphics[width=0.12\linewidth]{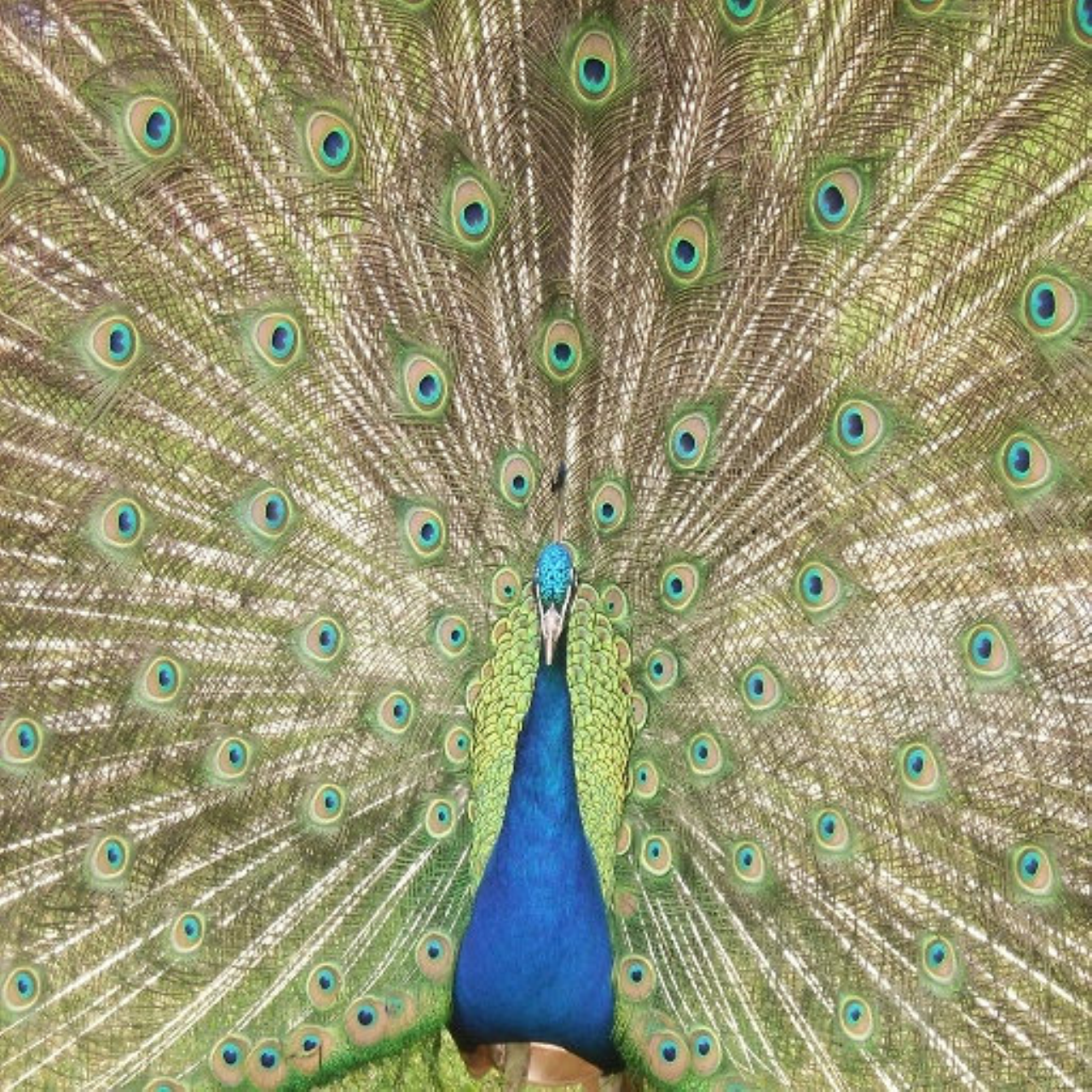}\vspace{0.51pt}
	\end{minipage}}
	\subfigure[Object part.]{
		\begin{minipage}[]{0.96\linewidth}
			\includegraphics[width=0.12\linewidth]{boat_1759.pdf}\vspace{0.51pt}
			\includegraphics[width=0.12\linewidth]{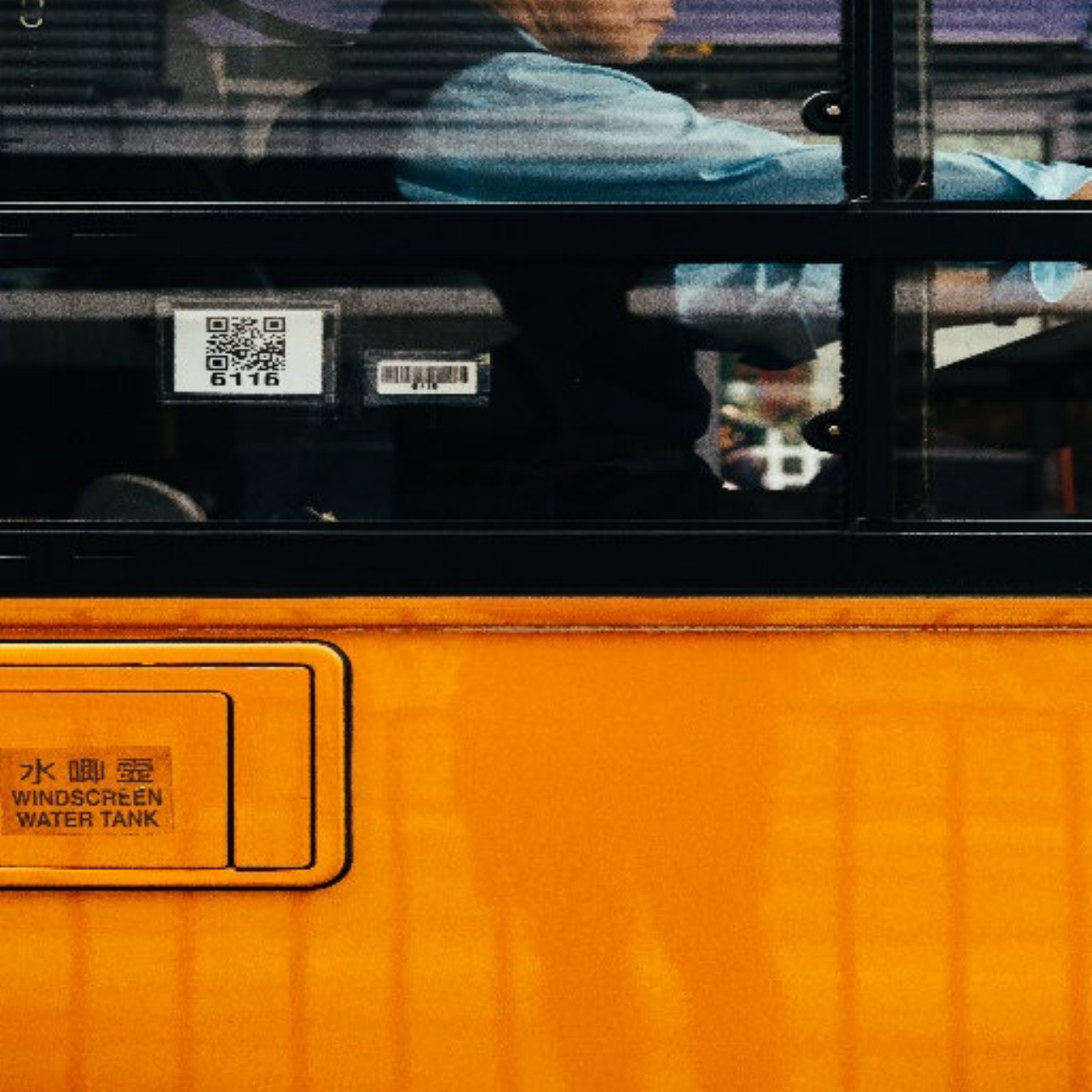}\vspace{0.51pt}
			\includegraphics[width=0.12\linewidth]{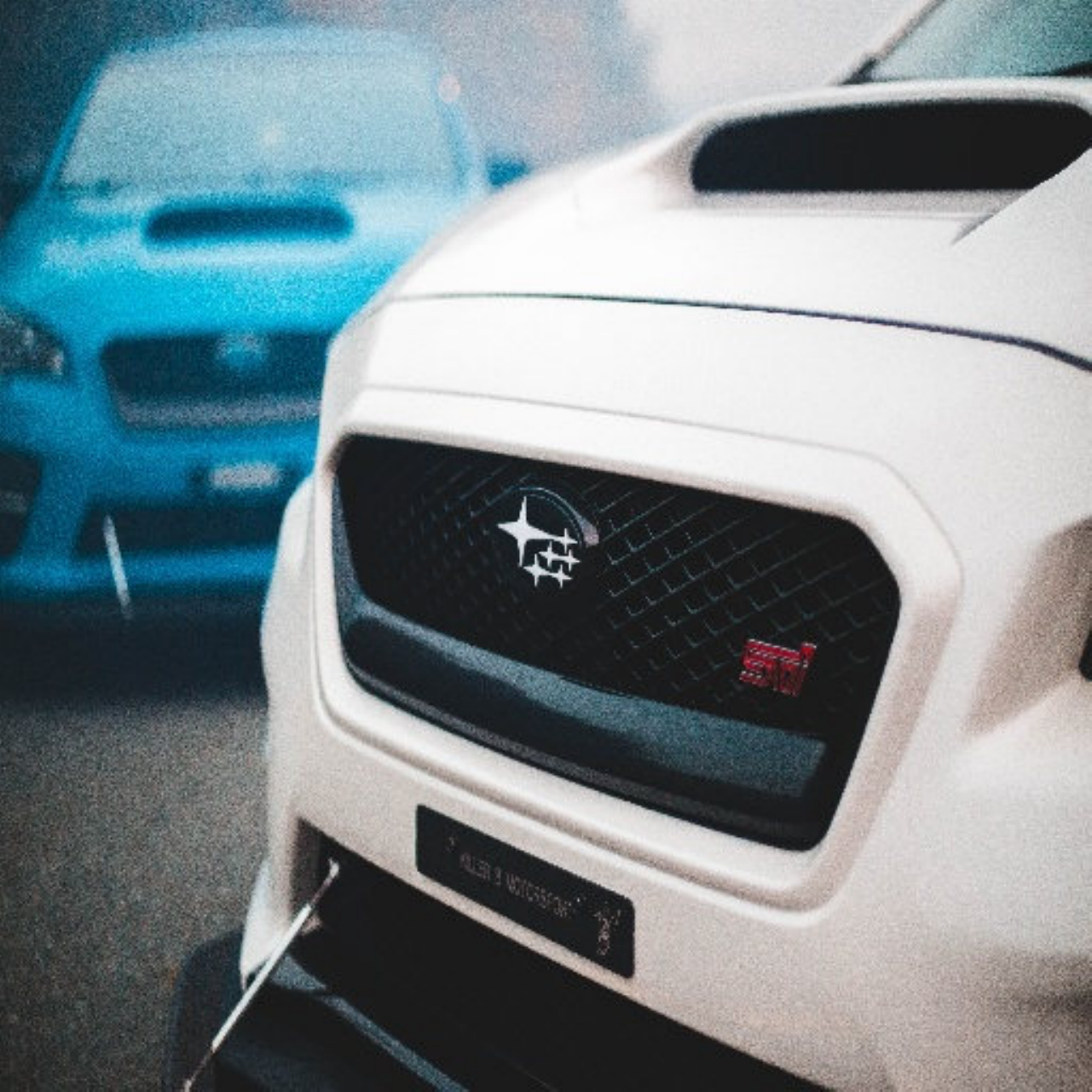}\vspace{0.51pt}
			\includegraphics[width=0.12\linewidth]{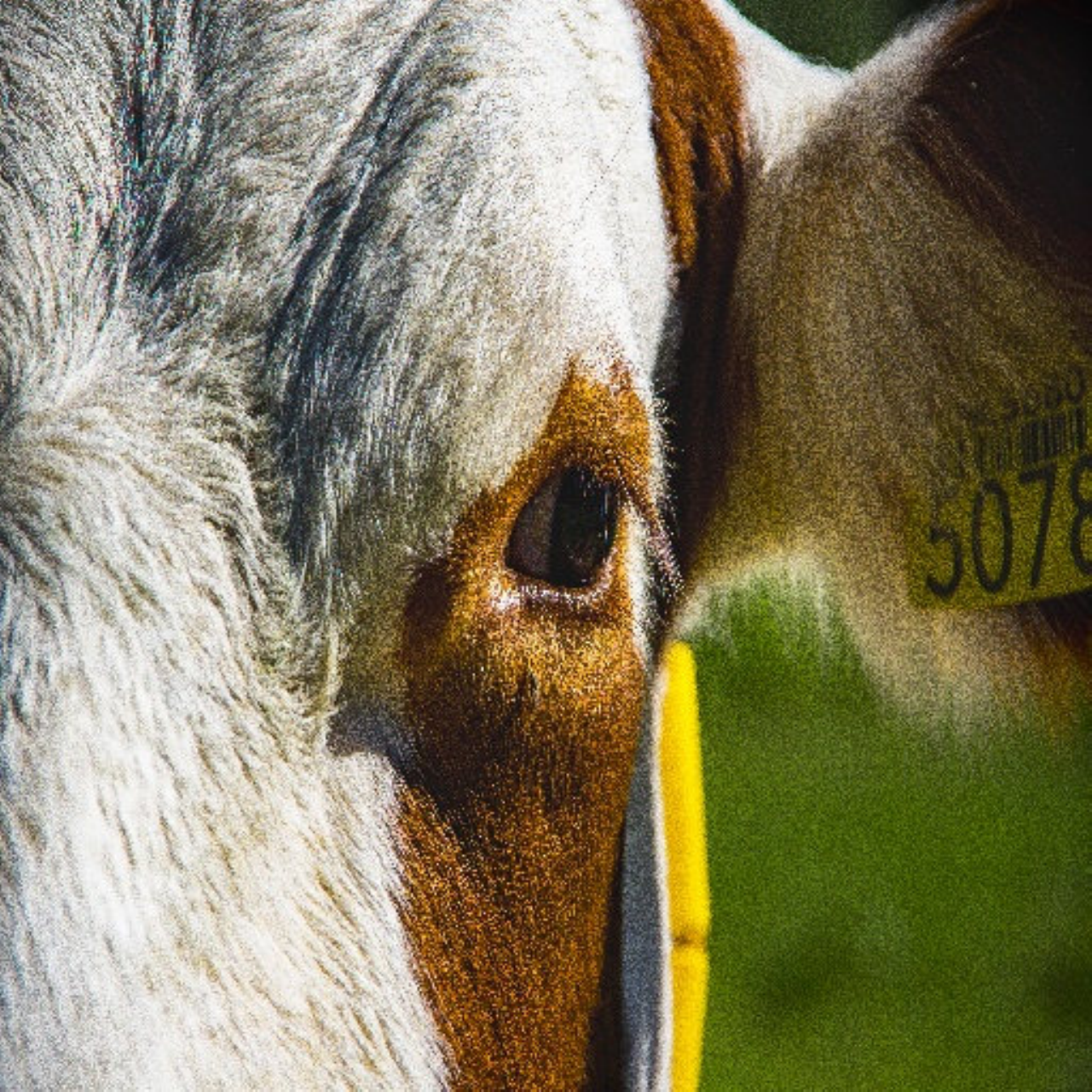}\vspace{0.51pt}
			\includegraphics[width=0.12\linewidth]{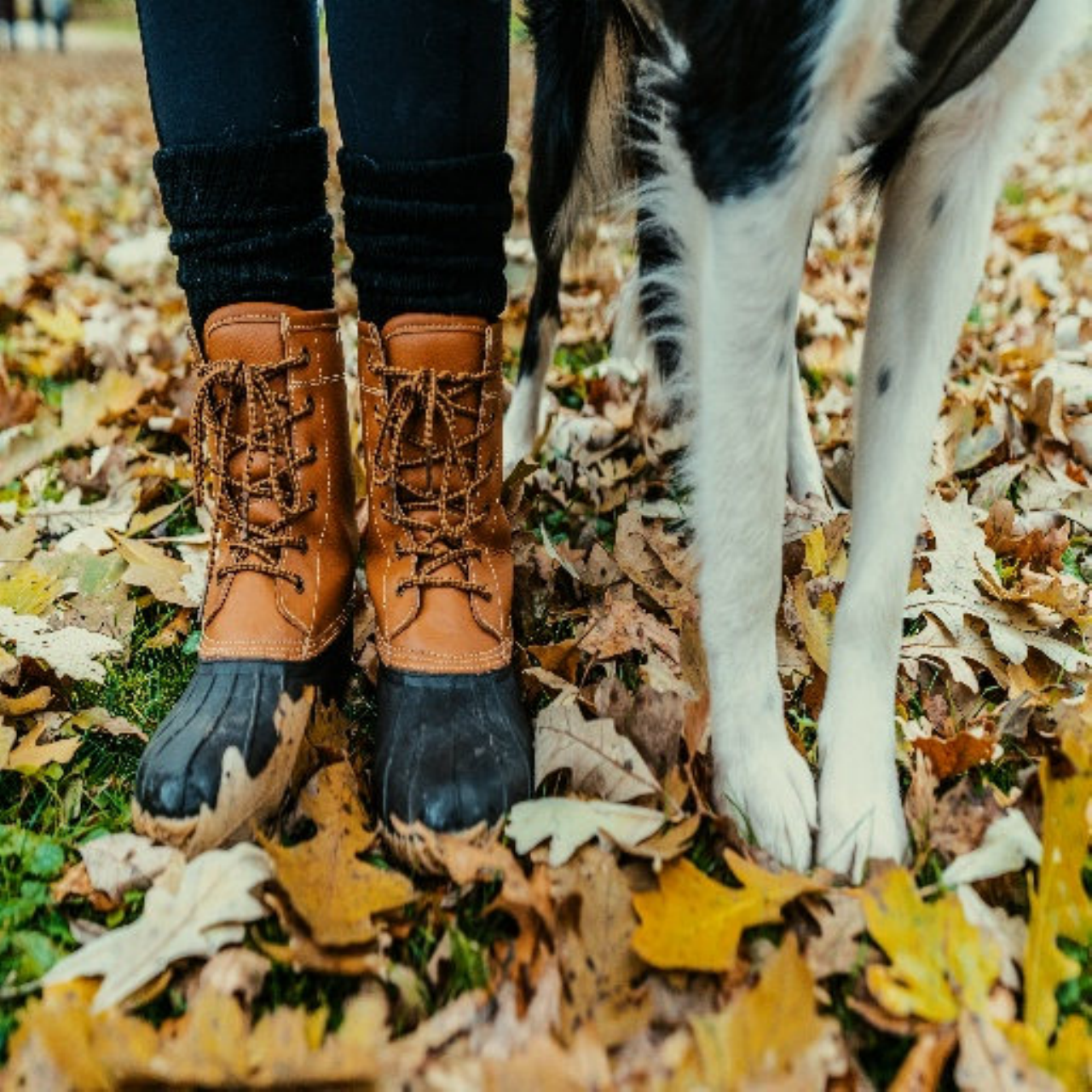}\vspace{0.51pt}
			\includegraphics[width=0.12\linewidth]{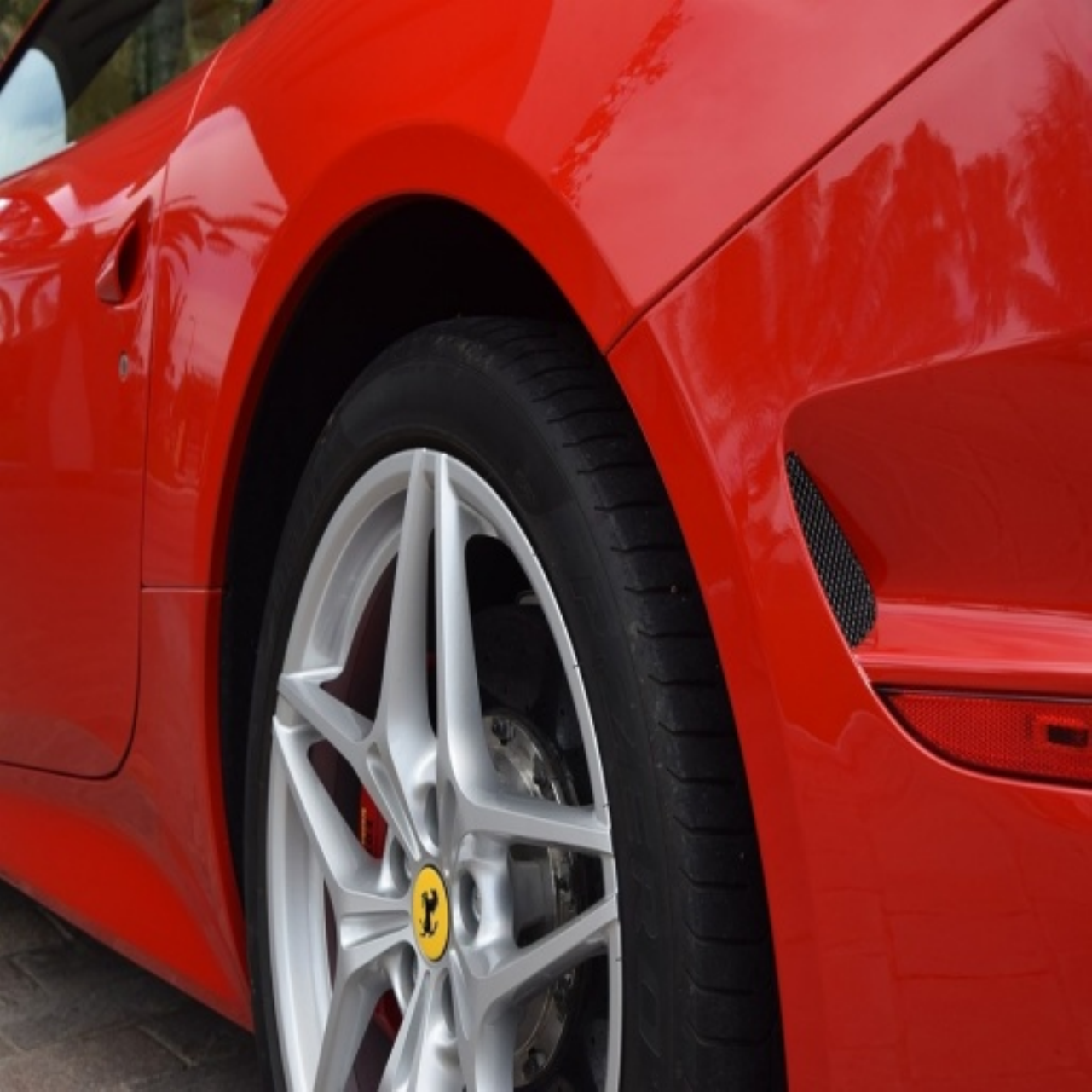}\vspace{0.51pt}
			\includegraphics[width=0.12\linewidth]{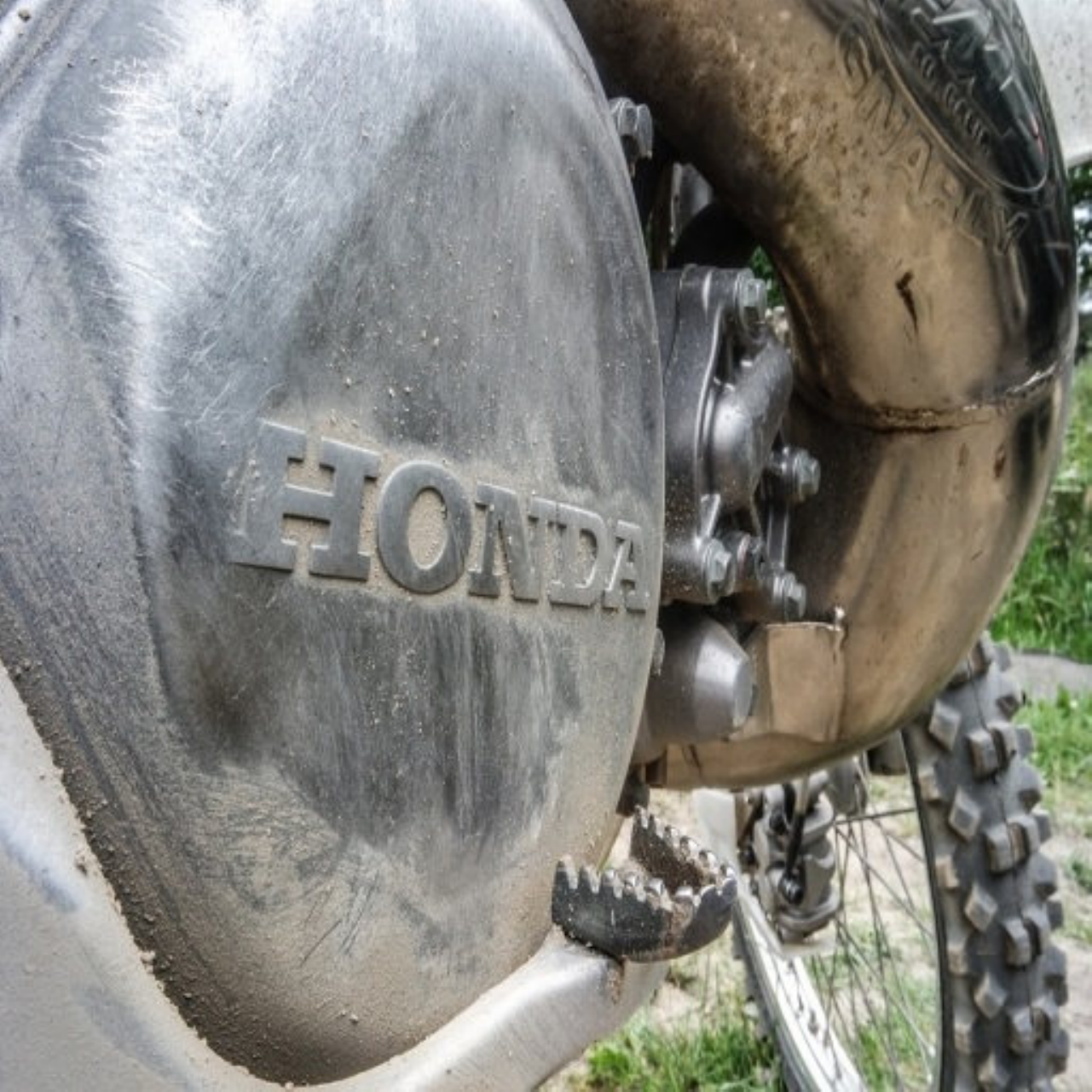}\vspace{0.51pt}
			\includegraphics[width=0.12\linewidth]{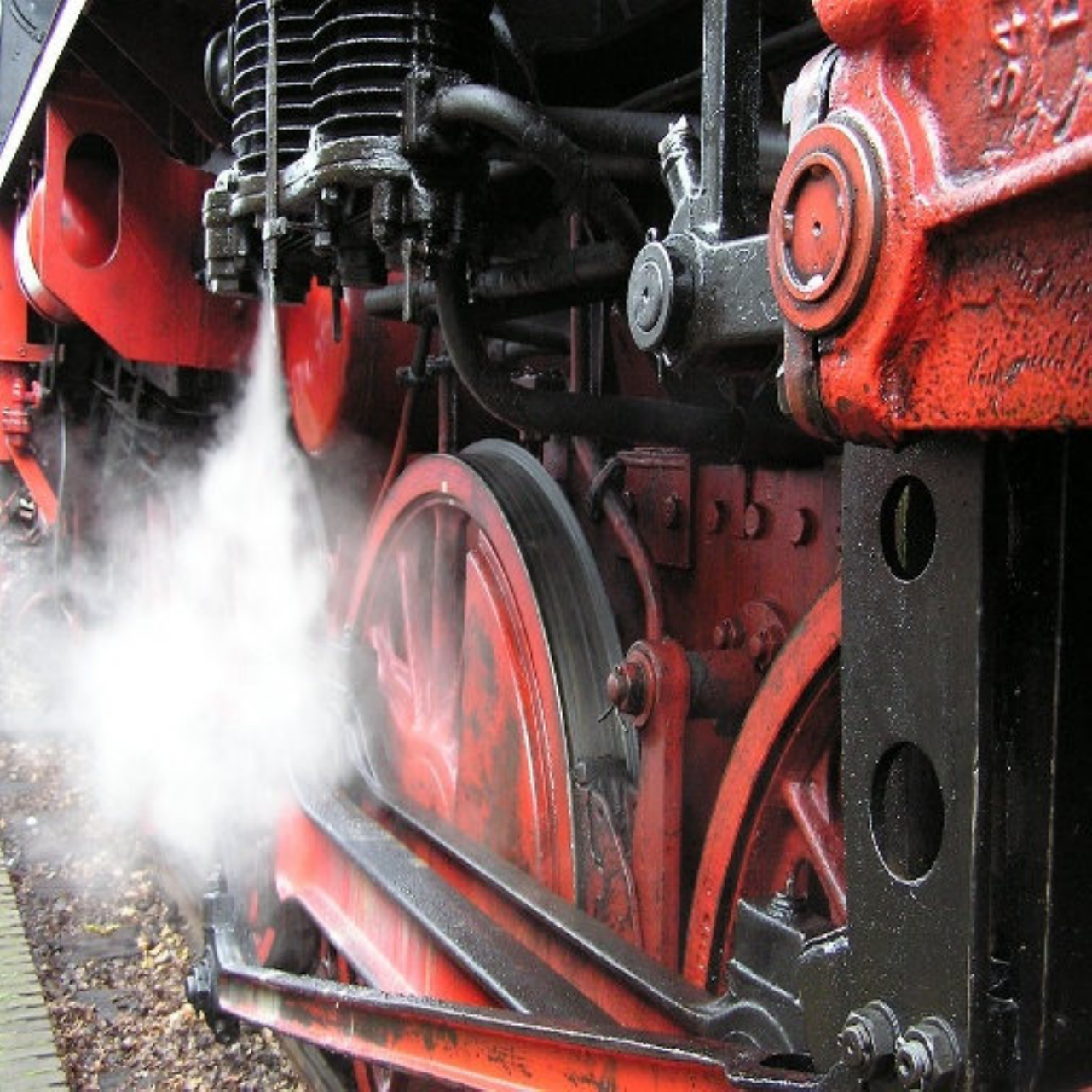}\vspace{0.51pt}
	\end{minipage}}
	\caption{Visualization and summary of common MAD failures of the competing segmentation models.}
	\label{fig:sum_failure}
\end{figure*}

Second,  RefineNet \citep{LinGS2017} shows strong competence in MAD, moving up at least three places in the aggressiveness and resistance rankings. The key to the model's success is the use of multi-resolution fusion and chained residual pooling for high-resolution and high-accuracy prediction. This is especially beneficial for segmenting images that only contain object parts (see Figs.~\ref{fig:boat}, \ref{fig:person} and~\ref{fig:sum_failure} (d)). In addition, the chained residual pooling module may also contribute to making better use of  background context. 

Third, the rankings of both EMANet \citep{Lixia2019} and PSANet \citep{zhao2018psanet} drop slightly in MAD compared with their excellent performance on PASCAL VOC. This reinforces the concern that ``delicate and advanced'' modules proposed in semantic segmentation (\eg, the attention mechanisms used in EMANet and PSANet) may have a high risk of overfitting to the extensively re-used benchmarks. As a result, the generalizability of EMANet and PSANet to the open visual world is inferior to segmentation models with simpler design, such as BlitzNet.

Fourth, the MAD performance of some low-weight architectures, ESPNetv2 and DiCENet, are worse than the competing models except FCN, which is consistent with their performance on PASCAL VOC. These results indicate that the trade-off between computational complexity and segmentation performance should be made in practice.




We conclude by visualizing the MAD failures of the segmentation models in Fig. \ref{fig:sum_failure}, where $\mathrm{mIoU} < 0.6$. Generally, the selected images are
visually much harder, containing novel appearances, untypical viewpoints, extremely large or small object scales, and only object parts.

\subsection{Ablation Studies}
\label{subsec_ablation}

\subsubsection{Sensitivity of Different Performance Measures}
\label{subsubsec_consistence}
We conduct ablation experiments to test the sensitivity of MAD results to different performance measures in Eq. \eqref{eq:sub}. Apart from mIoU, we adopt another two commonly used criteria in semantic segmentation, the frequency weighted IoU (FWIoU) and the mean pixel accuracy (MPA), which are defined  as
\begin{align}
\label{eq:FWIoU}
	\mathrm{FWIoU} = \frac{\sum_{y=0}^{\vert\mathcal{Y}\vert}{\frac{N_{yy}}{\sum_{y'=0}^{\vert\mathcal{Y}\vert}N_{yy'}+\sum_{y'=0}^{\vert\mathcal{Y}\vert}N_{y'y}-N_{yy}}}}{\sum_{y=0}^{\vert\mathcal{Y}\vert}\sum_{y'=0}^{\vert\mathcal{Y}\vert}N_{yy'}},
\end{align}
and
\begin{align}
	\label{eq:MPA}
	\mathrm{MPA} = \frac{1}{\vert\mathcal{Y}\vert+1}\sum_{y=0}^{\vert\mathcal{Y}\vert}{\frac{N_{yy}}{\sum_{y'=0}^{\vert\mathcal{Y}\vert}N_{yy'}}},
\end{align}
respectively, where $N_{yy'}$ has been specified in Eq. \eqref{eq:Nyy}. We follow the same procedure as in Sections \ref{subsec:sss} to construct two MAD sets under FWIoU and MPA, containing $662$ and $825$ distinct images, respectively. The subjective annotation procedure is also in accordance to Section \ref{subsec_subjective}. After collecting the ground-truth segmentation maps, we compare the global aggressiveness and resistance rankings under FWIoU and MPA, with those under mIoU, which serve as reference. Table~\ref{tab:srcc} shows the Spearman's rank correlation coefficient (SRCC) results, which are consistently high, indicating that the MAD rankings are quite robust to the choice of segmentation performance measures.

\subsection{Further Testing on WildDash}
\label{subsubsec_effectiveness}
We further validate the feasibility of MAD to expose semantic segmentation failures on another challenging dataset - WildDash \citep{2018WildDash} with unban-scene images. This dataset is designed specifically for testing segmentation models for autonomous driving.  We download $4,256$ images from its official website\footnote{\url{https://www.wilddash.cc/}} to construct $\mathcal{D}$.

\begin{table}[]
\centering
\caption{The sensitivity of the MAD rankings to the choice of segmentation performance measures in Eq. \eqref{eq:sub}. The SRCC values are computed using the aggressiveness and resistance rankings under mIoU as reference.}\label{tab:srcc}
\begin{tabular}{l|c}
\toprule
 & SRCC  \\ \hline
      Aggressiveness ranking under FWIoU    & 0.891 \\
       Aggressiveness ranking under  MPA    & 0.878 \\ \hline
        Resistance ranking under   FWIoU   & 0.879\\ Resistance ranking under   MPA   & 0.830 \\
\bottomrule
\end{tabular}
\end{table}

We test four recent semantic segmentation models, including HANet \citep{Choi2020Cars}, DGCNet \citep{zhang2019dual}, DANet \citep{fu2019dual}, and Panoptic-DeepLab \citep{cheng2020panoptic}, all trained mainly on the Cityscapes dataset \citep{Cordts2016Cityscapes}. Specifically, HANet includes an add-on module, with emphasis on  the distinct characteristics of  urban-scene images. We test HANet with the backbone ResNeXt-101~\citep{2017Aggregated} for its optimal performance. DGCNet exploits graph convolutional networks for capturing long-range contextual information. DANet uses position attention and channel attention modules, while Panoptic-DeepLab adopts the dual-decoder module for
semantic segmentation.  We follow the experimental protocols on PASCAL VOC with one key difference: we opt to a two-alternative forced choice (2AFC) method to minimize the human labeling effort. In particular,  instead of obtaining the ground-truth segmentation maps for the selected MAD images, participants are forced to choose the better prediction  from two competing models.
 Two experienced postgraduate students are invited to this subjective experiment.

\begin{figure*}[!ht]
\includegraphics[width=1\linewidth]{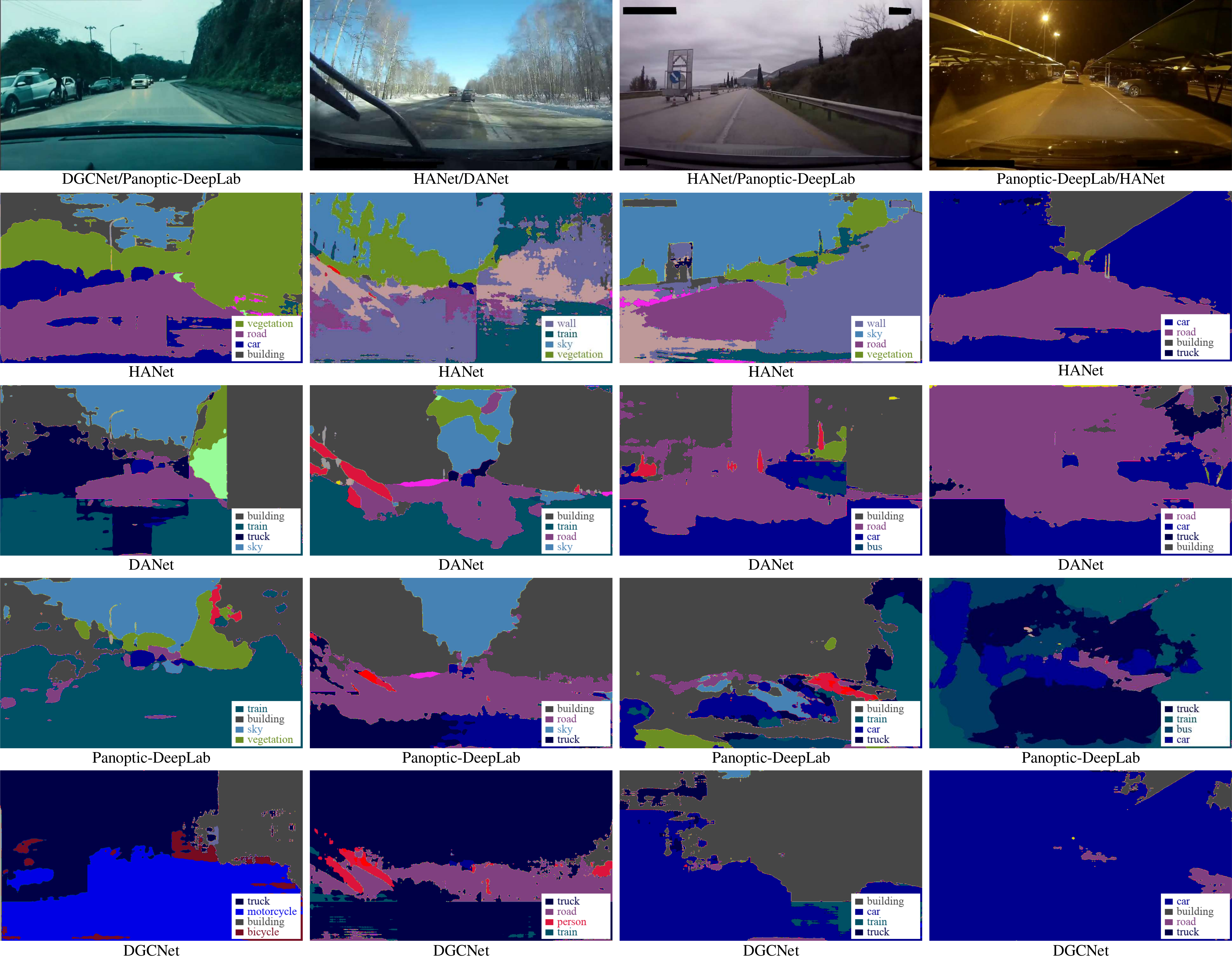}
\caption{The MAD images selected from WildDash~\citep{2018WildDash} together with the corresponding model predictions. It is clear that MAD successfully exposes strong transferable failures of the competing models.}
\label{fig:wilddash_harder}
\end{figure*}

We show the global ranking results in Table~\ref{tab:wilddashmad}, and make two consistent observations, which have been drawn from the experiments on PASCAL VOC. First, more training data leads to improved generalizability. For example, the best-performing HANet is trained on both coarsely and finely  annotated images in Cityscapes \citep{Cordts2016Cityscapes} as well as Mapillary~\citep{Mapillary}, while other test models are only trained on the finely annotated images in Cityscapes. Second, the incorporation of ``advanced'' modules in DGCNet, such as graph convolutional networks, may encourage overfitting to i.i.d. data from Cityscapes, with weak generalization to o.o.d. data from WildDash. We also show representative MAD samples and the four model predictions in Fig.~\ref{fig:wilddash_harder}. We find that MAD successfully exposes strong failures of the two associated models, which transfer well to falsify other segmentation methods.


\begin{table}[]
\centering
\caption{
\label{tab:wilddashmad}The MAD competition of four semantic segmentation models on WildDash \citep{2018WildDash}. The mIoU results are computed on Cityscapes \citep{Cordts2016Cityscapes}. $A$ and $R$ represent aggressiveness and resistance, respectively.}
\begin{tabular}{l|cccclllll}
\toprule Model  & mIoU/rank & $A$/$\Delta$ rank & $R$/$\Delta$ rank\\ \hline
HANet    & 0.832 / 1         & 1 / 0     & 1 / 0 \\
         DGCNet   & 0.820 / 2         & 4 / -2    & 4 / -2 \\
         DANet    & 0.815 / 3         & 2 / +1    & 2 / +1 \\
         Panoptic-DeepLab & 0.809 / 4  & 3 / +1  & 3 / +1 \\
\bottomrule
\end{tabular}
\end{table}

\section{Conclusion}
\label{sec_discu_reco}
In this paper, we have investigated the generalizability of semantic segmentation models by exposing their failures via the MAD competition. The main result is somewhat surprising: high accuracy numbers achieved by existing segmentation models on closed test sets do not result in robust generalization to the open visual world. In fact, most images selected by MAD are double-failures of the two associated segmentation models with strong transferability to falsify other methods. Fortunately, as the MAD samples are selected to best differentiate between two models, we are able to rank their relative performance in terms of aggressiveness and resistance through this failure-exposing process. We have performed additional experiments to verify that the MAD results are insensitive under different segmentation measures, and are consistent across different segmentation datasets. We hope that the proposed MAD for semantic segmentation would become a standard complement to the  close-set evaluation of a much wider range of dense prediction problems.

A promising future direction  is to jointly fine-tune the segmentation methods on the MAD set $\mathcal{S}$, which contains rich and diverse failures. We are currently testing this idea, and preliminary results indicate that existing models are able to learn from their MAD failures for improved segmentation performance. Moreover, if we relax the constraint of densely labeling $\mathcal{S}$ and opt for other forms of cheaper labels (\eg, object bounding boxes), we can significantly enlarge the size of $\mathcal{S}$, which may also be useful for boosting the segmentation performance. In addition, it is also interesting to see whether the MAD-selected images in the context of semantic segmentation are equally transferable to falsify computational models of related tasks, such as instance segmentation and object detection.








%
%

\begin{acknowledgements}
This work was partially supported by the National Key R$\&$D Program of China under Grant 2018AAA0100601, and the National Natural Science Foundation of China under Grants 62071407 and 61822109.
\end{acknowledgements}

%
\section*{Conflict of interest}
The authors declare that they have no conflict of interest.


\bibliographystyle{spbasic}      
\bibliography{main}   

%
%

\end{document}